\documentclass{article}

\usepackage{arxiv}
\usepackage[numbers]{natbib}

\usepackage[utf8]{inputenc} 
\usepackage[T1]{fontenc}    
\usepackage{hyperref}       
\usepackage{url}            
\usepackage{booktabs}       
\usepackage{amsfonts}       
\usepackage{nicefrac}       
\usepackage{microtype}      
\usepackage{lipsum}
\usepackage{graphicx}
\usepackage{caption}

\makeatletter
\def\blfootnote{\gdef\@thefnmark{}\@footnotetext}
\makeatother

\theoremstyle{plain}
\newtheorem{theorem}{Theorem}[section]

\newtheorem{lemma}[theorem]{Lemma}

\theoremstyle{definition}
\newtheorem{definition}[theorem]{Definition}

\theoremstyle{remark}

\title{Proactive DP: A Multiple Target Optimization Framework for DP-SGD}

\author{Marten van Dijk$^{1,2,3*}$, Nhuong V. Nguyen$^{4*}$, Toan N. Nguyen$^{4,5,6}$, \\ \textbf{Lam M. Nguyen}$^{}$\textbf{,}  \textbf{Phuong Ha Nguyen}$^{8}$ \\
$^{1}$ CWI Amsterdam, The Netherlands\\
$^{2}$ Department of Computer Science, Vrije Universiteit Amsterdam, The Netherlands \\
$^{3}$ 
Department of Electrical and Computer Engineering, University of Connecticut, CT, USA\\
$^{4}$ Department of Computer Science and Engineering, University of Connecticut, CT, USA \\
$^{5}$ Faculty of Information Technology, University of Science, Ho Chi Minh, Vietnam\\
$^{6}$ Vietnam National University, Ho Chi Minh city, Vietnam\\
$^{7}$ IBM Research, Thomas J. Watson Research Center, Yorktown Heights, NY, USA\\
$^{8}$ eBay, CA, USA\\
\\
\texttt{marten.van.dijk@cwi.nl}, \texttt{nhuong.nguyen@uconn.edu}, \texttt{nntoan@fit.hcmus.edu.vn},  \\
\texttt{LamNguyen.MLTD@ibm.com}, \texttt{phuongha.ntu@gmail.com}}

\begin{document}
\maketitle

\blfootnote{$^{*}$ These authors contributed equally.}

\blfootnote{}
\blfootnote{In memory of our dear friend and author of this paper Nhuong Nguyen.}

\begin{abstract}
We introduce a multiple target optimization framework for DP-SGD referred to as pro-active DP. In contrast to traditional DP accountants, which are used to track the expenditure of privacy budgets, the pro-active DP scheme allows one to {\it a-priori} select parameters of DP-SGD based on a fixed privacy budget (in terms of $\epsilon$ and $\delta$) in such a way to optimize the anticipated utility (test accuracy) the most.
To achieve this objective, we first propose significant improvements to the moment account method, presenting a closed-form $(\epsilon,\delta)$-DP guarantee that connects all parameters in the DP-SGD setup. 
We show that DP-SGD is $(\epsilon<0.5,\delta=1/N)$-DP if $\sigma=\sqrt{2(\epsilon +\ln(1/\delta))/\epsilon}$ with $T$ at least $\approx 2k^2/\epsilon$ and $(2/e)^2k^2-1/2\geq \ln(N)$, where $T$ is the total number of rounds, and $K=kN$ is the total number of gradient computations where $k$ measures $K$ in number of epochs of size $N$ of the local data set. We prove that our expression is close to tight in that if $T$ is more than a constant factor $\approx 4$ smaller than the lower bound $\approx 2k^2/\epsilon$, then the  $(\epsilon,\delta)$-DP guarantee is violated. 
The above DP guarantee can be enhanced in that
DP-SGD is $(\epsilon, \delta)$-DP if  
$\sigma = \sqrt{2(\epsilon+\ln(1/\delta))/\epsilon}$ with 
$T$ at least $\approx 2k^2/\epsilon$ together with two additional, less intuitive, conditions that allow larger $\epsilon\geq 0.5$. 
Our DP theory allows us to create a utility graph and DP calculator. These tools link privacy and utility objectives and search for optimal experiment setups, efficiently taking into account both accuracy and privacy objectives, as well as implementation goals. We furnish a comprehensive implementation flow of our proactive DP, with rigorous experiments to showcase the proof-of-concept.\footnote{We remember Nhuong, our dear friend, whose presence and kindness we so much miss. His careful and creative thinking together with his uncanny implementation skills leaves this paper as his footprint. His promising future cut short, he will live on in our heart and his spark forever visible in this paper and his scientific work.}
\end{abstract}

\section{Introduction}
\label{sec:intro_contribution}

DP-SGD \citep{abadi2016deep} was introduced for private machine learning training as it adapts distributed Stochastic Gradient Descent (SGD)~\cite{robbins} with Differential Privacy (DP)~\cite{BookDP}. Many different DP notions have been developed for better tracking the expenditure of privacy budgets during the training process such as ($\epsilon,\delta$)-DP~\cite{BookDP}, Concentrated Differential Privacy (CDP)~\cite{CDP}, Renyi-DP~\cite{BS15}, zero-CDP (zCDP)~\cite{zCDP}, f-DP~\cite{dong2021gaussian}. Generally, DP comes from adding Gaussian noise ${\cal N}(0,C^2\sigma^2{\bf I})$ to local (client-computed) mini-batch SGD updates after performing a clipping operation $x\rightarrow [x]_C=x/\max \{1,\| x\|/C\}$. 

Existing DP notions are inadequate for optimizing the parameters for DP-SGD to achieve a given privacy budget, utility (test accuracy) goal, and implementation goal (communication efficiency). Consequently, we initiated the study of a fresh DP notion called proactive DP. Our DP framework consists of three significant components. First, a tight closed DP formula encompassing all experimental setup parameters (the number of data points $N$, the mini-batch or sample size $s$  during local SGD iterations, the step size scheme, etc.) and privacy budget (in the form of an ($\epsilon$,$\delta$)-DP target), utility goal (i.e., test accuracy), and implementation goal (in particular, the number of communication rounds $T$ between server and client). We demonstrate a specific relationship among privacy parameters $\epsilon$, $\delta$, and $\sigma$ given by the closed form formula, $\sigma = \sqrt{2(\epsilon +\ln(1/\delta))/\epsilon}$. This formula is vital as it helps to establish a relationship between privacy budget and utility goal. Notably, noise variance $\sigma$ directly correlates with the test accuracy of the model. However, we aim to avert performing any blind private training because we can only determine the impact of $\sigma$ on test accuracy after the private training process concludes. As our second component, we introduce the utility graph method, which allows us to learn the impact of $\sigma$ on the test accuracy without performing any private training. This graph assists us in determining good privacy parameters (i.e., $\epsilon$ and $\delta$) and the corresponding values of noise $\sigma$ that meet privacy and utility goals. Finally, we employ a tool called DP calculator to efficiently calculate the remaining parameters of the optimal experiment setup, such as the stepsize scheme, sampling scheme, and the number of communication rounds. Our DP framework primarily contains a tight closed DP formula, utility graph, and DP calculator. It's worth mentioning that a tight and closed DP formula leads to better optimal experiment settings and a more efficient DP calculator.

We devised our tight closed DP formula by non-trivially improving the moment accountant technique \citep{abadi2016deep}. This approach has lacked a tight bound \citep{dong2021gaussian} due to the high analysis complexity which the authors of \citep{abadi2016deep} did not pursue, but did lead to a straightforward closed non-tight DP formula for the experimental framework. To demonstrate the tightness of our new DP formula, we leverage the findings from $f$-DP \citep{dong2021gaussian}. The $f$-DP framework supersedes all other existing frameworks because it has all the relevant information to derive known DP metrics. However, as stated in \citep{dong2021gaussian}, the "disadvantage is that the expressions it yields are more unwieldy: they are computer evaluable, so usable in implementations, but do not admit simple closed form" for the $f$-DP model.

Basically, together with a detailed implementation flow of proactive DP: 
\begin{itemize}
    \item We non-trivially improve the analysis of the moment accountant method in \citep{abadi2016deep} and show for the first time that $(\epsilon,\delta)$-differential privacy can be achieved for 
    \begin{equation} \sigma = \sqrt{2(\epsilon +\ln(1/\delta))/\epsilon} \label{eq:sed}
    \end{equation}
    in  parameter settings 
with (a) a reasonable DP guarantee by choosing $\delta\leq 1/N$ and $\epsilon$ smaller than $0.5$, where (b) for the total number $K$ of gradient computations over all local rounds performed on the local data set we have 
 $(2/e)^2
\cdot k^2 \geq 1/2 + \ln(1/\delta)$ with $k=K/N$ measuring $K$ in number of epochs of size $N$ (this condition is generally met in practice),
 and (c) 
$T$ is at least another constant 
($\approx 2$) times
$k^2/\epsilon$.



A precise formulation of our main result is given in Theorem \ref{thm:Main}. We notice that this theorem is extracted as a special case of the more general (but less readable and less intuitive) Theorem \ref{thm:simpleF} which applies to {\em any} choice of $\epsilon$ and $\delta$. In particular, the more general theorem can be used for $\epsilon\geq 0.5$ and can also be used in our proactive DP framework.

\item 
Confirmed by simulations, we show that by setting $T$ equal to the lower bound $T\approx 2k^2/\epsilon$, we optimize accuracy (as this minimizes the number of times/rounds when noise is aggregated into the global model at the server) and minimize round complexity. By using the $f$-DP framework, we prove that $T$'s condition of being at least  $\approx 2k^2/\epsilon$ cannot be made weaker in that $2k^2/\epsilon$ cannot be divided by more than a constant factor $\approx 4$ (otherwise, this conflicts with an asymptotical result proved by the $f$-DP framework). 
We conclude that setting 
\begin{equation}
T\approx 2k^2/\epsilon \label{eq:Tapprox}
\end{equation}
leads to a close to tight $(\epsilon,\delta)$-DP guarantee which also optimizes accuracy and minimizes round complexity (we are the first to show such a kind of tightness result for the moment accountant method).

\item We discuss the concept of a utility graph and DP calculator in order to efficiently determine suitable parameter settings based on our theory. Simulations based on (\ref{eq:sed}) and (\ref{eq:Tapprox}) show a significantly smaller $\epsilon$. For example, our theory applies to $\epsilon=0.15$ for the non-convex problem of the simple neural network LeNet \citep{lecun1998gradient} with cross entropy loss function for image classification of MNIST~\cite{lecun-mnisthandwrittendigit-2010} at a test accuracy of 93\%, compared to 98\% without differential privacy. A detailed comparison to the current state-of-the art is presented in Section~\ref{sec:related_work}.


\end{itemize}
\noindent
{\bf Outline:} We provide background in Section \ref{sec:DP-SGD}, where we define $(\epsilon,\delta)$-differential privacy, explain DP-SGD  as introduced by 
\citep{abadi2016deep}, and shortly introduce the $f$-DP framework. In Section \ref{sec:Main} we explain our main theory, where we start by discussing the theoretical result of the moment accountant method of \citep{abadi2016deep} and its limitation, which we improve  leading to our main contribution as given in (\ref{eq:sed}) with a tightness result for (\ref{eq:Tapprox}) based on the $f$-DP framework. We discuss the concept of a utility graph and show how our theory can be used to determine parameter settings for DP-SGD.
Experiments are in Section \ref{sec:experiment}.  A more general asynchronous SGD framework, the detailed differential privacy proofs and analysis,  additional experiments with extra details and a proposal for an algorithm that regularly updates parameters in DP-SGD are in the appendices.

\section{Differential Private SGD (DP-SGD)} \label{sec:DP-SGD}

\subsection{\texorpdfstring{$(\epsilon,\delta)$- Differential Privacy}{}} \label{sec:DP}

Differential privacy \citep{dwork2006calibrating, dwork2011firm,dwork2014algorithmic,dwork2006our} defines privacy guarantees for algorithms on databases, in our case a client's sequence of mini-batch gradient computations on his/her training data set. The guarantee quantifies into what extent the output  of a client (the collection of updates communicated to the server) can be used to differentiate among two adjacent training data sets $d$ and $d'$ (i.e., where one set has one extra element compared to the other set).

\begin{definition} \label{defDP} A randomized mechanism ${\cal M}: D \rightarrow R$ is $(\epsilon, \delta)$-DP (Differentially Private) if for any adjacent $d$ and $d'$ in $D$ and for any subset $S\subseteq R$ of outputs,
$$ Pr[{\cal M}(d)\in S]\leq e^{\epsilon} Pr[{\cal M}(d')\in S] + \delta,$$
where the probabilities are taken over the coin flips of mechanism ${\cal M}$.
\end{definition}
The privacy loss incurred by observing $o$ is given by
$$ L^{o}_{{\cal M}(d) \| {\cal M}(d')} = \ln \left( \frac{Pr[{\cal M}(d)=o]}{Pr[{\cal M}(d')=o]} \right).
$$
As explained in \citep{dwork2014algorithmic} $(\epsilon, \delta)$-DP ensures that
for all adjacent $d$ and $d'$ the absolute value of privacy loss will be bounded by $\epsilon$ with probability at least $1-\delta$. 
The larger $\epsilon$ the more certain we are about which of $d$ or $d'$ caused observation $o$. In order to have a reasonable security guarantee we assume $\epsilon<0.5$ such that $e^\epsilon<1.65$ is somewhat small.
When using differential privacy in machine learning we typically use $\delta=1/N$ (or $1/(10N)$) inversely proportional with the data set size $N$.

In order to prevent data leakage from  inference attacks in machine learning \citep{lyu2020threats} such as the deep leakage from gradients attack
\citep{ligengzhu201deepleakage,zhao2020idlg,geiping2020inverting} or the membership inference attack
\citep{shokri2017membership,nasr2019MIA,song2019MIA} a range of privacy-preserving methods have been proposed. 
Privacy-preserving solutions 
for federated learning are Local Differential Privacy (LDP) solutions \citep{abadi2016deep,abhishek2018privateFL,mohammad2020privateFL,stacey2018privateFL,meng2020privateFL,duchi2014local} and Central Differential Privacy (CDP) solutions \citep{mohammad2020privateFL,robin2017privateFL,mcmahan2017learning,nicolas2018privateFL,Yu2019CDP}. 
In LDP, the noise for achieving differential privacy is computed locally at each client and is added to the updates before sending to the server -- in this paper we also consider LDP. In CDP, a trusted server (aka trusted third party) aggregates received client updates into a global model; in order to achieve differential privacy the server adds noise to the global model before communicating it to the clients.



\subsection{DP-SGD}
\label{sec:dpsgd}


\begin{algorithm}[!ht]
\caption{DP-SGD: Local Model Updates with Differential Privacy}
\label{alg:DPs}
\begin{algorithmic}[1]
\Procedure{LocalSGDwithDP}{$d$}

\For{$i\in \{0,\ldots, T-1\}$}
 
       \State Receive the current global model $\hat{w}$ from  Server.

        \State Uniformly sample a random set $\{\xi_h\}_{h=1}^{s_i}\subseteq d$ 
        
    \State $h=0$, $U=0$ 
    \While{$h< s_{i}$}
        \State $g = [\nabla f(\hat{w}, \xi_h)]_C$ 
        \State ${U} = {U} + g$  
        \State $h$++
    \EndWhile
    
    \State $n\leftarrow {\cal N}(0,C^2 \sigma^2\textbf{I})$ 
    \State $U=U+n$
    
    \State Send $(i, U)$ to the Server. 
\EndFor
   
\EndProcedure

\end{algorithmic}
\end{algorithm}
   


We analyse the Gaussian based differential privacy method, called DP-SGD, of \citep{abadi2016deep}, depicted in Algorithm \ref{alg:DPs} in a distributed setting as described above. 
Rather than using the gradient $\nabla f(\hat{w}, \xi)$ itself, DP-SGD uses its clipped version $[\nabla f(\hat{w}, \xi)]_C$ where $[x]_C= x/\max\{1,\|x\|/C\}$. 
Clipping is needed because 
in general we cannot assume a bound $C$ on the gradients (for example, the bounded gradient assumption  is in conflict with strong convexity \citep{nguyen2018sgd}), yet
the added gradients need to be bounded by some constant $C$ in order for the DP analysis to go through.
%
%

DP-SGD uses a  mini-batch approach where before the start of the $i$-th local round a random min-batch of sample size $s_i$ is selected out of a local data set $d$ of size $|d|=N$. Here, we slighty generalize DP-SGD's original formulation which uses a constant $s_i=s$ sample size sequence, while our analysis will hold for a larger class of sample size sequences. The inner loop maintains the sum $U$ of gradient updates where each of the gradients correspond to the same local model $\hat{w}$ until it is replaced by a newer global model at the start of the outer loop. 
At the end of each local round the sum of updates $U$ is obfuscated with Gaussian noise ${\cal N}(0,C^2\sigma^2)$ added to each vector entry, and the result is transmitted to the server. 
%
The noised $U$ is  transmitted to the server who adds $U$ times the round step size $\bar{\eta}_i$ to its global model $\hat{w}$ (we discount averaging the sum represented by $U$ 
by scaling the step size inversely with $s_i$).
As soon as all clients have submitted their updates, the resulting new global model $\hat{w}$ is broadcast to all clients, who in turn replace their local models with the newly received global model (at the start of the outer loop). 



\subsection{\texorpdfstring{Tight $f$-DP Framework}{}}
\label{sec:GDPm}

Appendix \ref{app:GDP} summarizes the recent work by \citep{dong2021gaussian} that introduces the $f$-DP framework based on hypothesis testing. $f$-DP has $(\epsilon,\delta)$-DP as a special case in that a mechanism is $(\epsilon,\delta)$-DP if and only if it is $f_{\epsilon,\delta}$-DP with $f_{\epsilon,\delta}(\alpha) =
\max \{ 0, 1-\delta - e^{\epsilon}\alpha, (1-\delta-\alpha)e^{-\epsilon}\}$. They prove that DP-SGD is 
$C_{s/N}(G_{\sigma^{-1}})^{\otimes T}$-DP where $C_{s/N}$ is an operator representing the effect of subsampling, $G_{\sigma^{-1}}$ is a Gaussian function characterizing the differential privacy (called Gaussian DP) due to adding Gaussian noise, and operator $\otimes T$ describes composition over $T$ rounds. $C_{s/N}(G_{\sigma^{-1}})^{\otimes T}$-DP can be translated into a tight $(\epsilon,\delta)$-DP formulation.

Towards understanding how to a-priori set parameters for best utility and minimal privacy leakage, the tight $f$-DP formulation for DP-SGD can be translated into sharp privacy guarantees. However, 
as stated in the introduction by a citation from \citep{dong2021gaussian}, the expressions it yields are more unwieldy. 
Precisely, as said in \citep{dong2021gaussian}, ``the disadvantage
is that the expressions it yields are more unwieldy: they are computer evaluable, so usable in
implementations, but do not admit simple closed form." 
At best the expressions result in an algorithm that implements a method for keeping track (account for) spent privacy budget, called a differential privacy accountant.
This leads in \citep{zhu2021optimal} to a differential privacy accountant (using a complex characteristic function based on taking the Fourier transform) for a client to understand when to stop  helping the server to learn a global model. 

\section{Improved Moment Accountant Method} \label{sec:Main}

\citep{abadi2016deep} proves the following main result (rephrased using our notation by substituting $q=s/N$ in their work):
There exist constants $c_1$ and $c_2$ so that given a constant sample size sequence $s_i=s$
and  number of rounds $T$, for any $\epsilon < c_1 T (s/N)^2$, Algorithm \ref{alg:DPs} is $(\epsilon,\delta)$-DP for any $\delta>0$ if we choose
$$ \sigma \geq c_2 \frac{(s/N)\cdot \sqrt{T \ln (1/\delta)}}{\epsilon}.
$$
The interpretation of  this result is subtle: 
The condition on $\epsilon$  is equivalent to 
\begin{equation} 1/\sqrt{c_1} < z \mbox{ where } z=(s/N) \cdot \sqrt{T/\epsilon}.
\label{eq:originaleps}
\end{equation}
Substituting this into the bound for $\sigma$ yields
\begin{equation} \sigma \geq (c_2\cdot z) \cdot \sqrt{\frac{\ln (1/\delta)}{\epsilon}}.\label{interAbadi1}
\end{equation}
This formulation only depends on $T$ through the definition of $z$. Notice that $z$ may be as small as $1/\sqrt{c_1}$.
Therefore, $\sigma$ can potentially be as small as
\begin{equation} \sigma \geq \frac{c_2}{\sqrt{c_1}} \cdot \sqrt{\frac{\ln (1/\delta)}{\epsilon}}.
\label{eq:c12}
\end{equation}

This leads to the following questions: 
[Q1] Can we refine the theory of \citep{abadi2016deep} and compute an explicit constant $c_2/\sqrt{c_1}$ and show that (\ref{eq:c12}) yields $(\epsilon,\delta)$-DP for $\epsilon$ satisfying some constraint based on $T$, $s$, and $N$ (but without unknown constants)?
[Q2] Can we show that the refinement is  close to tight, implying that the refined analysis of the moment accountant method cannot be much improved?  [Q3] And once we have found such a refinement, how can we use this in practice? 

The next subsections provide affirmative answers to these questions. We stress that our characterization of a universal constant $c_2/\sqrt{c_1}$ which is close to tight is non-trivial as we need to develop new refined expressions which allow us to redefine the unknown constants $c_1$ and $c_2$ in the theory of \citep{abadi2016deep} as functions of $T$ and other parameters in order for us to determine a universal constant that tightly bounds $c_2/\sqrt{c_1}$.



\subsection{Main Contribution: Refined Analysis}

The next theorem answers question [Q1] in the affirmative (for explicit constant $\approx \sqrt{2}$).
Rather than applying the main result of \citep{abadi2016deep}, we can directly use the moment accountant method of their proof to analyse specific parameter settings. 
In Appendix \ref{appDP} we non-trivially improve the analysis of the moment accountant method and show that `constants' $c_1$ and $c_2$ can be chosen as functions of $T$ and other parameters and as a result we show 
that $\sigma$ can remain small up to a lower bound that only depends on the privacy budget, see (\ref{eq:sed}), (\ref{maineq}). 
%
%
The proof of the next theorem 
is detailed in Appendix \ref{appDP} 
where our
improved analysis leads to a first generally applicable Theorem \ref{thm1}.
As a consequence we derive a simplified characterization in the form of Theorem \ref{thm:simpleF}. Finally, we introduce more coarse bounds in order to extract the more readable and more interpretable Theorem \ref{thm:Main} below. We notice that the simulations in  Section \ref{sec:experiment} 
are 
based on parameters that satisfy constraints (\ref{treq1}, \ref{treq2}, \ref{treq3}, \ref{sigeps}) of Theorem \ref{thm:simpleF} as this leads to slightly better results (and also allows any $\delta>0$ and $\epsilon>0$, in particular, $\epsilon\geq 0.5$).

\begin{theorem} \label{thm:Main}
Let $\sigma$ and $(\epsilon,\delta)$ satisfy the relation
\begin{equation}
    \sigma = \sqrt{2(\epsilon+\ln(1/\delta))/\epsilon} \mbox{ with } \delta \leq 1/N \mbox{ and } \epsilon< 0.5 \label{maineq}
\end{equation}

For sample size sequence $\{s_i\}_{i=0}^{T-1}$
the total number of local SGD iterations is equal to $K=\sum_{i=0}^{T-1} s_i$. We define $k=K/N$ as the total number of  local SGD iterations measured in epochs (of size $N$).
Related to the sample size sequence we define
 the mean $\bar{s}$ and maximum $s_{max}$ and their quotient $\theta=s_{max}/\bar{s}$, where 
 \begin{eqnarray*}
&&  \bar{s} = \frac{1}{T} \sum_{i=0}^{T-1} s_i = \frac{K}{T}, 
 \ \
  s_{max} = \max \{ s_0, \ldots, s_{T-1}\}. 
   \end{eqnarray*}

Let $\gamma$ be the smallest solution satisfying
\begin{eqnarray*}
  \gamma \geq   && \hspace{-5mm} \frac{2}{1-\bar{\alpha}} + \\
 && \hspace{-5mm}
 \frac{2^4 \cdot \bar{\alpha}  }{1-\bar{\alpha}}
 \left( \frac{\sigma}{(1-\sqrt{\bar{\alpha}})^2} +\frac{1}{\sigma(1-\bar{\alpha})-2e\sqrt{\bar{\alpha}}}\frac{e^3}{\sigma}
 \right)e^{3/\sigma^2} \\
 && \hspace{-10mm}
 \mbox{ with }
 \bar{\alpha}= \frac{\epsilon }{\gamma k}.
 \nonumber
 \end{eqnarray*}
Parameter $\gamma=2+O(\bar{\alpha})$, which is close to $2$ for small $\bar{\alpha}$. We assume data sets of size $N\geq 10000$ and sample size sequences with $\theta\leq 6.85$. If 
\begin{eqnarray}
(2/e)^2
\cdot k^2 &\geq& 1/2 + \ln(1/\delta) \
\mbox{ and } \label{minK} \\
T &\geq&  \frac{ \gamma \theta^2 }{\epsilon} \cdot k^2, \label{minT}
\end{eqnarray}
then Algorithm \ref{alg:DPs} is $(\epsilon, \delta)$-differentially private.\footnote{Our theory holds in the more general asynchronous SGD framework  discussed in Appendix \ref{sec:asynMNDPSGD}.}
\end{theorem}

As presented in Section \ref{sec:intro_contribution}, current DP concepts are limited to privacy accounting, wherein the calculation of privacy depletion is contingent on the number of rounds $T$. In contrast, our DP theorem establishes relationships between experimental setup parameters such as $N,s_i,k,K,\theta,\gamma$, privacy budget defined by $\delta$ and $\epsilon$, utility goal, i.e., the test accuracy which directly depends on $\sigma$ and $C$, and implementation goal given by $T$. Notably, Theorem \ref{thm:Main} serves as the fundamental basis for developing proactive DP.

For completeness we mention that the more general Theorem \ref{thm:simpleF} states that Algorithm \ref{alg:DPs} is $(\epsilon, \delta)$-differentially private if $\sigma$ and $(\epsilon,\delta)$ satisfy 
$\sigma = \sqrt{2(\epsilon+\ln(1/\delta))/\epsilon}$ together with condition (\ref{minT}) 
(which is $\equiv$ (\ref{treq3})), 
and the less intuitive constraints $T\geq \max \{e\theta \sigma, \theta/h(\sigma) \} \cdot k$ 
(which is\footnote{After substituting (\ref{treq1}) with (\ref{eq:sed}) and $\bar{s}=kN/T$, 
multiplying both sides by $\theta T/N$, and dividing both sides by the min expression in (\ref{treq1}).} $\equiv$ (\ref{treq1})) and $\epsilon\leq \gamma h(\sigma)\cdot k$ (which is $\equiv$
(\ref{treq2})), 
where $h(x)=(\sqrt{1+(e/x)^2}-e/x)^2$. This allows larger $\epsilon\geq 0.5$.


\subsection{Main Contribution: Tightness}
\label{sec:maintight}

If all local data sets are iid
coming from the same source distribution\footnote{This is the case in big data analysis where each local data set represents a too small sample of the source distribution for learning  an accurate local model on its own. Hence, collaboration among multiple clients through a central server is needed to generate an accurate joint global model.}, then 
simulations in Section \ref{sec:experiment} show that for fixed $K=k N$ the best accuracy of the final global model  is achieved by choosing the largest possible mini-batch size $s$ or, equivalently, 
since $K=sT$, 
choosing the smallest possible number of rounds $T=\gamma \theta^2 k^2/\epsilon$ according to condition (\ref{minT}). 
Optimizing accuracy by choosing the smallest 
$T$ 
can be understood by observing that 
this implies that the least number of times noise is added and aggregated into the global model at the server (also larger mini-batches imply less noise relative to the size of the mini-batches).
As a secondary objective, a smaller number of rounds means less round communication.

For the above reasons we want to make $T$ as small as possible, and we are interested in $T$ meeting (\ref{minT}) with equality. Rephrasing [Q2], can Theorem \ref{thm:Main} be strengthened in that the same DP guarantee can hold for a smaller $T$ that violates (\ref{minT})? Appendix 
\ref{app:GDP} uses the $f$-DP framework\footnote{Algorithm \ref{alg:DPs} uses $C$ rather than $2C$ in line 11.
This still fits the $f$-DP framework because our analysis based on \citep{abadi2016deep} assumes a probabilistic (rather than a deterministic) sampling strategy as implemented in the Opacus library \citep{Xpacus}.
For a constant sample size sequence with sample sizes $s$, we can reinterpret the $s_i$ as the actual chosen probabilistic sample sizes with $\mathbb{E}[s_i]=s$ and apply our theory that holds for varying sample size sequences (we need to formulate an upper bound on $s_{max}$ which holds with probability 'close to 1' and this will determine $\theta$ in Theorem \ref{thm:Main} and Theorem \ref{thm:simpleF}).
} to prove into what extent Theorem \ref{thm:Main} is tight:

\begin{theorem} \label{thm:tight}
For $T=(\gamma \theta^2 k^2/\epsilon)/a$ with constant $a>2\gamma$, there exists a parameter setting 
that fits all conditions of Theorem \ref{thm:Main} except for condition (\ref{minT}) such that $(\epsilon,\delta)$-DP is violated.
\end{theorem}

The theorem shows that (\ref{minT}) in Theorem \ref{thm:Main} cannot be relaxed by dividing the right hand side of the inequality by a factor more than $2\gamma\approx 4$.
This is the first such type of tightness result for the moment accountant as introduced by \citep{abadi2016deep}.


\subsection{Main Contribution: A brief implementation flow based on Utility Graph and DP-calculator}

We outline a concise implementation process for proactive DP. The procedure for computing experimental setup parameters is referred to as DP calculator, and is exhaustively elaborated in  Appendix~\ref{app:DPaccSim} and Appendix ~\ref{appendix:DPcalculator}.

The most important mission in machine learning is achieving a good accuracy, therefore,  the added Gaussian noise cannot be too large and is constrained.
For this reason each client wants to  choose (i) the smallest possible clipping constant $C$ for the clipping operation used in DP-SGD such that SGD still bootstraps convergence,  and given $C$, set (ii) the standard deviation $\sigma$ of the added Gaussian noise for differential privacy to a maximum value beyond which we cannot expect to achieve sufficiently good (test) accuracy for the learning task at hand, and given  $C$ and $\sigma$ (leading to Gaussian noise ${\cal N}(0,C^2 \sigma^2\textbf{I})$), estimate (iii) the total number of gradient computations $K=kN$ needed to achieve (converge to at least) the target test accuracy.

Assuming that we are able to efficiently determine a suitable triple $(C,\sigma, k)$ (we explain how this can be done at the end of this section), we are able to apply Theorem \ref{thm:Main} for a constant sample size sequence with $s_i=s$ as follows\footnote{A more advanced algorithm is proposed in Appendix \ref{sec:practice} in which parameters are regularly updated. It is left as an open problem to implement this algorithm for learning tasks based on more complex data sets -- our current experiments do not need such adaptivity. }: 
\begin{itemize}
    \item We set $\delta=1/N$ (the typical value used in literature). Given $\delta$ and the previously determined $\sigma$ we use the equation in (\ref{maineq}) to solve for $\epsilon$. If $\epsilon\geq 0.5$, then Theorem \ref{thm:Main} is not applicable\footnote{If we use the more advanced algorithm of Appendix \ref{sec:practice}, which is based on the more general Theorem \ref{thm:simpleF}, then the constraint $\epsilon<0.5$ can be discarded and larger target $\epsilon$ are possible.} and $\sigma$ must be chosen larger which violates the target accuracy -- one may decide to lower the target accuracy and recompute a triple $(C,\sigma, k)$. If $\epsilon<0.5$, then condition (\ref{maineq}) is satisfied.\footnote{The client may have a DP guarantee for some target epsilon $\epsilon_{tar}$ in mind. If the computed $\epsilon>\epsilon_{tar}$, then the client should not participate in the collaborative training of the global model and should abort.} 

    Since Theorem \ref{thm:Main} is tight up to a "constant", see Theorem \ref{thm:tight}, the equation in (\ref{maineq}) solves for a close to tight $\epsilon$ that cannot be further decreased.
    This elucidates the importance of a Theorem \ref{thm:Main} that provides a close to {\em tight}  bound.
    
    \item Given $k$, $\sigma$ and $\epsilon$, we compute $\gamma$ (which is generally close to $2$). Notice that for a constant sample size sequence we have $\theta=1$ and we can verify condition (\ref{minK}). Since $k$  generally represents 50 or 100s of epochs and we only take the natural logarithm of $1/\delta=N$, condition (\ref{minK}) generally verifies in practice.
    \item This leaves us with the final condition (\ref{minT}) which needs to be satisfied in order to apply Theorem \ref{thm:Main} and conclude $(\epsilon,\delta)$-DP. As discussed in Section \ref{sec:maintight}, we want to meet condition (\ref{minT}) with equality as this yields the best test accuracy and minimal round complexity (among the $T$ that satisfy (\ref{minT}) for the previously fixed parameters $\gamma$, $\theta$ and $k$). As soon as $T$ is computed, we set $s=K/T$.
\end{itemize}

We now return  to the problem of how to efficiently learn how to select parameters $C$, $\sigma$ and $K$ (without having to conduct a full training, based on DP-SGD,  for various candidate parameter settings for empirical evidence).  We introduce  the concept of a utility graph where a ``best-case'' accuracy is depicted as a function of noise $\sigma$ and clipping constant $C$ in DP-SGD (see Section
\ref{sec:experiment}). 
In DP-SGD the last round of local updates is aggregated into an update of the global model, after which the global model is finalized. This means that the Gaussian noise added to a client local update of its last round is directly added as a perturbation to the final global model.
We have a best-case scenario if we neglect the added noise of all previous rounds. That is, the ``best-case'' accuracy for DP-SGD is the accuracy of a global model which is trained using SGD with clipping corresponding to $C$ and without adding Gaussian noise, after which Gaussian noise is added to the final model at the very end. The utility graph for a fixed $C$ depicts this ``best-case'' trade-off between test accuracy and $\sigma$.

To generate the graph, we fix a diminishing learning rate $\bar{\eta}_t$ (step size) from round to round and we fix the total number $K$ of local gradient computations that will be performed. Based on local training data and a-priori knowledge (possibly from transferring a public model of another similar learning task), a local client can run SGD locally without any added noise but with clipping corresponding to $C$. This learns a local model $w^*$ (which depends on $C$) and we compare how much accuracy is sacrificed by adding Gaussian noise $n\sim {\cal N}(0,C^2\sigma^2{\bf I})$; that is, we compute and depict the ratio ``$F(w^*+n)/F(w^*)$" as a function of $\sigma$ and we do this for various clipping constants $C$. This teaches us the range of $\sigma$ and $C$ combinations that may lead to sufficient accuracy (say at most a 10\% drop).\footnote{In some cases (not in the experiments in this paper), $F(w^*)$ can be very small and as a result $F(w^*+n)$ is not stable. A solution for this case can be to train  model $w^*$ to reach a sufficient good accuracy (for example 80\% or 90\%) and then stop. Now $F(w^*)$ is not so small and $F(w^*+n)/F(w^*)$ may still produce stable results. We hypothesize that the resulting utility graph transfers to a full training of model $w^*$.}

The rationale for employing the utility graph in the context of a single client is primarily derived from scenarios characterized by strong convexity, where  the optimal solution $w^*$, when perturbed by noise, is expected to remain in close proximity to $w^*$. Therefore, in the strong convex case we expect $\hat{w}$ (as a result of DP-SGD) to be close enough to $w^*$ to lead to sufficient prediction accuracy as anticipated by the utility graph. In the realm of nonconvex optimization it has been established that deep and expansive neural networks possess a multitude of robust global optima $w^*$, 
which are interconnected, as evidenced by research such as~\cite{nguyen2019connected}. Furthermore, research by~\cite{nguyen2020global}  suggests that deep learning models exhibit linear convergence when subjected to SGD. This phenomenon underpins the practical observation that running SGD on a given training dataset consistently yields global optima that perform with high accuracy on testing datasets. Such findings reinforce the expectation that the distribution of $\hat{w}$ (as a result of DP-SGD) will be sufficiently concentrated around the true optimum $w^*$ and lead to sufficient prediction accuracy as anticipated by the utility graph.


The main purpose of the utility graph is to efficiently identify ineffective pairs $(\sigma,C)$  without conducting a true private training. It is not guaranteed that the chosen pairs $(\sigma,C)$ will offer good test accuracy when running  the true private training DP-SGD, i.e., whether the test accuracy fits our requirement. Only after running DP-SGD  the true test accuracy is found out.

We summarize the execution flow for proactive DP. We begin by setting $\delta = \frac{1}{N}$ and using the utility graph to search for a suitable $\sigma$ by starting with a large value of $\sigma$ and gradually decreasing it. The goal is to find a value of $\sigma$ that yields an expected prediction accuracy greater than our desired accuracy or utility goal. After that, we make use of the equation in (\ref{maineq}) to compute $\epsilon$ and check if it is smaller than $0.5$. To work with the utility graph, we must determine the anticipated decrease in prediction accuracy compared to training without privacy preservation. If $\epsilon$ is greater than $0.5$, we must select another $\sigma$ by decreasing $\sigma$ until we are unable to obtain an appropriate value because our result is tight. Due to the tightness and simplicity of the equation in (\ref{maineq}), the computational cost of this process is small. If $\epsilon$ is less than $0.5$, we move forward in the process. Once we have obtained the optimal $\sigma$ and $\epsilon$, the implementation process progresses. In practical terms, $k$ is generally restricted to 50 or 100 epochs. After determining the desired values of $k$ and $\epsilon$, we use the DP calculator to find $\gamma$ and the best $T$. Finally, we can set $s=K/T$, where $K=kN$.

As an example of the above method, in Section \ref{sec:experiment} simulations for the LIBSVM data set 
show $(\epsilon=0.05,\delta=1/N)$-DP is possible  while achieving good accuracy with $\sigma\approx 20$.
\section{Experiments}
\label{sec:experiment}

\begin{figure*}[ht!]
\vspace{-.32cm}
  \centering
  \subfigure[]{\includegraphics[width=0.34\textwidth]{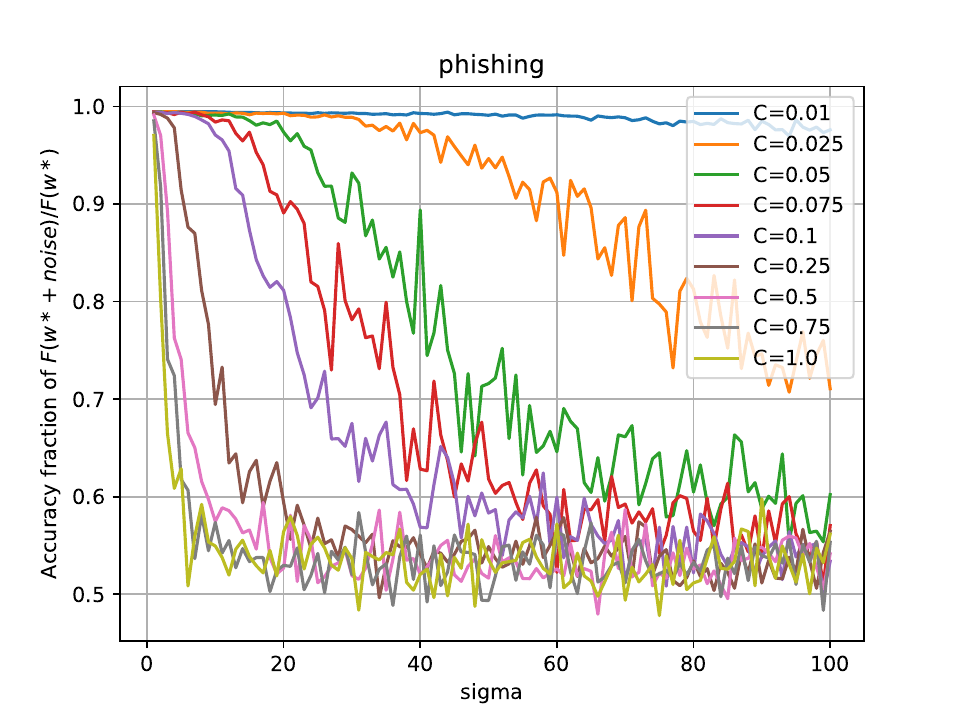}\label{fig:utility_phishing_main_0}}
  \hfill
  \subfigure[]{\includegraphics[width=0.32\textwidth]{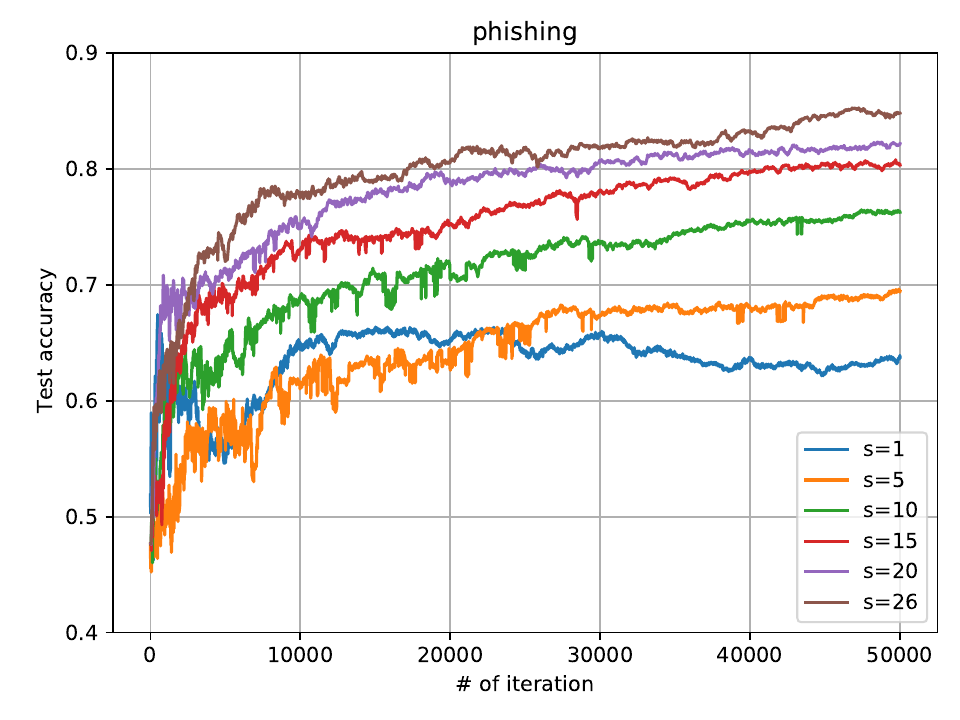}\label{fig:asyncDP_phishing_sample_main_0}}
  \hfill
  \subfigure[]{\includegraphics[width=0.32\textwidth]{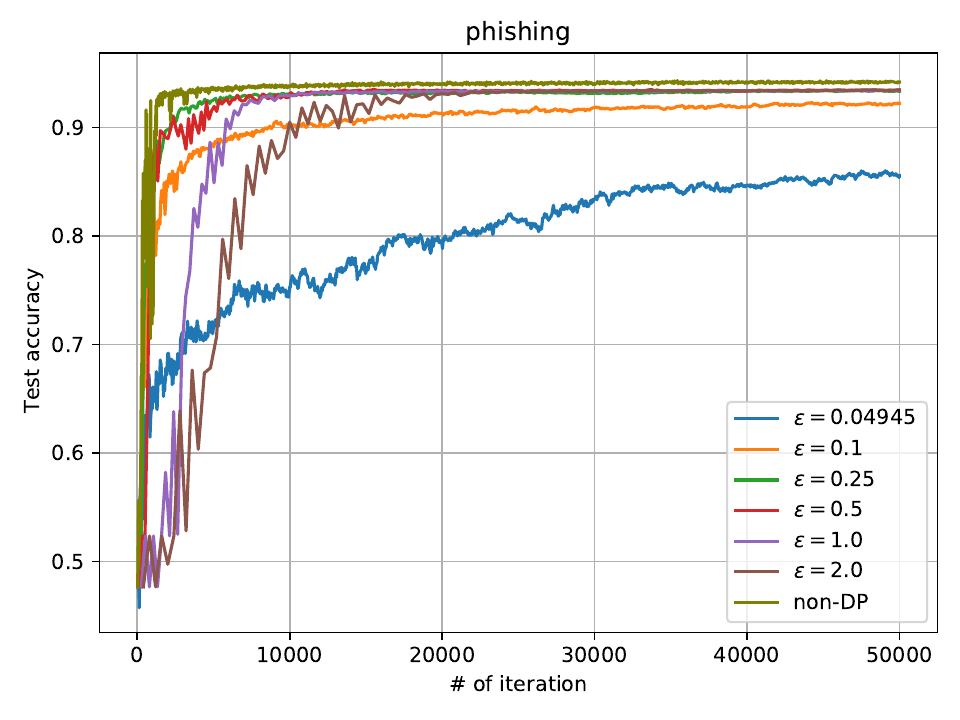}\label{fig:asyncDP_phishing_epsilon_main_0}}  
  \caption{Strongly convex. (a) Utility graph, (b) Different $s$, (c) Different $\epsilon$}
  \label{fig:exp_stronglyconvex}
\end{figure*}

\begin{figure*}[ht!]
\vspace{-.35cm}
  \centering
  \subfigure[]{\includegraphics[width=0.34\textwidth]{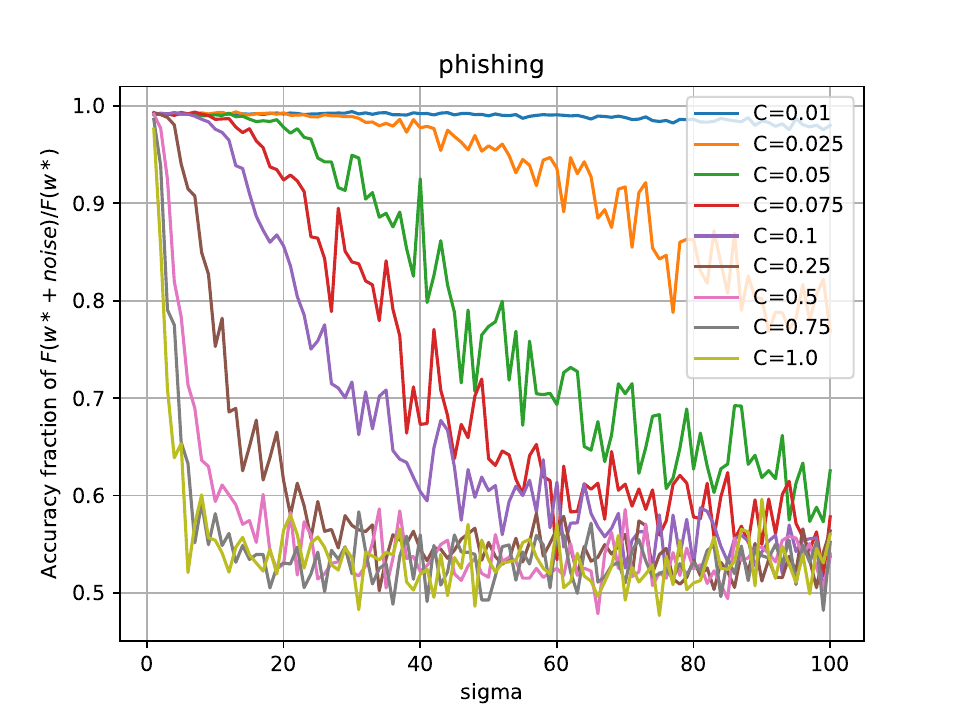}\label{fig:utility_phishing_main_1}}
  \hfill
  \subfigure[]{\includegraphics[width=0.32\textwidth]{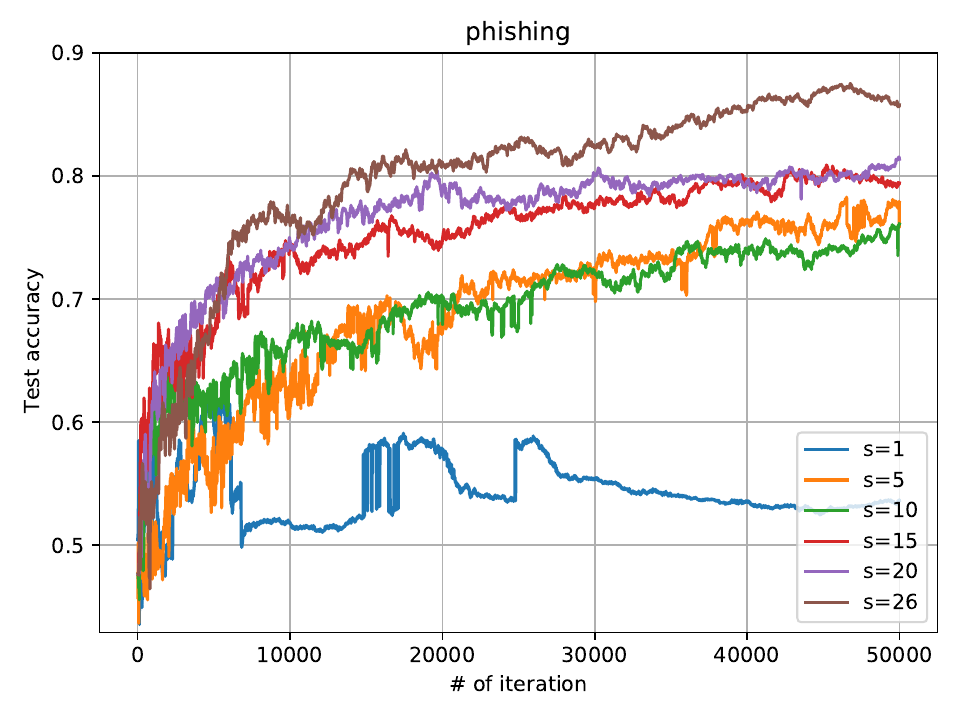}\label{fig:asyncDP_phishing_sample_main_1}}
  \hfill
  \subfigure[]{\includegraphics[width=0.32\textwidth]{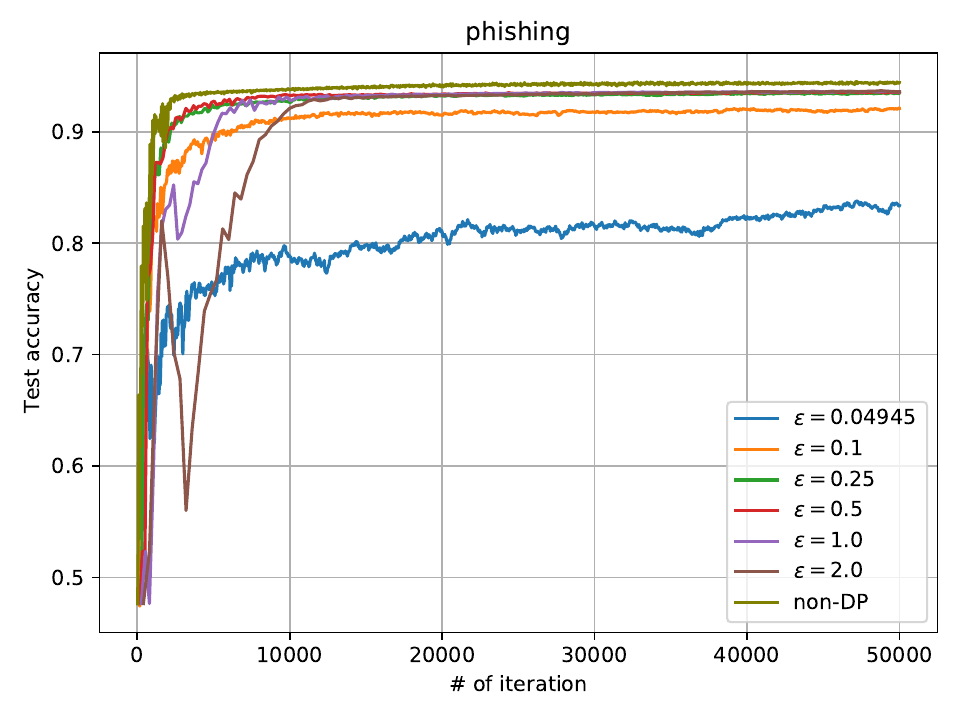}\label{fig:asyncDP_phishing_epsilon_main_1}}
  \caption{Plain convex. (a) Utility graph, (b) Different $s$, (c) Different $\epsilon$}
  \label{fig:exp_plainconvex}
\end{figure*}

\begin{figure*}[ht!]
\vspace{-.35cm}
  \centering
  \subfigure[]{\includegraphics[width=0.34\textwidth]{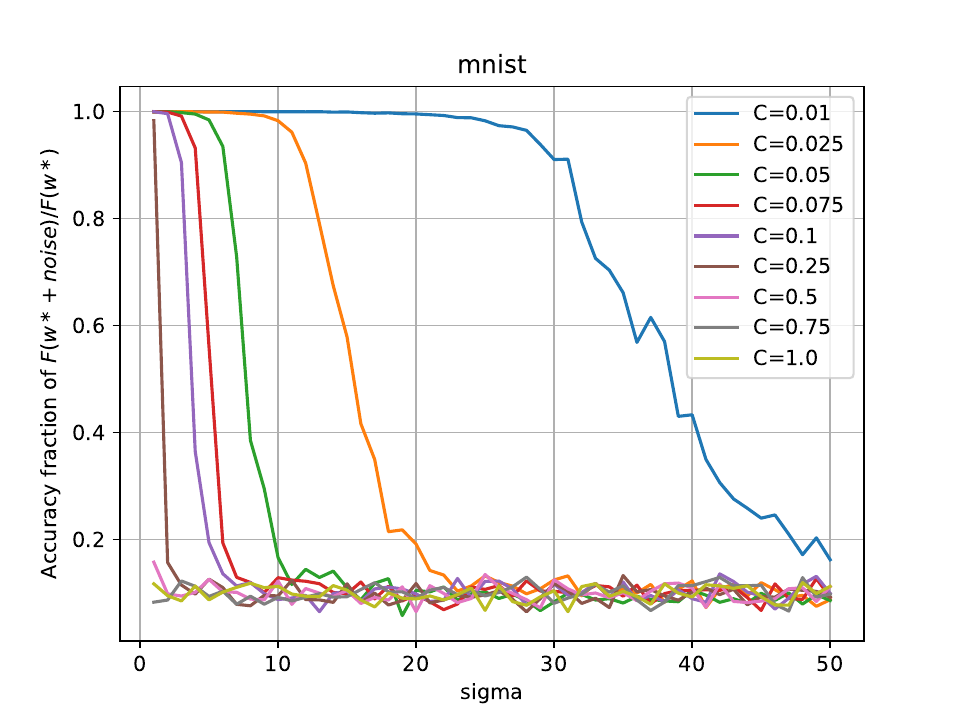}\label{fig:utility_mnist_main_1}}
  \hfill
  \subfigure[]{\includegraphics[width=0.32\textwidth]{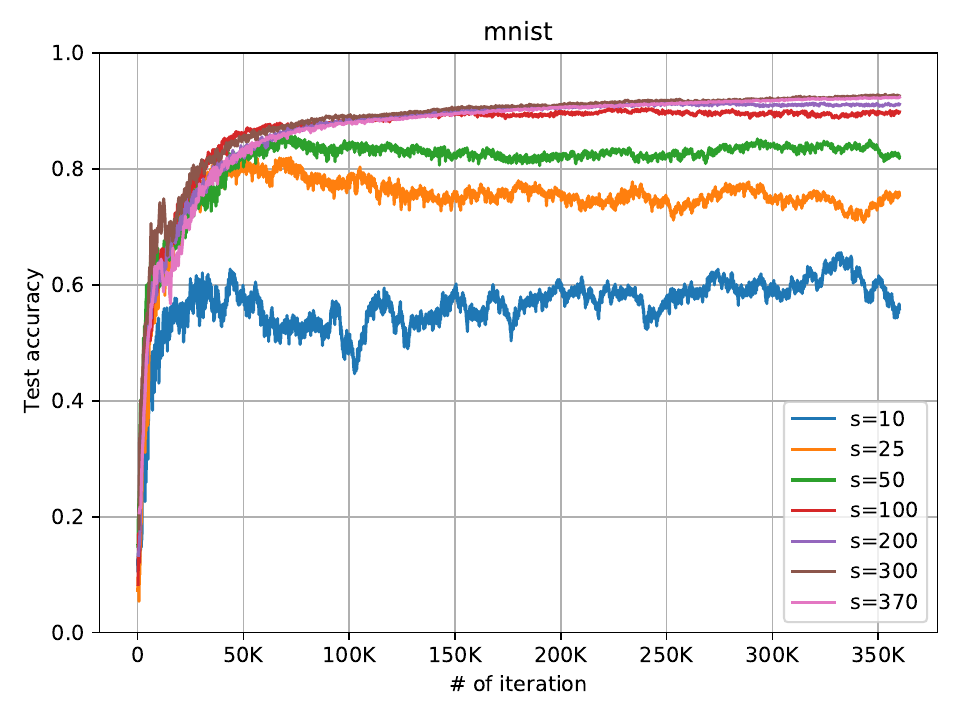}\label{fig:asyncDP_mnist_sample_main_1}}
  \hfill
  \subfigure[]{\includegraphics[width=0.32\textwidth]{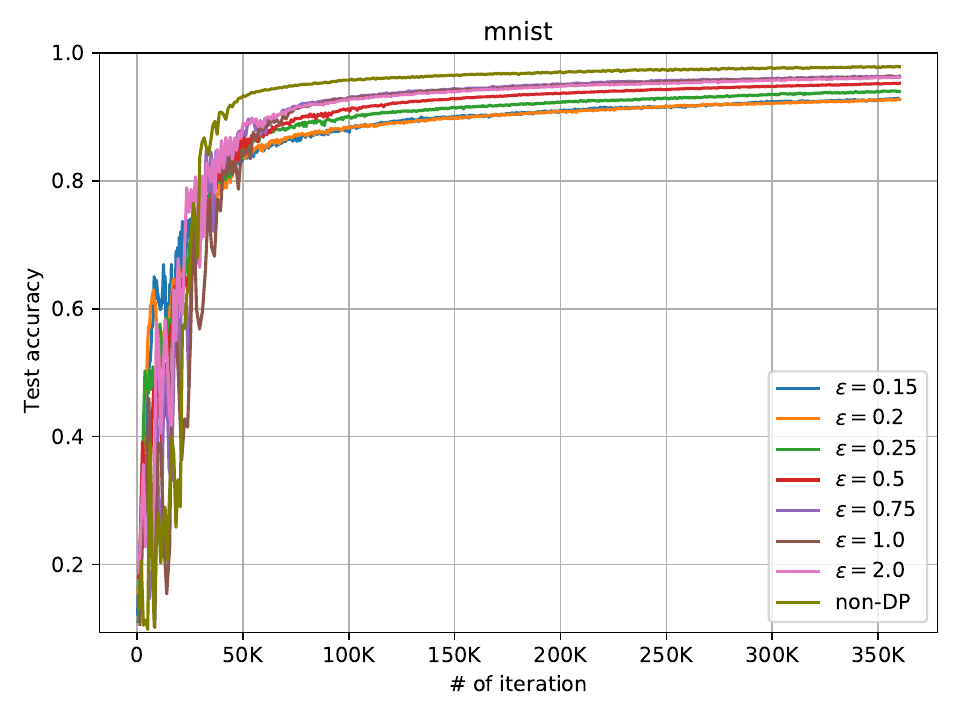}\label{fig:asyncDP_mnist_epsilon_main_1}}
  \caption{Non-convex. (a) Utility graph, (b) Different $s$, (c) Different $\epsilon$}
  \label{fig:exp_non_convex}
\end{figure*}


Our goal is to show that the more general asynchronous differential privacy framework (asynchronous DP-SGD which includes DP-SGD of Algorithm \ref{alg:DPs}) of Appendix \ref{sec:asynMNDPSGD} ensures a strong privacy guarantee, i.e, can work with very small  $\epsilon$ (and $\delta=1/N$), while having a good convergence rate to good accuracy. We refer to Appendix \ref{app:experiment} for simulation details and complete parameter settings.


\textbf{Simulation environment.} For simulating the asynchronous DP-SGD framework, we use multiple threads where each thread represents one compute node joining the training process. The experiments are conducted on Linux-64bit OS, with $16$ cpu processors, and 32Gb RAM.

\noindent
\textbf{Objective function.} We summarize experimental results of our asynchronous DP-SGD framework for strongly convex, plain convex and non-convex objective functions with \textit{constant sample size sequences}. As the plain convex objective function we use logistic regression: The weight vector $w$ and bias value $b$ of the logistic function can be learned by minimizing the log-likelihood function $J$:
\begin{equation} 
\nonumber
    J = - \sum_{i=1}^N [ y_{i} \cdot \log (\bar{\sigma}_i) 
    + (1 - y_{i}) \cdot \log 
    (1 - \bar{\sigma}_i) 
    ],
\end{equation}
where $N$ is the number of training samples $(x_i,y_i)$ with $y_i\in\{0,1\}$, and $\bar{\sigma}_i
= 1/(1 + e^{-(w^{\mathrm{T}}x_i + b)})$
is the sigmoid function. 
The goal is to learn a vector/model $w^*$ which represents a pair $\bar{w}=(w,b)$ that minimizes $J$.
Function $J$  changes into a strongly convex problem by adding ridge regularization with a regularization parameter $\lambda>0$, i.e., we minimize $\hat{J}=J+ \frac{\lambda}{2} \norm{\bar{w}}^2$ instead of $J$. That is,
\begin{equation}
    \hat{J} = - \sum_{i=1}^N [ y_{i}\log (\bar{\sigma}_i) 
    + (1 - y_{i})\log 
    (1 - \bar{\sigma}_i) 
    ] + \frac{\lambda}{2} \norm{\bar{w}}^2.
\end{equation}
For simulating non-convex problems, we choose a simple
neural network (LeNet)~\citep{lecun1998gradient} with cross entropy loss function for image classification.

\textbf{Parameter selection.} The parameters used for our distributed algorithm with Gaussian based differential privacy for strongly convex, plain convex and non-convex objective functions are described in Table~\ref{tbl:tbl_async_DP_paramter}. The clipping constant $C$ is set to $0.1$ for strongly convex and plain convex problems and $0.025$ for non-convex problem (this turns out to provide good utility).

\begin{table*}[!ht]
\caption{Common parameters of asynchronous DP-SGD framework with differential privacy}
\label{tbl:tbl_async_DP_paramter}
\begin{center}
\begin{scriptsize}
\scalebox{1.3}{
\begin{threeparttable}

\begin{tabular}{|l|c|c|c|c|c|c|}
\hline
                & \# of clients $n$ & Diminishing step size $\bar{\eta_t}$ & Regular $\lambda$  & Clipping constant $C$  
               \\ 
               \hline
               \hline
Strongly convex \ \ \ \ \ \ & 5 & $\frac{\eta_0}{1 + \beta {t}} \tnote{\ddag}$ & $\frac{1}{N}$ & 0.1                   
\\ 
\hline
Plain convex & 5 & $\frac{\eta_0}{1 + \beta {t}}$ or $\frac{\eta_0}{1 + \beta \sqrt{t}}$ & $N/A$ & 0.1                                   
\\ 
\hline
Non-convex & 5 & $\frac{\eta_0}{1 + \beta \sqrt{t}}$ & $N/A$ & 0.025
\\ 
\hline
\end{tabular}
    \begin{tablenotes}
       \item {\scriptsize $\ddag$ The $i$-th round step size $\bar{\eta}_i$ is computed by substituting  $t=\sum_{j=0}^{i-1} s_j$ into the diminishing step size formula}.
   \end{tablenotes}
   
\end{threeparttable}
}
\end{scriptsize}
\end{center}
\vskip -0.1in
\end{table*}

For the plain convex case, we can use  diminishing step size schemes $\frac{\eta_0}{1 + \beta \cdot t}$ or $\frac{\eta_0}{1 + \beta \cdot \sqrt{t}}$. In this paper, we focus our experiments for the plain convex case on $\frac{\eta_0}{1 + \beta \cdot \sqrt{t}}$. Here, $\eta_0$ is the initial step size and we perform a systematic grid search on parameter $\beta=0.001$ for strongly convex case and $\beta=0.01$ for both plain convex and non-convex cases.
Moreover, most of the experiments are conducted with $5$ compute nodes and $1$ central server.
When we talk about accuracy (from Figure~\ref{fig:asyncDP_phishing_sampling_method} and onward), we mean test accuracy defined as  the fraction of samples from a test data set that get accurately labeled by the classifier (as a result of training on a training data set by minimizing a corresponding objective function).

\noindent
\textbf{Asynchronous DP-SGD setting.} The experiments are conducted with $5$ compute nodes and $1$ central server. For simplicity, the compute nodes have  iid data sets.
%
%



\noindent
\textbf{Data sets.} All our experiments are conducted on LIBSVM~\cite{CC01a}\footnote{\scriptsize https://www.csie.ntu.edu.tw/~cjlin/libsvmtools/datasets/binary.html} and MNIST~\cite{lecun-mnisthandwrittendigit-2010}\footnote{http://yann.lecun.com/exdb/mnist/} data sets.

\subsection{Utility Graph}
\label{section:utility_graph}
Since we do not have a closed form to describe the relation between the utility of the model (i.e., prediction accuracy) and $\sigma$, 
we propose a
heuristic approach to learn the 
range of $\sigma$ from which we  may select $\sigma$ for finding the best $(\epsilon,\delta)$-DP.  
The utility graphs -- Figures~\ref{fig:utility_phishing_main_0}, ~\ref{fig:utility_phishing_main_1} and ~\ref{fig:utility_mnist_main_1}  -- show the fraction of test accuracy between the model ``$F(w + n)$" over the original model ``$F(w)$" (without noise), where $n\sim  {\cal N}(0,C^2 \sigma^2\textbf{I})$ (per round) for various values of the clipping constant $C$ and noise standard deviation $\sigma$. Intuitively, the closer $F(w + n)/F(w)$ to 1, the better accuracy w.r.t. to $F(w)$. Note that $w$ can be any solution and in the utility graphs, we choose $w=w^*$ with $w^*$ being  near to an optimal solution.  

The smaller $C$, the larger $\sigma$ can be, hence, $\epsilon$ can be smaller which gives stronger privacy. However, the smaller $C$, the more iterations (larger $K$) are needed for convergence. And if selected too small, no fast enough convergence is possible.
%
In next experiments we use clipping constant $C=0.1$, which gives a drop of at most $10\%$ in test accuracy 
for $\sigma \leq 20$ for both strongly convex and plain convex objective functions. 
To keep the test accuracy loss 
$\leq$ 10\% in the non-convex case, we choose 
$C=0.025$ and 
$\sigma \leq 12$.


\subsection{DP-SGD with Different Constant Sample Sizes}

Figure~\ref{fig:asyncDP_phishing_sample_main_0} and Figure~\ref{fig:asyncDP_phishing_sample_main_1} illustrate the test accuracy of our asynchronous DP-SGD with various constant sample sizes for the strongly convex and plain convex cases. Here, we use privacy budget $\epsilon=0.04945$ and noise $\sigma=19.2$. 
When we use a bigger constant sample size $s$, for example, $s=26$, our algorithm can achieve the desired performance, when compared to other constant sample sizes.\footnote{$s=26$ meets the lower bound 
on $T$; a larger $s$ violates this lower bound. The reason for having a somewhat small maximum possible $s$ is because of the relatively small data set size.}  The experiment is extended to the non-convex 
case
as shown in Figure~\ref{fig:asyncDP_mnist_sample_main_1}, where we can see a similar pattern.
Experimental results for other data sets 
are in Appendix \ref{app:experiment}. This confirms that our DP-SGD framework can converge to a decent accuracy while achieving 
a very small privacy budget $\epsilon$.

\subsection{DP-SGD with Different Levels of Privacy Budget}


Figure~\ref{fig:asyncDP_phishing_epsilon_main_0} and Figure~\ref{fig:asyncDP_phishing_epsilon_main_1} show that
our DP-SGD framework converges to better accuracy 
if $\epsilon$ is slightly larger. 
%
%
%
E.g., in the strongly convex case, 
privacy budget $\epsilon=0.04945$ achieves test accuracy 86\% compared to 93\% without differential privacy (hence, no added noise); $\epsilon=0.1$, still significantly smaller than what is reported in literature, achieves test accuracy 91\%. Figure~\ref{fig:asyncDP_mnist_epsilon_main_1} shows the test accuracy of our asynchronous DP-SGD 
for different privacy budgets $\epsilon$ in the non-convex case. For $\epsilon=0.15$, our framework can achieve a test accuracy of about $93\%$, compared to $98\%$ without differential privacy. These figures again confirm the effectiveness of our DP-SGD framework, which can obtain a strong differential privacy guarantee.

\section{Related Work}
\label{sec:related_work}


Our main contribution in this paper is an improved and tight analysis of the moment accountant method by \citep{abadi2016deep}. Since our theory goes beyond the theory developed by Abadi et al., we want to compare our work with \citep{abadi2016deep}. Our setup in terms of the model architecture, hyperparameters, etc., is different from Abadi et al.’s setup. However, with $\epsilon = 2$, we achieved $\approx$ $60\%$ test accuracy after $T = 350000$ iterations (Figure~\ref{fig:asyncDP_cifar10_epsilon}) which equals $k=7$ epochs ( $k = K/N = 350000/50000 = 7$). In Figure 6.1 of Abadi’s paper, they also achieved around $60\%$ test accuracy after 7 epochs.

However, if we analyze a different perspective where we are interested in the test accuracy deduction from the non-DP setting, then \citep{abadi2016deep} states that they used the model architecture from the Tensorflow tutorial which has $86\%$ accuracy, and this means they have $26\%$ accuracy deduction at epoch 7. Meanwhile, we used AlexNet which only has a reported $74.74\%$ accuracy for the non-DP setting which gives rise to a smaller $14.14\%$ accuracy deduction at epoch 7. 

For the above reason, we claim that we have a significant improved analysis of the accountant method in \citep{abadi2016deep}. Even with the state-of-the-art method of \cite{de2022unlocking} only achieves $65.9\%$ for CIFAR10 without pre-training data. However, they use WRN-40-4 (WideResnet) which has $98\%$ test accuracy, and this means they have a $32.1\%$ test accuracy deduction. In this case, our method still appears to be better as it can achieve a similar differential privacy and utility trade-of but for much more simpler neural network model. On the other hand, if we do not consider the model architecture, the test accuracy can be achieved to $56.8\%, 65.9\%, 73.5\%$ for $\epsilon = 1,2,4$, respectively as shown \cite{de2022unlocking}. Meanwhile, we achieve $\approx 60\%$ test accuracy for $\epsilon \in [0.5,3]$, hence, with $\epsilon = 1$ our method still yields better test accuracy and also allows us to report $\approx 60\%$ test accuracy for smaller $\epsilon$, i.e., better differential privacy.



\section{Conclusion}
\label{sec:conc}
We propose a new concept in DP coined proactive DP, which serves as a multi-target optimization framework for DP-SGD. The design of proactive DP is based on a significant improvement of the analysis of the moment account method together with two new computation tools - utility graph and DP calculator. These tools help to efficiently identify optimal experimental setups for DP-SGD. We have presented a detailed implementation process for our proactive DP, accompanied by rigorous experiments aimed at showcasing its proof-of-concept.

\section*{Acknowledgements}
This paper is supported by NSF grant CNS-1413996 “MACS: A Modular
Approach to Cloud Security.

\section*{Impact Statement}
This paper presents work whose goal is to advance the field of Machine Learning. There are many potential societal consequences of our work, none which we feel must be specifically highlighted here.
\bibliography{references}
\bibliographystyle{plainnat}

\clearpage
\appendix
\onecolumn


\section{Asynchronous Mini-Batch DP-SGD}
\label{sec:asynMNDPSGD}

All our theory, including the  theorems presented in the main body, holds in the asynchronous SGD framework as introduced in this appendix, where 
we provide a more general
{\em asynchronous mini-batch SGD algorithm} (which follows Hogwild!'s philosophy \citep{Hogwild,DeSaZhangOlukotunEtAl2015,zhang2016hogwild++,nguyen2018sgd,Leblond2018,van2020hogwild}) {\em with DP}.
The asynchronous setting allows clients to adapt their sample sizes to their processing speed and communication latency.

Algorithms\footnote{Our pseudocode uses the format from \citep{van2020hogwild}.} \ref{alg:DP1}, \ref{alg:DP2}, and \ref{alg:DP} explain in pseudo code our asynchronous LDP approach. It is based on the Hogwild!~~\citep{Hogwild}
 recursion 
\begin{equation}
 w_{t+1} = w_t - \eta_t  \nabla f(\hat{w}_t;\xi_t),\label{eqwM2a}
 \end{equation}
 where $\hat{w}_t$ represents the vector used in computing the gradient $\nabla f(\hat{w}_t;\xi_t)$ and whose vector entries have been read (one by one)  from  an aggregate of a mix of  previous updates that led to $w_{j}$, $j\leq t$.
 In a single-thread setting where updates are done in a fully consistent way, i.e. $\hat{w}_t=w_t$, yields SGD with diminishing step sizes $\{\eta_t\}$.

Recursion (\ref{eqwM2a}) models asynchronous SGD. 
The amount of 
 asynchronous behavior that can be tolerated is given by some function $\tau(t)$, see~\citep{nguyen2018sgd} where this is analysed for strongly convex objective functions:
We say that the sequence $\{w_t\}$  is consistent with delay function $\tau$  
if, for all $t$, vector $\hat{w}_t$ includes the aggregate 
of the updates up to and including those made during the $(t-\tau(t))$-th iteration, i.e., $$\hat{w}_t = w_0 - \sum_{j\in {\cal U}} \eta_j \nabla f(\hat{w}_j;\xi_j)$$ for some ${\cal U}$ with $\{0,1,\ldots, t-\tau(t)-1\}\subseteq {\cal U}$. 

In Algorithm \ref{alg:DP} the local SGD iterations all compute gradients based on the same local model $\hat{w}$, which gets substituted by a newer global model $\hat{v}_k$ as soon as it is received by the interrupt service routine \textbf{ISRReceive}. As explained in \textbf{ISRReceive} $\hat{v}_k$ includes all the updates from all the clients up to and including their local rounds $\leq k$. This shows that locally the delay $\tau$ can be estimated based on the current local round $i$ together with $k$. Depending on how much delay can be tolerated Setup defines $\Upsilon(k,i)$ to indicate whether the combination $(k,i)$ is permissible (i.e., the corresponding delay aka asynchronous behavior can be tolerated). It has been shown that for strongly convex objective functions (without DP enhancement) the convergence rate remains optimal even if the delay $\tau(t)$ is as large as $\approx \sqrt{t/\ln t}$~\citep{nguyen2018sgd}. Similar behavior has been reported for plain convex and non-convex objective functions in \citep{van2020hogwild}.

In Algorithm \ref{alg:DP} we assume that messages/packets never drop; they will be resent
but can arrive out of order. This guarantees that we get out of the "while $\Upsilon(k,i)$ is false loop" because at some moment the server receives all the updates in order to broadcast a new global model $\hat{v}_{k+1}$ and once received by \textbf{ISRReceive} this will increment $k$ and make $\Upsilon(k,i)$ true which allows \textbf{LocalSGDwithDP} to exit the wait loop. As soon as the wait loop is exited we know that all local gradient computations occur when $\Upsilon(k,i)$ is true which reflect that these gradient computations correspond to delays that are permissible (in that we still expect convergence of the global model to good accuracy).

\begin{algorithm}[!ht]
\caption{Client -- Local model with Differential Privacy}
\label{alg:DP1}

\begin{algorithmic}[1]
\Procedure{Setup}{$n$}: 
\vspace{1mm} \newline Initialize  sample size sequence $\{s_i\}_{i=0}^T$,  (diminishing) round step sizes $\{\bar{\eta}_i\}_{i=0}^T$, and a default global model ${\hat v}_0$ to start with.
\vspace{1mm} \newline Define a permissible delay function $\Upsilon(k,i)\in \{\textbf{True}, \textbf{False}\}$ which takes the current local round number $i$ and the round number $k$ of the last received global model into account to find out whether local SGD should wait till a more recent global model is received. $\Upsilon(\cdot,\cdot)$ can also make use of knowledge of the sample size sequences used by each of the clients.

\EndProcedure   
\end{algorithmic}
\end{algorithm}



\begin{algorithm}[!ht]
\caption{Client -- Local model with Differential Privacy}
\label{alg:DP2}
\begin{algorithmic}[1]
\Procedure{ISRReceive}{$\hat{v}_k$}: 
\vspace{1mm} \newline  This Interrupt Service Routine is called whenever a new broadcast global model $\hat{v}_k$ is received from the server. 
Once received, {\em the client's local model $\hat{w}$ is replaced with $\hat{v}_k$} (if no more recent global model $\hat{v}_{>k}$ was received  out of order before receiving this $\hat{v}_k$)
\vspace{1mm} \newline The server broadcasts global model $\hat{v}_k$ for global round number $k$ once the updates corresponding to local round numbers $\leq k-1$ 
from {\em all} clients have been received and have been aggregated into the global model. The server aggregates updates from clients into the current global model as soon as they come in.  This means that
$\hat{v}_k$ includes all the updates from all the clients up to and including their local round numbers $\leq k-1$ and potentially includes updates corresponding to later round numbers from subsets of clients. The server broadcasts the global round number $k$ together with $\hat{v}_k$. 
\EndProcedure  

\end{algorithmic}
\end{algorithm}


\begin{algorithm}[!ht]
\caption{Client -- Local model with Differential Privacy}
\label{alg:DP}
\begin{algorithmic}[1]
 \Procedure{LocalSGDwithDP}{$d$}

\State $i=0$, $\hat{w}=\hat{v}_{0}$ 
 
\While{\textbf{True}}
       \State {\bf while}  $\Upsilon(k,i)=\textbf{False}$
            {\bf do} nothing 
        {\bf end} \Comment{$k$ is the global round at the server.}

        \State Uniformly sample a random set $\{\xi_h\}_{h=1}^{s_i}\subseteq d$ 
        
    \State $h=0$, $U=0$ 
    \While{$h< s_{i}$}
        \State $g = [\nabla f(\hat{w}, \xi_h)]_C$ 
        \State ${U} = {U} + g$  
        \State $h$++
    \EndWhile
    
    \State $n\leftarrow {\cal N}(0,C^2 \sigma_i^2\textbf{I})$ 
    \State $U=U+n$
    \State $\hat{w}= \hat{w}+\bar{\eta}_i\cdot U$ 
    
    \State Send $(i, U)$ to the Server. 
    \State $i$++
\EndWhile
   
\EndProcedure

\end{algorithmic}
\end{algorithm}


        
    
    

In this paper we analyse the Gaussian based differential privacy method of \citep{abadi2016deep}. We use their clipping method; rather than using the gradient $\nabla f(\hat{w}, \xi)$ itself, we use its clipped version $[\nabla f(\hat{w}, \xi)]_C$ where $[x]_C= x/\max\{1,\|x\|/C\}$. Also, we use the same mini-batch approach where before the start of the $i$-th local round a random min-batch of sample size $s_i$ is selected. During the inner loop the sum of gradient updates is maintained where each of the gradients correspond to the same local model $\hat{w}$ until it is replaced by a newer global model. In supplementary material \ref{appDP} we show that this is needed for proving DP guarantees and that generalizing the algorithm by locally implementing the Hogwild! recursion itself (which updates the local model each iteration) does not work together with the DP analysis. So, our approach only uses the Hogwild! concept at a global round by round interaction level.

At the end of each local round the sum of updates $U$ is obfuscated with Gaussian noise;  Gaussian noise ${\cal N}(0,C^2\sigma_i^2)$ is added to each vector entry. In this general description $\sigma_i$ is round dependent, but our DP analysis in Supplementary Material \ref{appDP} must from some point onward assume a constant $\sigma=\sigma_i$ over all rounds. The noised $U$ times the round step size $\bar{\eta}_i$ is added to the local model after which a new local round starts again.

The noised $U$ is also transmitted to the server who adds $U$ times the round step size $\bar{\eta}_i$ to its global model $\hat{v}$. As soon as all clients have submitted their updates up to and including their local rounds $\leq k-1$, the global model $\hat{v}$, denoted as $\hat{v}_k$, is broadcast to all clients, who in turn replace their local models with the newly received global model. Notice that $\hat{v}_k$ may include updates from a subset of client that correspond to local rounds $\geq k$.

The presented algorithm adapts to asynchronous behavior in the following two ways: We explained above that the broadcast global models $\hat{v}_k$ themselves include a mix of received updates that correspond to local rounds $\geq k$ -- this is due to asynchronous behavior. Second, the sample size sequence $\{s_i\}$ does not necessarily need to be fixed a-priori during  \textbf{Setup} (the round step size sequence $\{\bar{\eta}_i\}$ does need to be fixed a-priori). In fact, the client can adapt its sample sizes $s_i$ on the fly to match its speed of computation and communication latency. This allows the client to adapt its local mini-batch SGD to its  asynchronous behavior due to the scheduling of its own resources. Our DP analysis holds for a wide range of varying sample size sequences. 

We notice that adapting sample size sequences on a per client basis still fits the same overall objective function as long as all local data sets are iid: This is because  iid implies that the execution of the presented algorithm can be cast in a single Hogwild! recursion where the $\xi_h$ are uniformly chosen from a common data source distribution ${\cal D}$. This corresponds to the stochastic optimization problem 
\begin{align}
\min_{w \in \mathbb{R}^d} \left\{ F(w) = \mathbb{E}_{\xi \sim \mathcal{D}} [ f(w;\xi) ] \right\},  \nonumber 
\end{align}
which defines objective function $F$ (independent of the locally used sample size sequences). Local data sets being iid in the sense that they are all, for example, drawn from car, train, boat, etc images benefit from DP in that car details (such as an identifying number plate), boat details, etc. need to remain private.

\section{Differential privacy proofs}
\label{appDP}


This appendix proves a key observation 
improving the DP moment accountant from \citep{abadi2016deep}. As shown in \citep{abadi2016deep}, for 
{\em any given $T\geq 0$} in one specific setting, there are many choices for $(\epsilon,\delta,\sigma)$ depending on two constants $(c_1,c_2)$ (see Theorem~\ref{thm:abadi}). 
We re-frame the problem as for a given $(\epsilon,\delta,\sigma)$ there are many choices $T$ depending on $K$ and sample size sequence $s$, where 
{\em $T \geq k^2/(\mbox{const}\cdot \epsilon)$} (see Theorem \ref{thm:Main}). 

This appendix provides the proof of Theorem \ref{thm:Main}. It follows a sequence of steps:
In Section \ref{sec:abadian} we discuss the analysis of \citep{abadi2016deep} and explain where we will improve. This leads in Section \ref{sec:improved} to an improved analysis yielding a first generally applicable Theorem \ref{thm1}; DP definitions/tools with a key lemma (generalized from \citep{abadi2016deep}) are discussed in  Section \ref{app:def} and the proof of Theorem \ref{thm1} is in Section \ref{app-dp1}.
As a consequence we derive in Section \ref{app:simplethm} a simplified characterization in the form of Theorem \ref{thm:simpleF}. Finally, we introduce more coarse bounds in order to extract the more readable Theorem \ref{thm:Main} in Section \ref{sec:mainproof}.


\subsection{DP-SGD Analysis by Abadi et al.} 
\label{sec:abadian}

\citep{abadi2016deep} proves the following theorem (rephrased using our notation substituting $q=s/N$):

\begin{theorem} \label{thm:abadi} There exist constants $c_1$ and $c_2$ so that given a sample size sequence $s_i=s$
and  number of steps $T$, for any $\epsilon < c_1 T (s/N)^2$, Algorithm \ref{alg:DPs} is $(\epsilon,\delta)$-differentially private for any $\delta>0$ if we choose
$$ \sigma \geq c_2 \frac{(s/N)\cdot \sqrt{T \ln (1/\delta)}}{\epsilon}.
$$
\end{theorem}


The interpretation of Theorem \ref{thm:abadi}, however, 
subtle: 
The condition on $\epsilon$ in Theorem \ref{thm:abadi} is equivalent to 
$$ 1/\sqrt{c_1} < z \mbox{ where } z=(s/N) \cdot \sqrt{T/\epsilon}.
$$
Substituting this into the bound for $\sigma$ yields
\begin{equation} \sigma \geq (c_2\cdot z) \cdot \sqrt{\frac{\ln (1/\delta)}{\epsilon}}.\label{interAbadi}
\end{equation}
This formulation only depends on $T$ through the definition of $z$. Notice that $z$ may be as small as $1/\sqrt{c_1}$. In fact, it is unclear how $z$ depends on $T$ since $T$ is equal to the total number $K$ of gradient computations over all local rounds performed on the local data set divided by the mini-batch size $s$, i.e., $T=K/s$, hence, $z=(K/N) \cdot \sqrt{1/(T\epsilon)}$. This shows that for fixed $K$ and $N$, we can increase $T$ as long as $1/\sqrt{c_1}<z$, or equivalently,
\begin{equation}
    T< c_1 (K/N)^2/\epsilon \label{upT}
\end{equation}
(notice that the original constraint on $\epsilon$ in Theorem \ref{thm:abadi} directly translates into this upper bound on $T$ by using $T=K/s$). Since $\sigma$ cannot be chosen too large (otherwise the final global model has too much noise), $\epsilon$, see (\ref{interAbadi}), cannot be very small. 
Therefore, (\ref{upT})
puts an upper bound on $T$ which is in general much less than $K$ for practically sized large data sets ($K$ equals the maximum possible number of rounds for mini-batch size $s=1$). 

Rather than applying Theorem \ref{thm:abadi}, we can directly use the moment accountant method of its proof to analyse specific parameter settings. It turns out that $T$ can be much larger than upper bound (\ref{upT}). In this paper we formalize this insight (by showing that `constants' $c_1$ and $c_2$ can be chosen as functions of $T$ and other parameters) and show a lower bound on $\sigma$ which does not depend on $T$ at all -- in fact $z$ in (\ref{interAbadi}) can be characterized as a constant independent of any parameters. This will show that $\sigma$ can remain small to at least a lower bound that only depends on the privacy budget. 


\subsection{A General Improved DP-SGD Analysis}
\label{sec:improved}


We generalize  Theorem \ref{thm:abadi} \citep{abadi2016deep}:

\begin{theorem} \label{thm1}
%
We assume 
that $\sigma = \sigma_i$ with 
$\sigma\geq 216/215$
for all rounds $i$.
Let
$$r=r_0\cdot 2^3 \cdot \left(\frac{1}{1-u_0}+\frac{1}{1-u_1} \frac{e^3}{\sigma^3}\right) e^{3/\sigma^2}
\mbox{ with } 
u_0=\frac{2\sqrt{r_0\sigma}}{\sigma-r_0} \mbox{ and } u_1=\frac{2e\sqrt{r_0\sigma}}{(\sigma-r_0)\sigma},$$
where $r_0$ is such that it satisfies
\begin{equation} r_0\leq 1/e, \ u_0<1, \mbox{ and } u_1<1.\nonumber
\end{equation}
Let the sample size sequence satisfy $s_{i}/N\leq r_0/\sigma$. 
For $j=1,2,3$ we define $\hat{S}_j$ (resembling an average over the sum of $j$-th powers of $s_{i}/N$) 
with related constants $\rho$ and $\hat{\rho}$:
$$\hat{S}_j = \frac{1}{T} \sum_{i=0}^{T-1} \frac{s_{i}^j }{N(N-s_{i})^{j-1}}, \ \
 \frac{\hat{S}_1\hat{S}_3}{\hat{S}_2^2}\leq \rho
 \mbox{ and }
 \frac{\hat{S}_1^2}{\hat{S}_2}\leq \hat{\rho} .$$ 

Let $\epsilon = c_1 T \hat{S}_1^2$.
Then, Algorithm \ref{alg:DP} is $(\epsilon, \delta)$-differentially private if

\begin{eqnarray*}
\sigma \geq \frac{2}{\sqrt{c_0}} \frac{\sqrt{ \hat{S}_2 T (\epsilon +\ln (1/\delta)) }}{\epsilon}
\mbox{ where } c_0=c(c_1)
 \mbox{ with } 
c(x) = \min \left \{ \frac{\sqrt{2r\rho x+1} -1}{r \rho x}, \frac{2}{\hat{\rho} x }  
 \right \}.
\end{eqnarray*}
\end{theorem}

This generalizes Theorem \ref{thm:abadi} 
where all $s_i=s$ 
are constant. 
First, Theorem \ref{thm1} covers a much broader class of sample size sequences that satisfy bounds on their 'moments' $\hat{S}_j$ (this is more clear as a consequence of Theorem \ref{thm1}).
Second, our detailed analysis provides a tighter bound in that it makes the relation between ``constants'' $c_0$ and $c_1$ explicit, contrary to \citep{abadi2016deep}. Exactly due to this relation $c_0=c(c_1)$ we are able to prove in Appendix \ref{app:simplethm}
Theorem \ref{thm:simpleF} as a consequence of Theorem \ref{thm1} by considering the case $c(c_1)=2/(\hat{\rho}c_1)$.

In order to prove Theorem \ref{thm1}, we first set up the differential privacy framework of \citep{abadi2016deep} in Appendix \ref{app:def}. 
Here we enhance a core lemma by proving a concrete bound rather than an asymptotic bound on the so-called $\lambda$-th moment which plays a crucial role in the differential privacy analysis. The concrete bound makes explicit the higher order error term in \citep{abadi2016deep}.

In Appendix \ref{app-dp1} we generalize  Theorem \ref{thm:abadi} of \citep{abadi2016deep} by proving Theorem \ref{thm:simpleF} using the core lemma of Appendix \ref{app:def}. 

\subsubsection{Definitions and Main Lemma}
\label{app:def}

We base our proofs on the framework and theory presented in \citep{abadi2016deep}. In order to be on the same page we repeat and cite word for word their definitions:

For neighboring databases $d$ and $d'$, a
mechanism ${\cal M}$, auxiliary input $\texttt{aux}$, and an outcome $o$,
define the privacy loss at $o$ as
$$ c(o;{\cal M}, \texttt{aux}, d, d')
= \ln
\frac{Pr[{\cal M}(\texttt{aux}, d) = o]}
{Pr[{\cal M}(\texttt{aux}, d') = o]}.
$$
For a given mechanism ${\cal M}$, we define the $\lambda$-th moment
$\alpha_{{\cal M}}(\lambda ; \texttt{aux}, d, d')$ as the log of the moment generating function evaluated at the value $\lambda$:
$$
\alpha_{{\cal M}}(\lambda ; \texttt{aux}, d, d')
=
\ln \textbf{E}_{o\sim {\cal M}(\texttt{aux},d)}[\exp(\lambda \cdot c(o;{\cal M}, \texttt{aux}, d, d'))].
$$
We define
$$ \alpha_{{\cal M}}(\lambda) = \max_{\texttt{aux},d,d'} 
\alpha_{{\cal M}}(\lambda ; \texttt{aux}, d, d')
$$
where the maximum is taken over all possible $\texttt{aux}$ and all
the neighboring databases $d$ and $d'$.


We first take Lemma 3 from \citep{abadi2016deep} and make explicit their order term $O(q^3\lambda^3/\sigma^3)$ with $q=s_{i,c}$ and $\sigma=\sigma_i$ in our notation. The lemma considers as mechanism ${\cal M}$ the $i$-th round of gradient updates and we abbreviate $\alpha_{{\cal M}}(\lambda)$ by $\alpha_i(\lambda)$. The auxiliary input of the mechanism at round $i$ includes all the output of the mechanisms of previous rounds (as in \citep{abadi2016deep}).







For the local mini-batch SGD  the mechanism ${\cal M}$ of the $i$-th round is given by 
$${\cal M}(\texttt{aux},d) = \sum_{h=0}^{s_{i}-1} [\nabla f(\hat{w},\xi_h)]_C +{\cal N}(0,C^2\sigma_i^2\textbf{I}),$$ where $\hat{w}$ is the local model at the start of round $i$
which is replaced by a new 
global model $\hat{v}$ 
as soon as a new $\hat{v}$ is received from the server (see \cal{ISRReceive}), and where $\xi_h$ are drawn from training data $d$, and $[.]_C$ denotes clipping (that is $[x]_C=x/\max\{1,\|x\|_2/C\}$). In order for ${\cal M}$ to be able to compute its output, it needs to know the global models received  in round $i$ and it needs to know the starting local model $\hat{w}$. To make sure ${\cal M}$ has all this information,
$\texttt{aux}$ represents  the collection of all outputs generated by the mechanisms of previous rounds $<i$ together with the global models received in round $i$ itself. 

In the next subsection we will use the framework of \citep{abadi2016deep} and apply its composition theory to derive bounds on the privacy budget $(\epsilon,\delta)$ for the whole computation consisting of $T$ rounds that reveal the outputs of the mechanisms for these $T$ rounds as described above.




We remind the reader that $s_i/N$ is the probability of selecting a sample from a sample set (batch) of size $s_i$ out of a training data set $d'$ of size $N=|d'|$;  $\sigma_i$ corresponds to the ${\cal N}(0,C^2\sigma_i^2 \textbf{I})$ noise added to the mini-batch gradient computation in round $i$ (see the mechanism described above).

\begin{lemma} \label{lemthree} 
Assume a constant $r_0<1$ and deviation $\sigma_i\geq 216/215$ 
such that $s_{i}/N\leq r_0/\sigma_i$.
Suppose that $\lambda$ is a positive integer with
$$ \lambda \leq \sigma_i^2 \ln \frac{N}{s_{i} \sigma_i}$$
and define
$$
U_0(\lambda) = \frac{2  \sqrt{\lambda r_0/\sigma_i}     
}
{
 \sigma_i -r_0
} \mbox{ and }
U_1(\lambda) = \frac{ 2e \sqrt{\lambda r_0/\sigma_i}   
}
{
 (\sigma_i-r_0) \sigma_i
}.
$$
Suppose $U_0(\lambda)\leq u_0<1$ and $U_1(\lambda)\leq u_1<1$ for some constants $u_0$ and $u_1$.
Define
\begin{eqnarray*}
r &=& r_0 \cdot 2^3 \left(\frac{1}{1-u_0} + \frac{ 1}{1-u_1} \frac{e^3}{\sigma_i^3} \right ) \exp(3/\sigma_i^2).
\end{eqnarray*}

Then,
\begin{equation*}
\alpha_i(\lambda) \leq  \frac{s_{i}^2 \lambda (\lambda+1)}{N(N-s_{i}) \sigma_i^2} +\frac{r}{r_0} \cdot \frac{s_{i}^3\lambda^2(\lambda+1)}{N (N-s_{i})^2\sigma_i^3}.
\end{equation*}
\end{lemma}





\vspace{.5cm}

\noindent
{\bf Proof.} 
The start of the proof of Lemma 3 in \citep{abadi2016deep} implicitly uses the proof of Theorem A.1 in \citep{dwork2014algorithmic}, which up to formula (A.2) shows how the 1-dimensional case translates into a privacy loss that corresponds to the 1-dimensional problem defined by $\mu_0$ and $\mu_1$ in the proof of Lemma 3 in \citep{abadi2016deep}, and which shows at the end of the proof of Theorem A.1 (p. 268 \citep{dwork2014algorithmic}) how the multi-dimensional problem transforms into the 1-dimensional problem. In the notation of Theorem A.1, $f(D)+{\cal N}(0,\sigma^2\textbf{I})$ represents the general (random) mechanism ${\cal M}(D)$, which for Lemma 3 in  \citep{abadi2016deep}'s notation 
should be interpreted as the batch computation 
$${\cal M}(d)=\sum_{h\in J} f(d_h) +{\cal N}(0,\sigma^2\textbf{I})$$ for a random sample/batch $\{ d_h\}_{h\in J}$. Here, $f(d_h)$ (by abuse of notation -- in this context $f$ does not represent the objective function) represent clipped gradient computations $\nabla f(\hat{w};d_h)$ where $\hat{w}$ is the last received global model with which round $i$ starts (Lemma 3 in \citep{abadi2016deep} uses clipping constant $C=1$, hence ${\cal N}(0,C^2\sigma^2\textbf{I})={\cal N}(0,\sigma^2\textbf{I})$).  

Let us detail the argument of the proof of Lemma 3 in \citep{abadi2016deep} in order to understand what flexibility is possible: We consider two data sets $d=\{d_1,\ldots,d_{N-1}\}$ and $d'=d + \{d_N\}$, where $d_N\not\in d$ represents a new data base element so that $d$ and $d'$ differ in exactly one element. The size of $d'$ is equal to $N$. We define 
vector $x$ as the sum
$$x=\sum_{J\setminus \{N\}} f(d_i).$$
Let
$$z=f(d_N).$$ 
If we consider data set $d$, then sample set $J\subseteq \{1,\cdots, N-1\}$ and mechanism ${\cal M}(d)$ returns
$${\cal M}(d)=\sum_{h\in J} f(d_h) +{\cal N}(0,\sigma^2\textbf{I})
=\sum_{h\in J\setminus \{N\}} f(d_h) +{\cal N}(0,\sigma^2\textbf{I}) = x + {\cal N}(0,\sigma^2\textbf{I}).
$$
If we consider data set $d'$, then $J\subseteq \{1,\cdots, N\}$ contains $d_N$ with probability $q=|J|/N$ ($|J|=s_i$ is the sample size used in round $i$). In this case
mechanism ${\cal M}(d')$ returns\footnote{This is actually a subtle argument: We do not have fixed constant sample sizes, instead we have probabilistic sample sizes with a predetermined expectation. The idea is to add each data element to the sample with probability $s_i/N$. This means that the sample size is equal to $s_i$ in expectation. This allows one to compare two samples that differ in exactly one element $d_N$ (as is done in this argument). If one uses fixed constant sample sizes, then ${\cal M}(d')=x+(1-q)\cdot {\cal N}(0,\sigma^2{\bf I})+ q\cdot {\cal N}(f(d_N)-f(d_h),\sigma^2{\bf I})$ for $z=f(d_N)$ and some $h\in J$. Now $\|f(d_N)-f(d_h)\|_2\leq \|f(d_N)\|_2+\|f(d_h)\|_2\leq 2C=2$ (for $C=1$) and we pay a factor 2 penalty. In the $f$-DP framework we actually consider the latter and work with fixed (non-probabilistic) constant sample sizes. In this paper and, what should have been assumed in \citep{abadi2016deep} and is actually implemented in the Opacus library \citep{Xpacus}, we assume a probabilistic sample size and safe the factor $2$. We notice that even if we aim at a constant sample size sequence with sample sizes $s$, we can reinterpret the $s_i$ as the actual chosen probabilistic sample size with $\mathbb{E}[s_i]=s$ and apply our theory that holds for varying sample size sequences (we need to formulate an upper bound on $s_{max}$ which holds with probability 'close to 1' and this will determine $\theta$ in Theorem \ref{thm:Main}). }
$${\cal M}(d')=\sum_{h\in J} f(d_h) +{\cal N}(0,\sigma^2\textbf{I})
=f(d_N) + \sum_{h\in J\setminus \{N\}} f(d_h) +{\cal N}(0,\sigma^2\textbf{I})
= z+x+ {\cal N}(0,\sigma^2\textbf{I})
$$
with probability $q$. It returns
$${\cal M}(d')=\sum_{h\in J} f(d_h) +{\cal N}(0,\sigma^2\textbf{I})
= \sum_{h\in J\setminus \{N\}} f(d_h) +{\cal N}(0,\sigma^2\textbf{I})
= x+ {\cal N}(0,\sigma^2\textbf{I})
$$
with probability $1-q$. Combining both cases shows that ${\cal M}(d')$ represents a mixture of two Gaussian distributions (shifted over a vector $x$):
$${\cal M}(d')= x + (1-q)\cdot {\cal N}(0,\sigma^2\textbf{I}) + q\cdot {\cal N}(z,\sigma^2\textbf{I}).$$

This high dimensional problem is transformed into a single dimensional problem 
at the end of the proof of Theorem A.1 (p. 268 \citep{dwork2014algorithmic}) by considering the one dimensional line from point $x$ into the direction of $z$, i.e., the line through points $x$ and $x+z$; the one dimensional line maps $x$ to the origin $0$ and $x+z$ to $\|z\|_2$. ${\cal M}(d)$ as wells as ${\cal M}(d')$ projected on this line are distributed as
$$ {\cal M}(d) \sim \mu_0 \mbox{ and } {\cal M}(d') \sim (1-q) \mu_0 + q \mu_1,$$
where
$$ \mu_0 \sim {\cal N}(0,\sigma^2) \mbox{ and } \mu_1 \sim {\cal N}(\|z\|_2,\sigma^2).
$$
In \citep{abadi2016deep} as well as in this paper the gradients are clipped (their Lemma 3 uses clipping constant $C=1$) and this implies
$$ \|z\|_2 = \| f(d_N)\|_2 \leq C=1.$$
Their analysis continues by assuming the worst-case in differential privacy, that is,
$$\mu_1 \sim {\cal N}(1,\sigma^2).
$$

Notice that the above argument analyses a local mini-batch SGD computation. 
Rather than using a local mini-batch SGD computation, can we use clipped SGD iterations which continuously update the local model:
\begin{equation}
 \hat{w}_{h+1}=\hat{w}_h-\eta_h \nabla [f(\hat{w}_h,\xi_h)]_C.
 \nonumber 
 \end{equation}
This should lead to faster convergence to good accuracy compared to a local minibatch computation.
However, 
the above arguments cannot proceed\footnote{Unless we assume a general upper bound on the norm of the Hessian of the objective function which should be large enough to cover a wide class of objective functions and small enough in order to be able to derive practical differential privacy guarantees.} because (in the notation used above where the $d_h$, $h\in J$, are the $\xi_h$, $h\in \{0,\ldots, s_i-1=|J|-1\}$)  selecting sample $d_N$ in iteration $h$ does not only influence the update computed in iteration $h$ but also influences all iterations after $h$ till the end of the round (because $f(d_N)$ updates the local model in iteration $h$ which is used in the iterations that come after). Hence, the dependency on $d_N$ is directly felt by $f(d_N)$ in iteration $h$ and indirectly felt in the $f(d_j)$ that are computed after iteration $h$. This means that we cannot represent distribution ${\cal M}(d')$ as a clean mix of Gaussian distributions with a mean $z$, whose norm is bounded by the clipping constant.








The freedom which we do have is replacing the local model by a newly received global model. This is because 
the 
updates $f(d_h)$, $h\in J$, computed locally in round $i$ 
have not yet been transmitted to the server and, hence, have not been aggregated into the global model that was received. In a way the mechanism ${\cal M}(d)$ is composed of two (or multiple if more newer and newer global models are received during the round) sums
$${\cal M}(d)=\sum_{h\in J_0} f_0(d_h) + \sum_{h\in J_1} f_1(d_h) +{\cal N}(0,\sigma^2\textbf{I}),$$
where $J=J_0 \cup J_1$ and $J_0$ represent local gradient computations, shown by $f_0(.)$,  based on the initial local model $\hat{w}$ and $J_1$ represent the local gradient computations, shown by $f_1(.)$, based on the newly received global model $\hat{v}$ which replaces $\hat{w}$. As one can verify, the above arguments are still valid for this slight adaptation.
%
%
As in Lemma 3 in \citep{abadi2016deep}
we can now translate our privacy loss to the 1-dimensional problem defined by $\mu_0 \sim {\cal N}(0,C^2\sigma^2)$ and $\mu_1 \sim {\cal N}(C,C^2\sigma^2)$ for $\|\nabla f(.,.)\|_2\leq C$ as in the proof of Lemma 3 (which after normalization with respect to $C$ gives the formulation of Lemma 3 in \citep{abadi2016deep} for $C=1$). 

The remainder of the proof of Lemma 3 analyses $\mu_0$ and the mix $\mu=(1-q)\mu_0+q\mu_1$ leading to  bounds for the expectations (3) and (4) in \citep{abadi2016deep} which only depend on $\mu_0$ and $\mu_1$. Here, $q$ is the probability of having a special data sample $\xi$ (written as $d_N$ in the arguments above) in the batch. In our algorithm $q=s_{i}/N$. So, we may adopt the statement of Lemma 3 and conclude for the $i$-th batch computation
\begin{equation*}
\alpha_i(\lambda) \leq  \frac{s_{i}^2 \lambda (\lambda+1)}{N(N-s_{i}) \sigma_i^2} +O \left(\frac{s_{i}^3\lambda^3}{N^3\sigma_i^3} \right).
\end{equation*}

In order to find an exact expression for the higher order term we look into the details of Lemma 3 of \citep{abadi2016deep}.
It computes an upper bound for the binomial tail
\begin{equation} 
\sum_{t=3}^{\lambda+1} {\lambda +1 \choose t} \mathbb{E}_{z\sim \nu_1}[((\nu_0(z)-\nu_1(z))/\nu_1(z))^t],
\label{tail}
\end{equation}
where
\begin{eqnarray}
&& \mathbb{E}_{z\sim \nu_1}[((\nu_0(z)-\nu_1(z))/\nu_1(z))^t] \nonumber
\\
&\leq &
\frac{(2q)^t(t-1)!!}{2 (1-q)^{t-1} \sigma^t} +
\frac{q^t}{(1-q)^t\sigma^{2t}} +
 \frac{(2q)^t \exp((t^2-t)/(2\sigma^2)) (\sigma^t (t-1)!! + t^t)
}
{
2 (1-q)^{t-1} \sigma^{2t}
} \nonumber \\
&=&
 \frac{(2q)^t (t-1)!!  ( 1+ \exp((t^2-t)/(2\sigma^2))  )
}
{
2 (1-q)^{t-1} \sigma^{t}
} 
+
 \frac{ q^t (1+ (1-q) 2^t\exp((t^2-t)/(2\sigma^2)) t^t )
}
{
2 (1-q)^{t} \sigma^{2t}
} \label{terms}  
\end{eqnarray}
%
%
%

Since $t\geq 3$, we have the coarse upper bounds
$$1\leq \frac{\exp((t^2-t)/(2\sigma^2))}{\exp((3^2-3)/(2\sigma^2))} \mbox{ and }
1\leq  \frac{(1-q) 2^t\exp((t^2-t)/(2\sigma^2)) t^t}{(1-q) 2^3\exp((3^2-3)/(2\sigma^2)) 3^3}.$$
By defining $c$ as $1$ plus the maximum of these two bounds,
$$c=
1+ \frac{\max \left \{1, 1/((1-q)\cdot 216) \right \}}{\exp(3/\sigma^2)},$$
we have (\ref{terms}) at most
\begin{equation}\leq
 \frac{(2q)^t (t-1)!!  c \exp((t^2-t)/(2\sigma^2))  
}
{
2 (1-q)^{t-1} \sigma^{t}
} 
+
 \frac{ q^t c (1-q) 2^t\exp((t^2-t)/(2\sigma^2)) t^t 
}
{
2 (1-q)^{t} \sigma^{2t}
}. \label{upterms}
\end{equation}
Generally (for practical parameter settings as we will find out), $q\leq 1-1/216$ which makes $c\leq 2$. In the remainder of this proof, we use $c=2$ and assume $q\leq 215/216$. In fact, assume in the statement of the lemma  that $\sigma=\sigma_i\geq 216/215$ which together with $q=s_i/N\leq r_0/\sigma_i$ and $r_0<1$ implies $q\leq 215/216$.

After multiplying (\ref{upterms}) with the upper bound for
$$ {\lambda +1 \choose t} \leq \frac{\lambda+1}{\lambda} \frac{\lambda^{t}}{t!}$$
and noticing that $(t-1)!!/t!\leq 1$ and $t^t/t! \leq e^t$
we get the addition of the following two terms
\begin{eqnarray*}
&& \frac{\lambda+1}{\lambda}\frac{\lambda^t (2q)^t    \exp((t^2-t)/(2\sigma^2))  
}
{
 (1-q)^{t-1} \sigma^{t}
} 
+
\frac{\lambda+1}{\lambda} \frac{ \lambda^t q^t (1-q) 2^t\exp((t^2-t)/(2\sigma^2)) e^t 
}
{
 (1-q)^{t} \sigma^{2t}
}.
\end{eqnarray*}
%
%
This is equal to
\begin{eqnarray}
&&
(1-q)\frac{\lambda+1}{\lambda} \left (\frac{\lambda 2q    \exp((t-1)/(2\sigma^2))  
}
{
 (1-q) \sigma
} \right )^t \nonumber
\\
&& \hspace{2cm} +
(1-q) \frac{\lambda+1}{\lambda} \left (\frac{ \lambda q   2\exp(1+(t-1)/(2\sigma^2)) 
}
{
 (1-q) \sigma^{2}
} \right )^t. \label{eqterms}
\end{eqnarray}
We notice that by using $t\leq \lambda+1$, $\lambda/\sigma^2 \leq \ln (1/(q\sigma))$ (assumption), and $q=s_{i,c}/N_c \leq r_0/\sigma$ we obtain
\begin{eqnarray*} 
\frac{\lambda 2q    \exp((t-1)/(2\sigma^2))  
}
{
 (1-q) \sigma
} &\leq &
\frac{\lambda 2q    \exp(\lambda/(2\sigma^2))  
}
{
 (1-q) \sigma
}
\leq 
\frac{2  \sqrt{\lambda q}     
}
{
 (1-q) \sigma
} = 
\frac{2  \sqrt{\lambda r_0/\sigma}     
}
{
 \sigma -r_0
} =U_0(\lambda)
\end{eqnarray*}
and
\begin{eqnarray*}
\frac{ \lambda q   2\exp(1+(t-1)/(2\sigma^2)) 
}
{
 (1-q) \sigma^{2}
}
&\leq &
\frac{ \lambda q   2 e\exp(\lambda/(2\sigma^2)) 
}
{
 (1-q) \sigma^{2}
}
\leq
\frac{ 2e \sqrt{\lambda q}   
}
{
 (1-q) \sigma^{2}
} = 
\frac{ 2e \sqrt{\lambda r_0/\sigma}   
}
{
 (\sigma -r_0) \sigma
} =U_1(\lambda).
\end{eqnarray*}



Together with our assumption on $U_0(\lambda)$ and $U_1(\lambda)$, this means that the binomial tail (\ref{tail}) is upper bounded by the two terms in (\ref{eqterms}) after substituting $t=3$, with the two terms multiplied by
$$ \sum_{j=0}^\infty U_0(\lambda)^j = \frac{1}
{1-U_0(\lambda)}\leq \frac{1}{1-u_0} \mbox{ and } \sum_{j=0}^\infty U_1(\lambda)^j = \frac{1}{1-U_1(\lambda)}\leq \frac{1}{1-u_1}$$
respectively.
For (\ref{tail}) this yields the upper bound
\begin{eqnarray*}
&& \frac{1}
{1-u_0} (1-q)\frac{\lambda+1}{\lambda} \left (\frac{\lambda 2q    \exp(1/\sigma^2)  
}
{
 (1-q) \sigma
} \right )^3
+
 \frac{1}{1-u_1}
(1-q)\frac{\lambda+1}{\lambda}
\left(\frac{ \lambda q   2\exp(1+1/\sigma^2) 
}
{
 (1-q) \sigma^{2}
} \right)^3 \\
&\leq &
\left( \frac{1}
{1-u_0}
  2^3 \exp(3/\sigma^2) 
+  \frac{1}{1-u_1}
\frac{ 2^3 \exp(3+3/\sigma^2)}{\sigma^3}
\right)
\cdot 
\frac{ \lambda^2 (\lambda+1) q^3}{(1-q)^2\sigma^3} .
\end{eqnarray*}
By the definition of $r$, we obtain the bound 
%
\begin{eqnarray*}
&\leq &
\frac{r}{r_0}
\cdot
\frac{\lambda^2(1+\lambda) q^3}{(1-q)^2\sigma^3},
\end{eqnarray*}
which finalizes the proof.

\subsubsection{Proof of Theorem \ref{thm1}}
\label{app-dp1}

The proof Theorem \ref{thm1} follows the line of thinking in the proof of Theorem 1 in \citep{abadi2016deep}. Our theorem applies to varying sample/batch sizes and for this reason introduces moments $\hat{S}_j$. Our theorem explicitly defines the constant used in the lower bound of $\sigma$ -- this is important for proving our second (main) theorem in the next subsection.

%
Theorem \ref{thm1} assumes $\sigma = \sigma_i$ for all rounds $i$ with $\sigma\geq 216/215$; 
constant $r_0\leq 1/e$ 
such that $s_{i}/N\leq r_0/\sigma$;  constant
\begin{eqnarray}
r &=& r_0 \cdot 2^3 \left(\frac{1}{1-u_0} + \frac{ 1}{1-u_1} \frac{e^3}{\sigma^3} \right )\exp(3/\sigma^2), \label{constr}
\end{eqnarray}
where
$$ u_0 = \frac{2  \sqrt{r_0\sigma}     
}
{
 \sigma -r_0
} \mbox{ and } u_1 = \frac{ 2e \sqrt{r_0 \sigma} 
}
{
 (\sigma-r_0) \sigma
}
$$
are both assumed $<1$. 


For $j=1,2,3$ we define\footnote{$s_{i}^j$ denotes the $j$-th power $(s_{i})^j$.}
$$\hat{S}_j = \frac{1}{T} \sum_{i=0}^{T-1} \frac{s_{i}^j}{N(N-s_{i})^{j-1}}
\mbox{ with } \frac{\hat{S}_1\hat{S}_3}{\hat{S}_2^2}\leq \rho, \
 \frac{\hat{S}_1^2}{\hat{S}_2}\leq \hat{\rho}.
$$
Based on these constants we define
$$c(x) = \min \left \{ \frac{\sqrt{2r\rho x+1} -1}{r \rho x}, \frac{2}{\hat{\rho} x}  
\right \}.
$$

Let $\epsilon = c_1 T \hat{S}_1^2$.
We want to prove Algorithm \ref{alg:DP} is $(\epsilon, \delta)$-differentially private if
$$\sigma \geq \frac{2}{\sqrt{c_0}} \frac{\sqrt{ \hat{S}_2 T (\epsilon +\ln (1/\delta)) }}{\epsilon} 
\mbox{ where } c_0=c(c_1).
$$

\vspace{.5cm}

\noindent
{\bf Proof.} 
For $j=1,2,3$, we define 
$$ S_j = \sum_{i=0}^{T-1} \frac{s_{i}^j}{N (N-s_{i})^{j-1}\sigma_i^j} \mbox{ and } 
S'_j = \frac{1}{T} \sum_{i=0}^{T-1} \frac{s_{i}^j \sigma_i^j}{N (N-s_{i})^{j-1}}.
$$
(Notice that $S'_1\leq r_0$.)
Translating Lemma \ref{lemthree} in this notation yields (we will verify the requirement/assumptions of Lemma \ref{lemthree} on the fly below)
$$\sum_{i=0}^{T-1} \alpha_i(\lambda) \leq S_2 \lambda (\lambda+1) + \frac{r}{r_0} S_3 \lambda^2(\lambda+1).$$

The composition Theorem 2 in \citep{abadi2016deep} shows that  our algorithm for client $c$ is $(\epsilon, \delta)$-differentially private for
\begin{eqnarray*}
\delta &\geq & \min_{\lambda}\exp \left (\sum_{i=0}^{T-1} \alpha_i(\lambda)-\lambda \epsilon \right ),
\end{eqnarray*}
where $T$ indicates the total number of batch computations and the minimum is over positive integers $\lambda$.
Similar to their proof we choose $\lambda$ such that
\begin{equation} S_2 \lambda (\lambda+1) + \frac{r}{r_0} S_3 \lambda^2(\lambda+1) -\lambda \epsilon \leq -\lambda \epsilon/2.
\label{Meq1}
\end{equation}
This implies that we can choose $\delta$ as small as $\exp(-\lambda\epsilon/2)$, i.e., if
\begin{equation}\delta \geq \exp(-\lambda\epsilon/2),\label{DPlam}
\end{equation}
then we have $(\epsilon,\delta)$-differential privacy.
%
%
After dividing by the positive integer $\lambda$, inequality (\ref{Meq1}) is equivalent to the inequality
 $$ S_2 (\lambda+1) + \frac{r}{r_0} S_3 \lambda (1+\lambda) \leq \epsilon/2,$$
 which is equivalent to
 $$ (\lambda+1) \left( 1+ \frac{r}{r_0} \frac{S_3}{S_2} \lambda \right) \leq \frac{\epsilon}{2S_2}.
 $$
 This is in turn implied by
 \begin{equation} \lambda +1 \leq c_0 \frac{\epsilon}{2S_2} 
 \label{Meq2}
 \end{equation}
 together with
 $$
 c_0 \frac{\epsilon}{2S_2} 
\left ( 1+ \frac{r}{r_0} \frac{S_3}{S_2} c_0 \frac{\epsilon}{2S_2} \right ) \leq \frac{\epsilon}{2S_2},
 $$
 or equivalently,
 \begin{equation}
 c_0 
\left ( 1+ \frac{r}{2r_0} \cdot c_0 \cdot \frac{S_3}{S_2^2}\epsilon  \right) \leq 1.
 \label{Meq3}
 \end{equation}
 We use
 \begin{equation} \epsilon = c_1\cdot T \hat{S}_1^2 = c_1 \cdot S_1 S'_1
 \label{eqepsc1}
 \end{equation}
 (for constant $\sigma_i=\sigma$).
 This translates our requirements (\ref{Meq2}) and (\ref{Meq3})  into
 \begin{equation} \lambda +1 \leq \frac{c_0c_1}{2} \frac{S_1S'_1}{S_2} \mbox{ and } \label{Meq5}
 \end{equation}
 \begin{equation}
c_0  \left (1+\frac{r}{2r_0}\cdot c_0 c_1 \frac{S_1S_3}{S_2^2}S'_1 \right ) \leq 1. 
 \label{Meq4}
 \end{equation}
 
 Since we assume
 $$\frac{S_1S_3}{S_2^2}= \frac{\hat{S}_1\hat{S}_3}{\hat{S}_2^2}\leq \rho $$
 and since we know that $S'_1\leq r_0$,
 requirement (\ref{Meq4}) 
 is implied by 
 $$ c_0(1+\frac{r \rho}{2} \cdot c_0 c_1 ) \leq 1, $$
 or equivalently
 \begin{equation} c_1 \leq \frac{1-c_0}{\frac{r \rho}{2}  c_0^2}.
 \label{Eqc1a}
 \end{equation}

 Also notice that for constant $\sigma_i=\sigma$ we have $S'_1=S_1\sigma^2/T$. 
 Together with 
 $$\frac{S_1^2}{S_2}= \frac{\hat{S}_1^2}{\hat{S}_2} T\leq \hat{\rho} T
 $$
 we obtain from (\ref{Meq5})
 \begin{equation} \lambda +1 \leq 
 \frac{c_0c_1}{2} \frac{S_1S'_1}{S_2}
 \leq 
 \frac{c_0c_1}{2} \hat{\rho}
 \sigma^2.
\label{upprlambda}
\end{equation}
Generally, if 
\begin{equation}
c_1\leq \frac{2}{\hat{\rho}c_0}, \label{Eqc1b}
\end{equation}
then (\ref{upprlambda}) implies $\lambda \leq \sigma^2$: Hence,
(a) 
for our choice of $u_0$ and $u_1$ in this theorem, $U_0(\lambda)\leq u_0$ and $U_1(\lambda)\leq u_1$ as defined in Lemma \ref{lemthree}, and
(b) the condition $\lambda \leq \sigma_i^2 \ln \frac{N_c}{s_{i,c}\sigma_i}$ is satisfied  (by assumption, $\frac{N_c}{s_{i,c}\sigma_i}\geq 1/r_0\geq e$).
This implies that Lemma \ref{lemthree} is indeed applicable.

For the above reasons we strengthen the requirement on $\epsilon$ (conditions 
(\ref{Eqc1a}) and (\ref{Eqc1b}) with (\ref{eqepsc1})) to
$$\epsilon \leq \min \left \{\frac{1-c_0}{\frac{r \rho}{2}  c_0^2}, \frac{2}{\hat{\rho} c_0} \right \} \cdot S_1S'_1$$
For constant $\sigma_i=\sigma$, we have
$$S_1S'_1= T \hat{S}_1^2,$$
hence, we need
\begin{equation} \epsilon \leq \min \left \{\frac{1-c_0}{\frac{r \rho}{2} 
c_0^2}, \frac{2}{\hat{\rho} c_0} \right \} \cdot T\hat{S}_1^2
\label{Meq6}
\end{equation}
Summarizing (\ref{Meq6}), (\ref{Meq2}), and (\ref{DPlam}) for some positive integer $\lambda$ proves $(\epsilon,\delta)$-differential privacy.


Condition (\ref{DPlam}) (i.e., $\exp(-\lambda \epsilon /2) \leq \delta $) is equivalent to
\begin{equation} \ln(1/\delta) \leq \frac{\lambda \epsilon}{2} \label{Meq7}
\end{equation}
If 
\begin{equation}\lambda = \lfloor c_0 \frac{\epsilon}{2S_2} \rfloor -1
\label{Meq9}
\end{equation}
is positive, then it satisfies (\ref{Meq2}) and we may use this $\lambda$ in (\ref{Meq7}). This yields the condition
$$ \ln(1/\delta) \leq \left(\lfloor c_0 \frac{\epsilon}{2S_2} \rfloor -1\right) \frac{\epsilon}{2},$$
which is implied by
$$ \ln(1/\delta) \leq \left( c_0 \frac{\epsilon}{2S_2} -2\right) \frac{\epsilon}{2} =
\frac{c_0}{4S_2}\epsilon^2 -\epsilon.$$
%
For constant $\sigma_i=\sigma$ we have $S_2=\hat{S}_2 T/\sigma^2$
and the latter inequality is equivalent to
\begin{equation}
\sigma \geq  \frac{2}{\sqrt{c_0}} \frac{\sqrt{\hat{S}_2} \sqrt{T (\epsilon +\ln (1/\delta))}}{\epsilon}.
\label{Meq8}
\end{equation}
Summarizing, if (\ref{Meq6}), (\ref{Meq8}), and the lambda value (\ref{Meq9}) is positive, then this shows $(\epsilon,\delta)$-differential privacy.

The condition (\ref{Meq9}) being positive follows from
$$\frac{4S_2}{c_0}\leq \epsilon.$$
Substituting $S_2=\hat{S}_2 T/\sigma^2$ yields the equivalent condition
$$\frac{4T\hat{S}_2}{\sigma^2 c_0}\leq \epsilon$$
or
$$ \sigma\geq \frac{2}{\sqrt{c_0}} \sqrt{\hat{S}_2} \frac{\sqrt{T\epsilon}}{\epsilon},$$
which is implied by (\ref{Meq8}).
Summarizing, if (\ref{Meq6}) and (\ref{Meq8}),  then this shows $(\epsilon,\delta)$-differential privacy.
Notice that (\ref{Meq8}) corresponds  to Theorem 1 in \citep{abadi2016deep} where all $s_{i}$ are constant implying $\sqrt{\hat{S}_2}=q/\sqrt{1-q}$  in their notation. 

We are interested in a slightly different formulation:
 Given
$$ c_1= \min \left \{\frac{1-c_0}{\frac{r\rho}{2} c_0^2}, \frac{2}{\hat{\rho} c_0} \right \} $$
what is the maximum possible $c_0$ (which minimizes $\sigma$ implying more fast convergence to an accurate solution). We need to satisfy $c_0\leq 2/(\hat{\rho}c_1)$ and
$$ \frac{r\rho}{2} c_1 c_0^2 +c_0 -1 \leq 0,$$
that is,
$$( c_0 +1/(r\rho c_1))^2 \leq 1/ \left(\frac{r\rho}{2} c_1 \right)+ 1/(r\rho c_1)^2,$$
or
$$ c_0 \leq \sqrt{1/ \left(\frac{r\rho}{2} c_1 \right) + 1/ \left(r\rho c_1 \right)^2} - 1/(r \rho c_1) =\frac{\sqrt{2 r\rho c_1+1} -1}{r \rho c_1}
.$$
We have
$$ c_0 = \min \left \{ \frac{\sqrt{2 r\rho c_1+1} -1}{r \rho c_1}, 2/(\hat{\rho} c_1)  
\right \} = c(c_1).
$$
This finishes the proof.

\subsection{A Simplified Characterization}
\label{app:simplethm}

So far, we  have generalized  Theorem \ref{thm:abadi} in Appendix
 \ref{appDP}
 in a non-trivial way by analysing increasing sample size sequences, by making explicit the higher order error term in \citep{abadi2016deep},  and by providing a precise functional relationship among the constants $c_1$ and $c_2$ in Theorem  \ref{thm:abadi}. The resulting Theorem \ref{thm1} is still hard to interpret. The next theorem is a consequence of Theorem \ref{thm1} and brings us the interpretation we look for.

\begin{theorem} \label{thm:simpleF}
For sample size sequence $\{s_i\}_{i=0}^{T-1}$
the total number of local SGD iterations is equal to $K=\sum_{i=0}^{T-1} s_i$. We define
 the mean $\bar{s}$ and maximum $s_{max}$ and their quotient $\theta$ as
\begin{eqnarray*}
&& \bar{s} = \frac{1}{T} \sum_{i=0}^{T-1} s_i = \frac{K}{T}, \ \
 s_{max} = \max \{ s_0, \ldots, s_{T-1}\}, \
 \mbox{ and } \ \
  \theta = \frac{s_{max}}{\bar{s}}.
  \end{eqnarray*}
We define
\begin{align*}
    h(x) &= \left(\sqrt{1+(e/x)^2}-e/x\right)^2,  \ \ 
    g(x) = \min \left \{ \frac{1}{ex}, h(x) \right \},
\end{align*}
%
and denote by $\gamma$ the smallest solution satisfying
%
%
%
%
\begin{eqnarray*}
&& \gamma  \geq  \frac{2}{1-\bar{\alpha}} + 
\frac{2^4 \cdot \bar{\alpha}  }{1-\bar{\alpha}}
\left( \frac{\sigma}{(1-\sqrt{\bar{\alpha}})^2} +\frac{1}{\sigma(1-\bar{\alpha})-2e\sqrt{\bar{\alpha}}}\frac{e^3}{\sigma}
\right)e^{3/\sigma^2}
\mbox{ with }
\bar{\alpha}= \frac{\epsilon N}{\gamma K}.
\nonumber
\end{eqnarray*}

If the following requirements are satisfied:
\begin{eqnarray}
\bar{s} &\leq& \frac{g \left (\sqrt{2(\epsilon+\ln(1/\delta))/\epsilon} \right )}{\theta} \cdot N, \label{treq1} \\
\epsilon &\leq& \gamma  h(\sigma)\cdot \frac{K}{N}, \label{treq2} \\
\epsilon &\geq& \gamma \theta^2 \cdot \frac{K}{N} \cdot \frac{\bar{s}}{N}, \mbox{ and} \label{treq3}\\
\sigma &\geq& \sqrt{2(\epsilon+\ln(1/\delta))/\epsilon}, \label{treq4}
\end{eqnarray}
then Algorithm \ref{alg:DP} is $(\epsilon, \delta)$-differentially private.
\end{theorem}

Its proof follows from analysing the requirements stated in Theorem \ref{thm1}. We will focus on the case where $c(x)=\frac{2}{\hat{\rho}x}$, which turns out to lead to practical parameter settings as discussed in the main body of the paper.

\noindent
{\bf Requirement on $r$ -- (\ref{req2f}):}
In Theorem \ref{thm1} we use
$$r=r_0\cdot 2^3 \cdot \left(\frac{1}{1-u_0}+\frac{1}{1-u_1} \frac{e^3}{\sigma^3}\right) e^{3/\sigma^2}$$
with 
$$u_0=\frac{2\sqrt{r_0\sigma}}{\sigma-r_0} \mbox{ and } u_1=\frac{2e\sqrt{r_0\sigma}}{(\sigma-r_0)\sigma},$$
where $r_0$ is such that it satisfies
\begin{equation} r_0\leq 1/e, \ u_0<1, \mbox{ and } u_1<1.\label{req1}
\end{equation}
In our application of Theorem \ref{thm1} we substitute $r_0=\alpha \sigma$. This translates the requirements of (\ref{req1}) into
\begin{equation}
  \alpha \leq \frac{1}{e\sigma}, \ \alpha<1, \mbox{ and }   \sigma > \frac{2e\sqrt{\alpha}}{1-\alpha}. \label{req2}
\end{equation}
As we will see in our derivation, we will require another lower bound (\ref{boundsigma}) on $\sigma$. We will use (\ref{boundsigma}) together with
$$
  \alpha \leq \frac{1}{e \sqrt{2(\epsilon+\ln(1/\delta))/\epsilon}}, \alpha < 1, \mbox{ and }   \sqrt{2(\epsilon+\ln(1/\delta))/\epsilon} > \frac{2e\sqrt{\alpha}}{1-\alpha} 
$$
to imply the needed requirement (\ref{req2}). These new bounds on $\alpha$ are in turn equivalent to
\begin{eqnarray}
&&\alpha \leq g(\epsilon, \delta) \mbox{ where }
\label{req2f} \\ && 
g(\epsilon, \delta) = \min \left\{ \frac{\sqrt{\epsilon}}{e\sqrt{2(\epsilon+\ln(1/\delta))}},  
\left(\sqrt{1+ \frac{e^2\epsilon}{2(\epsilon+\ln(1/\delta))}}-\frac{e\sqrt{\epsilon}}{\sqrt{2(\epsilon+\ln(1/\delta))/\epsilon}}\right)^2
 \right\} \nonumber
\end{eqnarray}
(notice that this implies $\alpha<1$). 

Substituting $r_0=\alpha\sigma$ in the formula for $r$ yields the expression
\begin{eqnarray}
r&=& 2^3\cdot \left( \frac{\sigma}{(1-\sqrt{\alpha})^2} +\frac{1}{\sigma (1-\alpha)-2e\sqrt{\alpha}}\frac{e^3}{\sigma}
\right) \cdot 
e^{3/\sigma^2} (1-\alpha)\alpha.\label{expr}
\end{eqnarray}

\noindent
{\bf Requirement on $s_i/N$ -- (\ref{req3}):}
In Theorem \ref{thm1} we also require $s_i/N\leq r_0/\sigma$ which translates into
\begin{equation}
s_i/N \leq \alpha. \label{req3}
\end{equation}

\noindent
{\bf Requirement on $\sigma$ -- (\ref{boundsigma}) and (\ref{loweps}):}
In Theorem \ref{thm1} we restrict ourselves to the case where function $c(x)$ attains the minimum $c(x)=2/(\hat{\rho} x)$. This happens when
$$\frac{\sqrt{2r\rho x+1}-1}{r\rho x}\geq \frac{2}{\hat{\rho}x}.$$
This is equivalent to
\begin{equation}x\geq 2r \frac{\rho}{\hat{\rho}^2}+\frac{2}{\hat{\rho}}.\label{lowx}
\end{equation}
Notice that in the lower bound for $\sigma$ in Theorem \ref{thm1} we use $c_0=c(x)$ for $x=c_1$, where $c_1$ is implicitly defined by
$$\epsilon=c_1 T \hat{S}_1^2$$
or equivalently
$$c_1 = \frac{\epsilon}{T\hat{S}_1^2}.$$
To minimize $\epsilon$, we want to minimize $c_1=x$. That is, we want $c_1=x$ to match the lower bound (\ref{lowx}). This lower bound is smallest if we choose the smallest possible $\rho$ (due to the linear dependency of the lower bound on $\rho$). Given the constraint on $\rho$ this means we choose
$$\rho= \frac{\hat{S}_1\hat{S}_3}{\hat{S}_2^2}.$$

For $c_1=x$ satisfying (\ref{lowx}) we have
$$c_0=c(c_1) = \frac{2}{\hat{\rho}x}.$$
Substituting this in the lower bound for $\sigma$ attains
$$\sigma \geq \frac{2}{\sqrt{c(c_1)}} \frac{\sqrt{\hat{S}_2T(\epsilon +\ln(1/\delta))}}{\epsilon} =
\sqrt{\frac{\hat{\rho}\hat{S}_2}{\hat{S}_1^2}} \sqrt{2(\epsilon+\ln(1/\delta))/\epsilon}.$$
In order to yield the best test accuracy we want to choose the smallest possible $\sigma$. Hence, we want to minimize the lower bound for $\sigma$ and therefore choose the smallest $\hat{\rho}$ given its constraints, i.e.,
$$\hat{\rho}=\frac{\hat{S}_1^2}{\hat{S}_2}.$$
This gives
\begin{equation} \sigma \geq \sqrt{2(\epsilon+\ln(1/\delta))/\epsilon}. \label{boundsigma}
\end{equation}
Notice that this lower bound implies $\sigma\geq 216/215$ and for this reason we do not state this as an extra requirement.

Our expressions for $\rho$, $\hat{\rho}$, and $c_1$ with $x=c_1$ shows that lower bound (\ref{lowx}) holds if and only if
\begin{equation}
\epsilon \geq \left(2r \frac{\hat{S}_3}{\hat{S}_1}+2\hat{S}_2\right) T.\label{loweps}
\end{equation}

\noindent
{\bf Requirement implying (\ref{loweps}):}
The definition of moments $\hat{S}_j$ imply
$$\hat{S}_1 = \frac{K}{TN}$$
and, since $s_i/N\leq \alpha<1$,
$$\hat{S}_j\leq \alpha^j /(1-\alpha)^{j-1}.$$
Lower bound (\ref{loweps}) on $\epsilon$ is therefore implied by
\begin{equation}
\epsilon \geq 2r\frac{\alpha^3}{(1-\alpha)^2} \frac{T^2 N}{K} + 2 \frac{\alpha^2}{1-\alpha} T.
\label{loweps1}
\end{equation}

We substitute
\begin{equation}
T = \beta \frac{K}{N} \label{Tbeta}
\end{equation} 
in (\ref{loweps1}) which yields the requirement
\begin{equation} \epsilon \frac{N}{K} \geq \frac{2r}{\alpha(1-\alpha)^2} (\alpha^2\beta)^2 + \frac{2}{1-\alpha} (\alpha^2\beta).
\label{loweps2}
\end{equation}
This inequality is implied by the combination of the following two inequalities:
\begin{equation}\alpha^2\beta \leq \frac{\epsilon N}{\gamma K}\label{loweps3a} \end{equation}
and
\begin{equation} 1\geq 
\frac{2r}{\alpha(1-\alpha)^2} \frac{\epsilon N}{K} \frac{1}{\gamma^2} + \frac{2}{1-\alpha} \frac{1}{\gamma}. \label{loweps3b}
\end{equation}

Inequality (\ref{loweps3b}) is equivalent to
\begin{equation} \gamma \geq 
\frac{2r}{\alpha(1-\alpha)^2} \frac{\epsilon N}{\gamma K}  + \frac{2}{1-\alpha}.\label{loweps3b1}
\end{equation}
This implies $$\gamma \geq \frac{2}{1-\alpha}\geq 2.$$
Also notice that $$\frac{1}{\beta}=\frac{K}{TN}=\hat{S}_1\leq \alpha$$
from which we obtain
$$1\leq \alpha \beta.$$
Let us define
\begin{equation}
    \bar{\alpha}=\frac{\epsilon N}{\gamma K}. \label{baradef}
\end{equation}
Inequalities $\gamma\geq 2$ and $1\leq \alpha\beta$ together with (\ref{loweps3a}) and the definition of $\bar{\alpha}$
imply 
\begin{equation} \alpha \leq \alpha^2\beta \leq \frac{\epsilon N}{\gamma K}= \bar{\alpha}\leq \frac{\epsilon N}{2 K}.\label{alphbarup}
\end{equation}

We will require 
\begin{equation}\bar{\alpha}< 1 \label{bara1} \end{equation}
and also $\sigma(1-\bar{\alpha})-2e\sqrt{\bar{\alpha}}>0$ i.e,
\begin{equation}
\sigma > \frac{2e\sqrt{\bar{\alpha}}}{1-\bar{\alpha}}.\label{bara2}
\end{equation}
Bounds (\ref{bara1}) and (\ref{bara2}) are equivalent to
\begin{equation}
    \bar{\alpha} \leq h(\sigma) \mbox{ where } h(\sigma) = \left(\sqrt{1+(e/\sigma)^2}-e/\sigma\right)^2. \label{bara3}
\end{equation}

With condition (\ref{bara3}) in place we may derive the upper bound
\begin{eqnarray*} && \frac{2r}{\alpha(1-\alpha)^2} \\
&=&
 \frac{2^4}{1-\alpha} \left( \frac{\sigma}{(1-\sqrt{\alpha})^2} +\frac{1}{\sigma(1-\alpha)-2e\sqrt{\alpha}}\frac{e^3}{\sigma}
\right)e^{3/\sigma^2} 
\\
&\leq&
\frac{2^4}{1-\bar{\alpha}} \left( \frac{\sigma}{(1-\sqrt{\bar{\alpha}})^2} +\frac{1}{\sigma(1-\bar{\alpha})-2e\sqrt{\bar{\alpha}}}\frac{e^3}{\sigma}
\right)e^{3/\sigma^2}
\end{eqnarray*}
because all denominators are decreasing functions in $\alpha$ and remain positive for $\alpha \leq \bar{\alpha}$.
Similarly,
$$\frac{2}{1-\alpha}\leq \frac{2}{1-\bar{\alpha}}.$$
These two upper bounds combined with (\ref{baradef})
show that (\ref{loweps3b1}) is implied by choosing
$$ \gamma = \gamma(\sigma, \epsilon N/K), $$
where $\gamma(\sigma, \epsilon N/K)$ is defined as the smallest solution of $\gamma$ satisfying
\begin{eqnarray}
\gamma && \geq  \frac{2}{1-\bar{\alpha}} + \label{gamma} \\
&& \frac{2^4 \cdot \bar{\alpha}  }{1-\bar{\alpha}} \left( \frac{\sigma}{(1-\sqrt{\bar{\alpha}})^2} +\frac{1}{\sigma(1-\bar{\alpha})-2e\sqrt{\bar{\alpha}}}\frac{e^3}{\sigma}
\right)e^{3/\sigma^2}, \nonumber
\end{eqnarray}
where $\bar{\alpha}= (\epsilon N/K) / \gamma$.
The smallest solution $\gamma$ will meet (\ref{gamma}) with equality. For this reason the minimal solution $\gamma$ will be at most the right hand side of (\ref{gamma}) where $\gamma$ is replaced by its lower bound $2$; this is allowed because this increases $\bar{\alpha}$ to the upper bound in (\ref{alphbarup}) and we know that the right hand side of (\ref{gamma}) increases in $\bar{\alpha}$ up to the upper bound in (\ref{alphbarup}) if the upper bound satisfies
$$ \frac{\epsilon N}{2K} \leq h(\sigma).$$
This makes requirement (\ref{bara3}) slightly stronger -- but in practice this stronger requirement is already satisfied because $K$ is several epochs of $N$ iterations making $\frac{\epsilon N}{2K}\ll 1$ while $\sigma\gg 1$ for small $\epsilon$ implying that $h(\sigma)$ is close to $1$.

Notice that
$\gamma = 2 + O(\bar{\alpha})$, hence, for small $\bar{\alpha}$ we have $\gamma\approx 2$.
A more precise asymptotic analysis reveals
$$ \gamma = 2 + (2+
2^4 \cdot   \left( \sigma + \frac{e^3}{\sigma^2}
\right)e^{3/\sigma^2} ) \bar{\alpha}  + O(\bar{\alpha}^{3/2}).
$$
Relatively large $\bar{\alpha}$ closer to $1$ will yield $\gamma \gg 2$. 

Summarizing
$$ 
\{ (\ref{Tbeta}), (\ref{loweps3a}), (\ref{baradef}), (\ref{bara3}), (\ref{gamma}) \} \Rightarrow (\ref{loweps}).
$$

\noindent 
{\bf Combining all requirements -- resulting in (\ref{gammadef}), (\ref{reqfins}), and (\ref{boundsigma}), or equivalently (\ref{ineqeps}), (\ref{ineqbars}),  and (\ref{boundsigma}):}
The combination of requirements (\ref{Tbeta}) and (\ref{loweps3a}) is equivalent to
\begin{equation}
\alpha \leq \sqrt{\frac{\epsilon }{\gamma T}} \label{alphaeps}
\end{equation}
(notice that $T$ and $\beta$ are not involved in any of the other requirements including those discussed earlier in this discussion, hence, we can discard (\ref{Tbeta}) and substitute this in (\ref{loweps3a})).
The combination of (\ref{baradef}), (\ref{bara3}), and (\ref{gamma}) is equivalent to
\begin{equation}
\frac{\epsilon N}{\gamma K}  \leq  
h(\sigma)
\mbox{ with } \gamma = \gamma \left (\sigma, \frac{\epsilon N}{K} \right ) \label{gammadef}
\end{equation}
(for the definition of $h(.)$ see (\ref{bara3}) and for $\gamma(.,.)$ see (\ref{gamma})).

We may now combine (\ref{alphaeps}),
(\ref{req2f}), and (\ref{req3}) into a single requirement
\begin{equation}
s_i/N \leq \min \left \{ g(\epsilon,\delta), \sqrt{\frac{\epsilon }{\gamma T}} \right \} \label{reqfins}
\end{equation}
(for the definition of $g(.,.)$ see (\ref{req2f})).
This shows that (\ref{gammadef}), (\ref{reqfins}), and (\ref{boundsigma}) (we remind the reader that the last condition is the lower bound on $\sigma \geq \sqrt{2(\epsilon+\ln(1/\delta))/\epsilon}$) implies $(\epsilon,\delta)$-DP by Theorem \ref{thm1}.

Let us rewrite these conditions. We introduce the mean $\bar{s}$ of all $s_i$ defined by
$$\bar{s} = \frac{1}{T} \sum_{i=0}^{T-1} s_i = \frac{K}{T}
$$
and we introduce the maximum $s_{max}$ of all $s_i$ defined by
$$ s_{max} = \max \{s_0,\ldots, s_{T-1}\}.$$
We define $\theta$ as the fraction
\begin{equation}
    \theta = \frac{s_{max}}{\bar{s}}. \label{thetaf}
\end{equation}
This notation allows us to rewrite 
$$
s_i/N \leq  \sqrt{\frac{\epsilon }{\gamma T}}$$
from (\ref{reqfins}) as
$$ \gamma  \frac{K}{N}  \frac{\bar{s}}{N}  \theta^2 \leq \epsilon.$$
From this we obtain that the requirements (\ref{gammadef}) and (\ref{reqfins}) are equivalent to 
\begin{equation}
\gamma \left (\sigma, \frac{\epsilon N}{K} \right ) \cdot \frac{K}{N}  \frac{\bar{s}}{N}  \theta^2 \leq \epsilon \leq \gamma \left (\sigma, \frac{\epsilon N}{K} \right ) \cdot h(\sigma)\frac{K}{N}
\label{ineqeps}
\end{equation}
and 
\begin{equation}
\theta \bar{s} \leq g(\epsilon,\delta) N.
\label{ineqbars}
\end{equation}
This alternative description shows that (\ref{ineqeps}), (\ref{ineqbars}), and (\ref{boundsigma}) with definitions for $h(.)$, $\gamma(.,.)$, $g(.,.)$, and $\theta$ in 
(\ref{bara3}), (\ref{gamma}), (\ref{req2f}), and (\ref{thetaf}) implies $(\epsilon,\delta)$-DP.
This proves Theorem \ref{thm:simpleF} (after a slight rewrite of the definitions of functions $h(.)$ and $g(.,.)$).

\subsection{Proof of the Main Theorem}
\label{sec:mainproof}

Theorem \ref{thm:simpleF} can already be used to a-priori set hyperparameters given DP and accuracy targets. Still, as discussed below, by making slight approximations (leading to slightly stronger constraints) we obtain the easy to interpret Theorem \ref{thm:Main} discussed in Section \ref{sec:Main}.

We set $\sigma$ as large as possible with respect to the accuracy we wish to have. Given this $\sigma$ we want to max out on our privacy budget. That is, we satisfy (\ref{treq4}) with equality, 
\begin{equation} \sigma = \sqrt{\frac{2(\epsilon +\ln(1/\delta))}{\epsilon}}. \label{sigeps}
\end{equation}
%
We discuss (\ref{sigeps}) with 
constraints (\ref{treq1}), (\ref{treq2}), and (\ref{treq3}) below:

\vspace{3mm}

\noindent
\textbf{Replacing (\ref{treq1}) and (\ref{treq2}):}
In practice, we need a sufficiently strong DP guarantee, hence, $\delta \leq 1/N$ and $\epsilon$ is small enough, typically $\leq 0.5$.
This means that we will stretch $\sigma$ to at least $\sqrt{2+4\ln N}$. A local data set of size $N=10000$ requires $\sigma\geq 6.23$; a local data set of size $N=100000$ requires $\sigma \geq 6.93$. 
For such $\sigma\geq 6$  we have $h(\sigma)\geq h(6)=0.42$ (since $h(\sigma)$ is increasing in $\sigma$). (For reference, $h(10)=0.58$, and for $\sigma \gg 1$ we have $h(\sigma)\approx 1$.)
From (\ref{sigeps}) we infer that $g(\sqrt{2(\epsilon+\ln(1/\delta))/\epsilon})=g(\sigma)= \min \{1/(e\sigma), h(\sigma)\}$. One can verify that $h(\sigma)-1/(e\sigma )$ is positive and increasing for $\sigma\geq 2.5$, hence, $g(\sigma)=1/(e\sigma)$ for $\sigma\geq 6$.
%
This reduces requirement (\ref{treq1}) to $\bar{s}\leq N/(e\sigma \theta)$ and requirement (\ref{treq2}) to  $\epsilon \leq 0.42\cdot \gamma K/N$. Notice that (\ref{treq3}) in combination with $\epsilon \leq \frac{\gamma \theta}{e\sigma} K/N$ implies condition $\bar{s}\leq N/(e\sigma \theta)$. This implies that (\ref{treq1}) and (\ref{treq2}) are satisfied for
$\epsilon \leq \min \{0.42 \cdot \gamma, \gamma \theta/(e\sigma) \} \cdot \frac{K}{N}$
 or, equivalently,
 $K \geq \epsilon \cdot \max\{2.4/\gamma, e\sigma/(\gamma \theta)\}$ epochs of size $N$.
 If $\theta \leq 6.85$, then $\sigma\geq 6 \geq 0.88 \cdot \theta = 2.4\cdot \theta/e$, hence, $\max\{2.4/\gamma, e\sigma/(\gamma \theta)\}=e\sigma/(\gamma \theta)$ and this reduces the condition on $K$ to
 %
 %
 $$K \geq \epsilon \sigma \cdot  \frac{e}{\gamma \theta}
 = \sqrt{2\epsilon(\epsilon+ \ln (1/\delta))} \cdot e/(\gamma \theta)
 \mbox{ epochs of size } N, $$
 where the equality follows from (\ref{sigeps}). 
In practical settings, $K$ consists of multiple (think 50 or 100s of) epochs (of size $N$) computation and this is generally true. We conclude that (\ref{treq1}) and (\ref{treq2}) are automatically satisfied by (\ref{treq3}) for general practical settings with $\delta\leq 1/N$, $\epsilon$ typically smaller than $0.5$, $N\geq 10000$, $\theta\leq 6.85$, and $K\geq \sqrt{2\epsilon (\epsilon +\ln (1/\delta))} \cdot e/(\gamma \theta)$ epochs, i.e.,
$$k \geq \sqrt{2\epsilon (\epsilon +\ln (1/\delta))} \cdot e/(\gamma \theta),$$
where $k=K/N$ as defined in Theorem \ref{thm:Main}.
By using $\epsilon\leq 0.5$, we can further weaken this condition to
$k\geq \sqrt{1/2 +\ln (1/\delta))} \cdot e/(\gamma \theta)$, or equivalently,
$$
(\gamma \theta /e)^2 k^2 \geq 1/2 + \ln(1/\delta).
$$
By using $\gamma\theta \geq 2$, we obtain condition (\ref{minK}) in Theorem \ref{thm:Main}.



\vspace{3mm}

\noindent
\textbf{Remaining constraint  (\ref{treq3}):} 
%
%
By using (\ref{sigeps}), (\ref{treq3}) 
can be equivalently recast as an upper bound on $\sigma$,
$$ \sigma \leq \sqrt{ \frac{2 (\epsilon +\ln(1/\delta))}{\gamma \theta^2 \cdot (K/N) \cdot (\bar{s}/N)}}.$$
Here, $\gamma$ is a function of $\sigma$ because $\gamma$ depends on $\epsilon$ in $\bar{\alpha}$ which is a function of $\sigma$ through (\ref{sigeps}). However, the definition of $\gamma$ shows that for small $\epsilon$, $\gamma$ is close to 2 and this gives 
$\sqrt{ \ln(1/\delta)/(\theta^2 \cdot (K/N) \cdot (\bar{s}/N))}$
as a good approximation of the upper bound.
Substituting $\bar{s}=K/T$ yields 
\begin{equation} \sigma \leq \frac{N\sqrt{T}}{K}
\sqrt{ \frac{2 (\epsilon+\ln(1/\delta))}{\gamma \theta^2}}. \label{sigT}
\end{equation}
For $\gamma\approx 2$ and $\theta=1$ (constant sample size), this upper bound compares to taking $c_2z\approx \sqrt{2}$ in (\ref{interAbadi}); we go beyond the analysis presented in
\citep{abadi2016deep} in a non-trivial way.

If $N\sqrt{T}/K$ is large enough, larger than the relatively small  $\sigma \sqrt{\theta^2(\gamma/2)/(\epsilon + \ln(1/\delta))}$, then upper bound (\ref{sigT}) is satisfied. That is, for given $K$ and $N$, we need $T$ to be large enough, or equivalently the mean sample/mini-batch  size $\bar{s}=K/T$  small enough. 
Squaring both sides of (\ref{sigT}) and moving terms yields the equivalent lower bound
\begin{equation}
T\geq \frac{\gamma}{2} \frac{\sigma^2 \theta^2 }{\epsilon + \ln(1/\delta)} \cdot (K/N)^2, \nonumber
\end{equation}
which after substituting (\ref{sigeps}) gives
\begin{equation}
T\geq  \frac{ \gamma \theta^2 }{\epsilon} \cdot (K/N)^2, \nonumber 
\end{equation}
which is condition (\ref{minT}) in Theorem \ref{thm:Main}.
In other words $T$ is at least a factor $\gamma \theta^2 /\epsilon$
larger than the square of the overall amount of local SGD computations measured in epochs (of size $N$). Notice that we have a lower bound on $T$ rather than an upper bound as in (\ref{upT})  from the theorem presented in \citep{abadi2016deep}.

\vspace{3mm}

\noindent 
\textbf{Remark Increasing Sample Size Sequence:}
We notice that polynomial increasing sample size sequences $s_i \sim qN i^p$ have $\bar{s}\approx [q N T^{p+1}/(p+1)]/T$ and $s_{max}=q N T^p$, hence, $\theta=1+p$. This shows that our theory covers e.g. linear increasing sample size sequences as discussed in \citep{van2020hogwild}, where is explained how this implies reduced round communication -- another metric which one may  trade-off against accuracy and total local number $K$ of gradient computations. 

\vspace{3mm}
\setcounter{footnote}{0}
\noindent 
\textbf{Remark Reusing the Local Data Set:}
We stress that  $T$ cannot be chosen arbitrarily large in Theorem \ref{thm:Main} as it is restricted by $K=k N$. Also $k$ cannot grow arbitrarily large since $kN=K\geq T \geq \frac{\gamma \theta^2}{\epsilon} \cdot k^2$, hence, $k\leq \frac{\epsilon}{\gamma \theta^2}\cdot N$.
This upper bound on $k$ does impose a constraint after which $(\epsilon,\delta)$-DP cannot be guaranteed -- so, $K$ and, hence, $T$ cannot increase indefinitely without violating the privacy budget.\footnote{We just derived that $K=kN\leq \frac{\epsilon}{\gamma \theta^2} \cdot N^2$. And we notice that besides the upper bound $T\leq K \leq \frac{\epsilon}{\gamma \theta^2} \cdot N^2$, we can also directly transform
condition $T\geq \frac{\gamma \theta^2}{\epsilon} \cdot k^2$ 
into $T\leq \frac{\epsilon}{\gamma \theta^2} \cdot (N/\bar{s})^2$
by substituting $k=\bar{s}T/N$ and rearranging terms.} 
Here we notice that repeated use of the same data set over multiple learning problems (one after another) is allowed as long as the number of epochs of gradient computing satisfies the upper bound $k\leq \frac{\epsilon}{\gamma \theta^2}\cdot N$.
Hence, the larger $N$ the more collaborative learning tasks the client can participate in. For typical values $\epsilon=0.2$, $\gamma \theta^2 \approx 2$, and a data set of size $N=10000$ we have $k\leq 1000$, which may accommodate about 10-20 learning tasks.

\textbf{Remark 
 Choosing $\epsilon$:}
%
%
We notice that, since $T\leq K=kN$ (this corresponds to the smallest possible mini-batch size $s=1$), lower bound (\ref{minT}) implies $kN\geq k^2/\epsilon$, hence, $\epsilon\geq k/N$ and we must have $\epsilon=\Omega(1/N)$.
Therefore, the smallest possible $\epsilon$ is $\Theta(1/N)$ and leads to $\sigma=\Theta(\sqrt{N\ln N})$ according to (\ref{maineq}). 
We notice that the theory in \citep{DN2002} 
for a similar but not exactly the same setting of DP-SGD strongly suggests for DP-SGD that unless the added Gaussian perturbation is as large as $\sqrt{N}$ almost the whole database can be recovered by a polynomial (in $N$) adversary; $\sigma=\Theta(\sqrt{N\ln N})$ seems needed if one wants cryptographical strong security.
However, in general, $\sigma=\Theta(\sqrt{N\ln N})$ is too large for sufficient accuracy. In practice
we choose $\epsilon=\theta(1)$:

In order to attain an accuracy comparable to the non-DP setting where no noise is added, the papers cited in Section \ref{sec:DP} generally require large $\epsilon$ (such that $\sigma$ can be small enough) -- which gives a weak privacy posture (a weak bound on the privacy loss). For example, when considering LDP (see Section \ref{sec:DP}), $10\%$ deduction in accuracy yields only $\epsilon = 50$ in \citep{abhishek2018privateFL}
and $\epsilon = 10.7$ in \citep{mohammad2020privateFL}, while  \citep{stacey2018privateFL,meng2020privateFL} show solutions for a much lower $\epsilon = 0.5$.
Similarly, when considering CDP (see Section \ref{sec:DP}), in order to remain close to the accuracy of the non-DP setting  
\citep{mohammad2020privateFL} requires $\epsilon = 8.1$, \citep{robin2017privateFL} requires $\epsilon = 8$, and \citep{mcmahan2017learning}  requires $\epsilon = 2.038$.

The theory presented in this paper allows relatively small Gaussian noise for small $\epsilon$: 
We only need to satisfy the main equation (\ref{maineq}). 
For example, in Section \ref{sec:experiment} simulations for the LIBSVM data set 
show $(\epsilon=0.05,\delta=1/N)$-DP is possible  while achieving good accuracy with $\sigma\approx 20$.
Such small $\epsilon$ is a significant improvement over existing literature.

\section{Tight Analysis using Gaussian DP} \label{app:GDP}

\citep{dong2021gaussian} explain an elegant alternative DP formulation based on hypothesis testing. From the attacker's perspective, it is natural to formulate the  problem of distinguishing two neighboring data sets $d$ and $d'$ based on the output of a DP mechanism ${\cal M}$ as a hypothesis testing problem:
$$H_0: \mbox{ the underlying data set is }d \ \ \ \ \mbox{ versus } \ \ \ \  H_1: \mbox{ the underlying data set is }d' .$$
We define the Type I and Type II errors by
$$\alpha_\phi = \textbf{E}_{o\sim {\cal M}(d)}[\phi(o)]  \mbox{ and } \beta_\phi = 1- \textbf{E}_{o\sim {\cal M}(d')}[\phi(o)],
$$
where $\phi$ in $[0,1]$ denotes the rejection rule which takes the output of the DP mechanism as input. We flip a coin and reject the null hypothesis with probability $\phi$. The optimal trade-off between Type I and Type II errors is given by the trade-off function
$$ T({\cal M}(d),{\cal M}(d'))(\alpha) = \inf_\phi \{ \beta_\phi \ : \ \alpha_\phi \leq \alpha \},$$ 
for $\alpha \in [0,1]$, where the infimum is taken over all measurable rejection rules $\phi$. If the two hypothesis are fully indistinguishable, then this leads to the trade-off function $1-\alpha$. We say a function $f\in [0,1]\rightarrow [0,1]$ is a trade-off function if and only if it is convex, continuous, non-increasing, and $f(x)\leq 1-x$ for $x\in [0,1]$. We define a mechanism ${\cal M}$ to be $f$-DP if 
$$
 T({\cal M}(d),{\cal M}(d')) \geq f
$$
for all neighboring $d$ and $d'$.
Proposition 2.5 in \citep{dong2021gaussian} is an adaptation of a result in \citep{wasserman2010statistical} and states that a mechanism is $(\epsilon,\delta)$-DP if and only if the mechanism is $f_{\epsilon,\delta}$-DP, where
$$f_{\epsilon,\delta}(\alpha) =
\max \{ 0, 1-\delta - e^{\epsilon}\alpha, (1-\delta-\alpha)e^{-\epsilon}\}.
$$
We see that $f$-DP has the $(\epsilon,\delta)$-DP  formulation as a special case. It turns out that the DP-SGD algorithm can be tightly analysed by using $f$-DP.

\vspace{3mm}

\noindent
{\bf Gaussian DP:}
In order to proceed \citep{dong2021gaussian} first defines Gaussian DP as another special case of $f$-DP as follows: We define the trade-off function
$$G_\mu(\alpha) = T({\cal N}(0,1),{\cal N}(\mu,1))(\alpha) = \Phi( \Phi^{-1}(1-\alpha) - \mu ),$$
where $\Phi$ is the standard normal cumulative distribution of ${\cal N}(0,1)$. We define a mechanism to be $\mu$-Gaussian DP if it is $G_\mu$-DP. Corollary 2.13 in \citep{dong2021gaussian} shows  that a mechanism is $\mu$-Gaussian DP if and only if it is $(\epsilon, \delta(\epsilon))$-DP for all $\epsilon\geq 0$, where
\begin{equation} \delta(\epsilon) = \Phi(-\frac{\epsilon}{\mu}+\frac{\mu}{2}) - e^{\epsilon} \Phi(-\frac{\epsilon}{\mu}-\frac{\mu}{2}).
\label{eq:gdp}
\end{equation}

\vspace{3mm}

\noindent
{\bf Subsampling:} Besides implementing Gaussian noise, DP-SGD also uses sub-sampling: For a data set $d$ of $N$ samples, ${\tt Sample}_s(d)$ selects a subset of size $s$ from $d$ uniformly at random. We define convex combinations
$$ f_p(\alpha) = p f(\alpha) + (1-p) (1-\alpha)$$
with corresponding $p$-sampling operator 
$$ C_p(f) = \min \{ f_p, f_p^{-1} \}^{**},
$$
where the conjugate $g^*$ of a function $g$ is defined as
$$ g^*(y) = \sup_x \{ yx -g(x) \}.$$ 
Theorem 4.2 in \citep{dong2021gaussian} shows that if a mechanism ${\cal M}$ on data sets of size $N$ is $f$-DP, then the subsampled mechanism ${\cal M}\circ {\tt Sample}_{s}$ is $C_{s/N}(f)$-DP.

\vspace{3mm}

\noindent
{\bf Composition:}
The tensor product $f\otimes g$ for trade-off functions $f=T(P,Q)$ and $g=T(P',Q')$ is well-defined by 
$$f\otimes g = T(P\times P',Q\times Q').$$
Let $y_i \leftarrow {\cal M}_i(\texttt{aux},d)$ with $\texttt{aux}=(y_1,\ldots, y_{i-1})$. Theorem 3.2 in \citep{dong2021gaussian} shows that if ${\cal M}_i(\texttt{aux},.)$ is $f$-DP for all $\texttt{aux}$, then the composed mechanism ${\cal M}$, which applies ${\cal M}_i$ in sequential order from $i=1$ to $i=T$, is $f^{\otimes T}$-DP.

\vspace{3mm}

\noindent
{\bf Tight Analysis DP-SGD:}
We are now able to formulate the differential privacy guarantee of DP-SGD since it is a composition of  subsampled Gaussian DP mechanisms. Theorem 5.1 in \citep{dong2021gaussian} states that DP-SGD in Algorithm \ref{alg:DPs} is\footnote{Their DP-SGD algorithm uses noise ${\cal N}(0,(2C)^2\sigma^2 {\bf I})$ compared to ${\cal N}(0,C^2\sigma^2 {\bf I})$ in our version of the DP-SGD algorithm. This is related to the earlier footnote on probabilistic versus constant sample sizes. The analysis in the $f$-DP framework normalizes with respect to $2C$, while we normalize with respect to $C$. The end result is that for fixed $\sigma$ the proven bounds in the $f$-DP framework also hold for this paper.  (Thus we do not need to compensate for a factor $2$ and use $\sigma/2$ in the $f$-DP framework in order to compare with our DP-SGD parameter setting.)}
$$ C_{s/N}(G_{\sigma^{-1}})^{\otimes T}\mbox{-DP}.$$
Since each of the theorems and results from \citep{dong2021gaussian} enumerated above are exact, we have a tight analysis.

\vspace{3mm}

\noindent
{\bf Our Goal:} We want to understand the behavior of the DP guarantee in terms of $s$, $N$, $T$, and $\sigma$. Our goal is to have an easy interpretation of the DP  guarantee  so that we can  select ``good"  parameters $s$, $N$, $T$, and $\sigma$ a-priori; good in terms of achieving at least our target accuracy without depleting our privacy budget. 
If we know how the differential privacy budget is being depleted over DP-SGD iterations, then we can optimize parameter settings in order to attain best performance, that is, best accuracy of the final global model (the most important target when we work with machine learning modelling). According to our best knowledge, all the current-state-of-the art privacy accountants do not allow us to achieve this goal because they are only
privacy loss accountants 
and do not offer  ahead-planning. 
It is not sufficient to only rely on a differential privacy accountant (see e.g., \citep{zhu2021optimal} as follow-up work of \citep{dong2021gaussian}) for a client to understand when to stop  helping the server to learn a global model. 

When talking about  accuracy, we mean how much loss in prediction/test accuracy is sacrificed by fixing a $\sigma$ (and clipping constant $C$). Our theory maps $\sigma$ directly to an $(\epsilon,\delta)$-DP guarantee independent of the number of rounds $T$. This allows use to characterize the trade-off between accuracy and privacy budget.
All the current-state-of-the art privacy loss frameworks do not offer this property.

We notice that \citep{dong2021gaussian} makes an effort to interpret the $C_{s/N}(G_{\sigma^{-1}})^{\otimes T}$-DP guarantee. Their Corollary 5.6 provides a precise expression based on integrals, themselves again depending on $p=s/N$ and $\mu=\sigma^{-1}$ in our notation. This still does not lead to the intuition we seek as we cannot extract how to select parameters $\sigma$, $s$ and $T$ given a data set of size $N$, given a privacy budget, and given a utility that we wish to achieve. We further explain this point in next paragraphs. 

In what follows, we 
seek a relationship between $\sigma$, $s$, $T$, $\epsilon$, $\delta$, and $N$ 
for Gaussian DP
based on Corollary 5.4 in \citep{dong2021gaussian}. Corollary 5.4 in \citep{dong2021gaussian} provides an \textbf{asymptotic analysis} which is a step forward to the kind of easy to understand interpretation we seek for: It states that if both $T\rightarrow \infty$ and $N\rightarrow \infty$ such that $s\sqrt{T}/N \rightarrow c$ for some constant $c>0$ (and where $s$ is a function of $N$ that may tend to $\infty$ as well), then the DP-SGD algorithm 
is $\mu$-Gaussian DP for
\begin{equation}
\mu =  c \cdot \tau^{-1} \mbox{ with } 
\tau^{-1}=\sqrt{2}\cdot \sqrt{e^{\sigma^{-2}} \Phi(3\sigma^{-1}/2) + 3 \Phi(-\sigma^{-1}/2) -2 }.
\label{asym}
\end{equation}

In Section \ref{app:GDPproof} we show that $\tau^{-1} = \sigma^{-1} +O(\sigma^{-2})$ and we show that for $\mu\leq \epsilon\leq 1$, $\mu$-Gaussian DP translates   to the DP-SGD algorithm being $(\epsilon,\delta)$-DP for $\delta \ll \epsilon \ll 1$ if
\begin{eqnarray*}
\tau &\approx& \frac{ (c/2)
\sqrt{2(\ln(1/\delta) + \ln(\epsilon) - O(\ln  \ln (1/\delta))) }}{\epsilon} \mbox{ with } s\sqrt{T}/N\rightarrow c.
\end{eqnarray*}

We see a similar $s\sqrt{T}/N$ dependency in Theorem  \ref{thm:abadi}  by \citep{abadi2016deep}.
The difference is that  Theorem  \ref{thm:abadi} holds in a \textbf{non-asymptotic} setting. That is, $T$ and $N$ do not need to tend to $\infty$ whereas the expression above does require taking such a limit. Of course, one can analyse the convergence rate of achieving the limit $\mu$ given $T$ and $N$ tending to infinity. When doing such an analysis one may find expressions of Gaussian DP guarantees  as a function of $T$ and $N$ that hold for all concrete values of $T$ and $N$. This may lead to results that strengthen our Theorem  \ref{thm:abadi} (we leave this as an open problem).
It is clear that the above asymptotic result is still insufficient for our purpose: How do we a-priori select concrete parameters $\sigma$, $s$, and $T$ given concrete parameters for $N$, a given privacy budget and utility that we wish to achieve?

In this paper we decided to generalize the proof method of Theorem  \ref{thm:abadi} rather than working with the complex integrals that provide the exact characterization of $f$-DP for the DP-SGD algorithm as stated above. This approach allows us to obtain the non-asymptotic 
result of Theorem \ref{thm:Main} which shows into large extent the independence of $T$, which is not immediately understood from Theorem  \ref{thm:abadi} and the corollary discussed above. 
Section \ref{app:tight} shows a first result on the tightness of  our Theorem \ref{thm:Main}. The advantage of our result is that it is easy to interpret and we do not need to fully rely on an accountant method to keep track of spent privacy budget while participating in learning a global model based on local data.



\subsection{\texorpdfstring{Translation to $(\epsilon,\delta)$-DP}{}} \label{app:GDPproof}

We first observe that by using $e^x=1+x +O(x^2)$ and $\Phi(x) = \frac{1}{2} + \frac{e^{-x^2/2}}{\sqrt{2\pi}}(x+O(x^3)) = \frac{1}{2} +\frac{x}{\sqrt{2\pi}}+O(x^3)$, 
a first order approximation of $\tau^{-1}$ is equal to $\sigma^{-1}+O(\sigma^{-2})$ (hence, $\tau^{-1}\approx \sigma^{-1}$  for large $\sigma$). 


For $x\geq 0$, we have the approximation
$$ \Phi(-x) = \frac{e^{-x^2/2}}{\sqrt{2 \pi}} \left(\frac{1}{x}-\frac{1}{x^3} + O(\frac{1}{x^5}) \right).
$$
Let $y\leq x$. Together with $-(x-y)^2/2= 2xy - (x+y)^2/2$ we derive
\begin{eqnarray*}
 \Phi(-x+y) - e^{2xy} \Phi(-x-y) 
&=&
\frac{e^{-(x-y)^2/2}}{\sqrt{2 \pi}} (\frac{1}{x-y}-\frac{1}{(x-y)^3} + O(\frac{1}{(x-y)^5}  ) 
\\ 
&&
-
\frac{e^{-(x-y)^2/2}}{\sqrt{2 \pi}} (\frac{1}{x+y}-\frac{1}{(x+y)^3} + O(\frac{1}{(x+y)^5}) )
\\ &=&
\frac{e^{-(x-y)^2/2}}{\sqrt{2 \pi}} (\frac{2y}{x^2-y^2}-\frac{6yx^2+2y^3}{(x^2-y^2)^3} + O(\frac{1}{x^5}) )
\\ &=&
\frac{e^{-x^2(1-y/x)^2/2}}{\sqrt{2 \pi}} (\frac{2y}{x^2(1-(y/x)^2)}+O(\frac{y}{x^4} +\frac{1}{x^5}) )
\end{eqnarray*}
If we assume 
$$ \mu \leq \epsilon \leq 1,$$
then $(\mu/2)/(\epsilon/\mu)=\mu^2/(2\epsilon) \leq \epsilon/2\leq 1$, $\epsilon/\mu \geq 1$,  and $\epsilon\leq 1$. We can use the above formulas and
approximate (\ref{eq:gdp}) as follows:
\begin{eqnarray*}
\delta &= &
\Phi(-\frac{\epsilon}{\mu}+\frac{\mu}{2}) - e^{\epsilon} \Phi(-\frac{\epsilon}{\mu}-\frac{\mu}{2})
\\
&=&
\frac{e^{-(\epsilon/\mu)^2(1-\mu^2/(2\epsilon))^2/2}}{\sqrt{2 \pi}} \left(\frac{\mu^3}{\epsilon^2(1-(\mu^2/(2\epsilon))^2)}+O(\frac{\mu^5}{\epsilon^5}) \right).
\end{eqnarray*}
This gives
$$
1/\mu = 
\frac{\sqrt{2(\ln(1/\delta) + \ln(\frac{\mu^3}{\sqrt{2\pi}(\epsilon^2-\mu^4/4)}+O((\frac{\mu}{\epsilon})^5)))}}{\epsilon-\mu^2/2},
$$
hence,
$$
\tau = 
\frac{(c/2)\sqrt{2(\ln(1/\delta) + \ln(\frac{\mu^3}{\sqrt{2\pi}(\epsilon^2-\mu^4/4)}+O((\frac{\mu}{\epsilon})^5)))}}{\epsilon-\mu^2/2}
%
%
\mbox{ with } s\sqrt{T}/N\rightarrow c \mbox{ and } \tau^{-1} = \sigma^{-1} +O(\sigma^{-2}).
$$
For small $\epsilon$, we can approximate $\tau$ as
$$
\tau \approx \frac{ c \cdot
\sqrt{2(\ln(1/\delta) + \ln(\frac{\epsilon}{\sqrt{2\pi}}(\frac{\mu}{\epsilon})^3+O((\frac{\mu}{\epsilon})^5)))}}{\epsilon}.
$$
Since $c\tau^{-1}=\mu \leq   \epsilon$, we may write
$\mu = \epsilon/b$ for some $b\geq 1$. Notice that $c/\epsilon =\tau /b$. This leads to the approximation
$$
b \approx 
\sqrt{2(\ln(1/\delta) + \ln(\frac{\epsilon}{\sqrt{2\pi}}\frac{1}{b^3}+O(\frac{1}{b^5})))}.
$$
For a concrete choice of $N$, we select in practice by default $\delta=1/N$ and
$N=\Omega(1/\epsilon)$, that is, $\epsilon=\Omega(1/N)$ (also notice that good accuracy can only be achieved for $\sigma$  small enough, that is, $\epsilon$ is generally orders of magnitude larger than $1/N$). For this reason, we assume $\epsilon \gg \delta$ (we fix $\epsilon$ and $\delta$, i.e., we do not choose $\delta$ as a function of $N$, after which take the limit $N\rightarrow \infty$; notice that this is only a theoretical analysis in an attempt to understand the relationship between various parameters).
Then substituting $\sqrt{2 \ln (1/\delta)}$ for $b$ at the right hand side yields
$$
b \approx 
\sqrt{2(\ln(1/\delta) + \ln(\epsilon) - O(\ln  \ln (1/\delta)))}.
$$
Substituting back in the expression for $\tau$ proves for $\delta\ll \epsilon\ll 1$,
$$
\tau \approx \frac{c
\sqrt{2(\ln(1/\delta) + \ln(\epsilon) - O(\ln  \ln (1/\delta))) }}{\epsilon} \mbox{ with } s\sqrt{T}/N\rightarrow c \mbox{ and } \tau^{-1} = \sigma^{-1} +O(\sigma^{-2}).
$$

\subsection{Asymptotic Tightness of Theorem \ref{thm:Main}}
\label{app:tight}

We consider the special case where $T$ meets its lower bound: $T=\frac{\gamma \theta^2}{\epsilon} \cdot k^2$ (notice that our experiments and understanding show that this is a good setting for best accuracy). 
Let 
$$ c= \sqrt{\frac{\epsilon}{\gamma \theta^2}}= \frac{k}{\sqrt{T}}= \frac{K}{\sqrt{T} N} = \frac{s \sqrt{T}}{N},$$
where we consider a constant step size $s$ (hence, $\theta=1$).
Substituting this in the final formula for $\tau$ of Section \ref{app:GDPproof} yields for $\epsilon\gg \delta$ 
$$
\tau \approx \sqrt{\frac{ 
\ln(1/\delta) + \ln(\epsilon) - O(\ln  \ln (1/\delta)) }{\gamma  \epsilon}}.
$$
Here, $\tau\approx \sigma$ for $\sigma\gg 1$ and we see that compared to Theorem \ref{thm:Main} this formula attains a factor $\sqrt{2\gamma}\approx 2$ smaller $\sigma$ for the same $(\epsilon,\delta)$.
This shows that in this asymptotic setting for $N\rightarrow \infty$, $T\rightarrow \infty$, $\epsilon\gg \delta$, and $\sigma\gg 1$, Theorem \ref{thm:Main} is up to a factor $\sqrt{2\gamma}$ tight. In other words, the formula in Theorem \ref{thm:Main} pays a factor in tightness in order to hold for general concrete parameter settings (not just the asymptotic setting). 

The factor $\sqrt{2\gamma}$ seems large. However, if the lower bound on $T$ can be tightened to $T\geq \frac{\theta^2}{2\epsilon} \cdot k^2$, then the constant $c$ above can be increased to $c=\sqrt{2\epsilon/\theta^2}$ leading to a formula for $\tau$ that matches the formula for $\sigma$ in Theorem \ref{thm:Main} implying that it is tight in the asymptotic setting. In other words, rather than expecting to be able to lower the constant $2$ in $\sigma= \sqrt{2(\epsilon+\ln(1/\delta))/\epsilon}$, we can focus on 
how much the lower bound of $T$ can be reduced by a small factor (notice that the derivations in Theorem \ref{thm:simpleF} that lead to this lower bound may be tightened up). We expect the bound $\sigma\geq  \sqrt{2(\epsilon+\ln(1/\delta))/\epsilon}$ to be quite tight, while the lower bound $T\geq T_{min}=\frac{\gamma \theta^2}{\epsilon} \cdot k^2$ can at most be reduced by a factor $2\gamma$ to $T\geq T_{min\_asym}=\frac{\theta^2}{2\epsilon} \cdot k^2$. For smaller $T<T_{min\_asym}$, the asymptotic setting for $N\rightarrow \infty$, $T\rightarrow \infty$, $\epsilon\gg \delta$, and $\sigma\gg 1$ contradicts the tightness of the asymptotic result (\ref{asym}) of the $f$-DP framework. This proves Theorem \ref{thm:tight}.
We notice that this is also (unsurprisingly) confirmed by experiments in Appendix \ref{app:DPaccSim} where we implement the $f$-DP accountant in order to compute which $\epsilon$ is actually being achieved.

\vspace{3mm}

\noindent 
\textbf{Remark on Choosing a Larger $T$:}
Theorem  \ref{thm:tight} shows that in 
Theorem \ref{thm:Main} $T$'s required lower bound $\gamma \theta^2 k^2/\epsilon$ cannot be made smaller by more than a constant factor $2\gamma \approx 4$ (otherwise, this conflicts with an asymptotical result proved by the $f$-DP framework). This shows that choosing $T$ equal to the lower bound $\gamma \theta^2 k^2/\epsilon$ is {\em close to tight} in order to achieve $(\epsilon,\delta)$-DP. 

Of course, larger $T$ also satisfy  lower bound (\ref{minK})  implying the same $(\epsilon,\delta)$-DP guarantee.
We notice that a larger $T$ can meet $\gamma \theta^2 \cdot k^2/\epsilon'$ for a smaller $\epsilon'$ leading to a close to tight $(\epsilon',\delta)$-DP guarantee if we choose a larger  $\sigma$, 
which can be done 
if this still leads to sufficient accuracy.
Intuitively,  a $T$ larger than the lower bound $\gamma \theta^2 k^2/\epsilon$ invests in the potential of  improved differential privacy  (i.e., $\epsilon'\leq \epsilon$) which we do not need if we only require the $(\epsilon,\delta)$-DP guarantee. Better is to sacrifice this potential and meet the lower bound so that accuracy of the final global model is optimized.
Since the lower bound can at most be a constant factor $2\gamma \approx 4$ smaller, we cannot improve the accuracy much more by reducing $T$ further without violating the $(\epsilon,\delta)$-DP guarantee\footnote{Notice that, especially for non-convex problems, $T$ should also not be too small since we want a large enough number of rounds for updating the global model regulary for convergence.}.

\subsection{\texorpdfstring{Interpreting Theorem \ref{thm:Main} in the $f$-DP Framework}{}}
\label{app:GDP-T}

As a theoretical consequence (side result), for fixed $k$, 
we formulate  $(\epsilon,\delta)$-DP guarantees for varying $\epsilon$ which can be used to show  $f$-DP for a (non-trivial) trade-off function $f$ that depends on a target $\epsilon$ and $\delta$ but does not depend on the choice of $T$ in the range 
$\gamma \theta^2 k^2/\epsilon\leq T\leq K=kN$ (where $sT=K=kN$): 

In the $f$-DP framework, if a mechanism is $f$-DP, then it is $(\epsilon,\delta)$-DP for all $(\epsilon,\delta)$ for which $f\geq f_{\epsilon,\delta}$. When considering DP-SGD for parameters $\sigma$ and $T$ and the other hyper parameters fixed, there exists some function $h_{\sigma,T}$ such that DP-SGD is $f$-DP if and only if $h_{\sigma,T}\geq f$. This function implies $(\epsilon,\delta)$-DP for all $h_{\sigma,T}\geq f_{\epsilon,\delta}$.
Conversely, if we want to realize $(\epsilon,\delta)$-DP for some target privacy budget defined by $(\epsilon,\delta)$, then we need to choose $\sigma$ and $T$ such that $h_{\sigma,T}\geq f_{\epsilon,\delta}$. 

We notice that Theorem \ref{thm:Main} can be cast in the $f$-DP framework as it allows us to formulate an appropriate (non-tight) $\hat{h}_{\sigma,T}$  for which DP-SGD is $\hat{h}_{\sigma,T}$-DP as follows: We first notice that by the definition of $h_{\sigma,T}$ we have
$$ h_{\sigma,T}\geq \hat{h}_{\sigma,T}.
$$
By fixing $\sigma$ and $T$, we may freely choose $(\epsilon,\delta)$ 
as long as the conditions of Theorem \ref{thm:Main} are satisfied: $(\epsilon,\delta)\in {\cal H}(\sigma,T)$ with
$$
{\cal H}(\sigma, T)
=
\left\{
(\epsilon,\delta) \ : \ 
\begin{array}{cc}
 \sigma \geq \sqrt{2(\epsilon+\ln(1/\delta))/\epsilon}      \\
 \delta\leq 1/N      \\
 \epsilon<0.5  \\
 k\geq \sqrt{2\epsilon \ln(1/\delta)}\cdot e/(\gamma(\sigma,\epsilon,k)\cdot \theta) \\
T\geq \gamma(\sigma,\epsilon,k)\cdot \theta^2 k^2/\epsilon 
\end{array}
\right\},
$$
 where $\gamma$ is a function of $\sigma$, $\epsilon$ and $k=K/N$. Hyper parameters $K$ and $N$ (and, therefore also $k$) are fixed.
For a given $T$, we consider a constant sample size $s=K/T$ from round to round, hence, we use $\theta=1$ in the definition of ${\cal H}(\sigma,T)$.
By defining 
$$
\hat{h}_{\sigma,T}(\alpha) = \sup_{(\epsilon,\delta) \in {\cal H}(\sigma,T)}
 f_{\epsilon,\delta}(\alpha),
$$
we have that DP-SGD is $\hat{h}_{\sigma,T}$-DP.

%
%
%
%
%
%
%

Now consider a target epsilon $\epsilon=\epsilon_{target}$ and define a new set
$$
{\cal G}(\sigma, \epsilon_{target})
=
\left\{
(\epsilon,\delta) \ : \ 
\begin{array}{cc}
 \sigma \geq \sqrt{2(\epsilon+\ln(1/\delta))/\epsilon}      \\
 \delta\leq 1/N      \\
 \epsilon<0.5  \\
 k\geq \sqrt{2\epsilon \ln(1/\delta)}\cdot e/(\gamma(\sigma,\epsilon,k)\cdot \theta) \\
\epsilon \geq \epsilon_{target}
\end{array}
\right\}.
$$
%
Notice that ${\cal G}(\sigma,\epsilon_{target})$ is independent of $T$, but does satisfy the property
$$ {\cal G}(\sigma,\epsilon_{target})\subseteq {\cal H}(\sigma, T) \mbox{ for } 
T\geq \gamma \theta^2 k^2/\epsilon_{target}.$$
Therefore,
$$
g_{\sigma,\epsilon_{low}}(\alpha) = \sup_{(\epsilon,\delta) \in {\cal G}(\sigma,\epsilon_{target})}
 f_{\epsilon,\delta}(\alpha),
$$
is also independent of $T$ while it satisfies
$$ \hat{h}_{\sigma,T} \geq g_{\sigma,\epsilon_{target}} \mbox{ for } 
T\geq \gamma \theta^2 k^2/\epsilon_{target}.
$$
Together with $h_{\sigma,T} \geq \hat{h}_{\sigma,T}$ we conclude
$$h_{\sigma,T} \geq g_{\sigma,\epsilon_{target}} \mbox{ for } 
T\geq \gamma \theta^2 k^2/\epsilon_{target}.
$$



This shows how Theorem \ref{thm:Main} translates to the $f$-DP framework in that as long as the number of rounds is large enough, that is, $K\geq T\geq \gamma \theta^2 k^2/\epsilon_{target}$, we have that there always remains some $f$-DP privacy guarantee given by $g_{\sigma,\epsilon_{target}}$. Notice that $g_{\sigma,\epsilon_{target}}$ is independent of $T$. So, even in the limit for larger $T$, not all of the privacy budget is being depleted. For larger $T$, $f_{\sigma,T}$ 
remains at least $g_{\sigma,\epsilon_{target}}$. 
Since the $f$-DP framework and its resulting DP accountant provide a tight analysis, this shows that an increasing number $T$ of rounds revealing more and more local updates does not linearly increase the privacy leakage! Instead, a larger $T$ gives rise to more updates that each leak privacy and gives rise to subsampling of smaller mini-batches which amplifies the differential privacy guarantee more, hence, less leakage per round. The total resulting leakage remains bounded in that we can always guarantee $g_{\sigma,\epsilon_{target}}$-DP (even for larger $T$).
This implies that the tight $f$-DP based privacy accountant will have a limit (upper bound) on its reported privacy leakage for increasing $T$ ($\leq K$).

\section{Experiments}
\label{app:experiment}


We provide  experiments to support our theoretical findings, i.e.,  convergence  of our proposed asynchronous distributed learning framework with differential privacy (DP)  to a sufficiently accurate solution. 
We cover strongly convex, plain convex and non-convex  objective functions over  iid local data sets.

We introduce
our experimental set up in Section~\ref{subsec:hyperparameter}. Section~\ref{subsec:utility_graph} provides utility graphs for different data sets and objective functions. A utility graph  helps choosing the maximum possible noise $\sigma$, in relation to the value of
the clipping constant $C$, for which decent accuracy can be achieved.
Section~\ref{subsec:exp_asynDP} provides  detailed experiments for our asynchronous differential privacy SGD framework (asynchronous DP-SGD) with different types of objective functions (i.e., strongly convex, plain convex and non-convex objective functions), different types of constant sample size sequences and different levels of privacy guarantees (i.e., different privacy budgets $\epsilon$). 

\setcounter{footnote}{0} 

All our experiments are conducted on LIBSVM~\citep{CC01a}\footnote{https://www.csie.ntu.edu.tw/~cjlin/libsvmtools/datasets/binary.html}
, MNIST~\citep{lecun-mnisthandwrittendigit-2010} 
\footnote{http://yann.lecun.com/exdb/mnist/}, and CIFAR$10$ \footnote{https://www.cs.toronto.edu/~kriz/cifar.html} data sets.

\subsection{Experiment settings}
\label{subsec:hyperparameter}

\textbf{Simulation environment.} For simulating the asynchronous DP-SGD framework, we use multiple threads where each thread represents one compute node joining the training process. The experiments are conducted on Linux-64bit OS, with $16$ cpu processors, and 32Gb RAM.

\textbf{Objective functions.} Equation~($\ref{eq_logstic_reg}$) defines the plain convex logistic regression problem. The weight vector $w$ and the bias value $b$ of the logistic function can be learned by minimizing the log-likelihood function $J$:
\begin{equation} 
\label{eq_logstic_reg}
    J = - \sum_{i=1}^N [ y_{i} \cdot \log (\bar{\sigma}_i) + (1 - y_{i}) \cdot \log (1 - \bar{\sigma}_i) ], \text{ (plain convex)}
\end{equation}
where $N$ is the number of training samples $(x_i,y_i)$ with $y_i\in\{0,1\}$ and $\bar{\sigma}_i$
is defined by
\begin{equation} \nonumber 
    \bar{\sigma}_i= \frac{1}{1 + e^{-(w^{\mathrm{T}}x_i + b)}},
\end{equation}
which is the sigmoid function with parameters $w$ and $b$.
Our goal is to learn a vector $w^*$ which represents a pair $\bar{w}=(w, b)$  that minimizes $J$. 

Function $J$ can be changed into a strongly convex problem $\hat{J}$ by adding a regularization parameter $\lambda>0$:
\begin{equation} \nonumber 
    \hat{J} = - \sum_{i=1}^N [ y_{i} \cdot \log (\sigma_i) + (1 - y_{i}) \cdot \log (1 - \sigma_i) ] + \frac{\lambda}{2}\norm{w}^{2}, \text{ (strongly convex).}
\end{equation}
where $\bar{w} = (w, b)$ is vector $w$ concatenated with bias value $b$. In practice, the regularization parameter $\lambda$ is set to $1/N$ \citep{roux2012stochastic}.

For simulating non-convex problems, 
we choose a simple neural network (LeNet)~\citep{lecun1998gradient}
for MNIST data set and AlexNet~\citep{krizhevsky2012imagenet} for CIFAR$10$ data set with cross entropy loss function for image classification.

The loss functions for the strong, plain, and non-convex problems represent the objective function $F(.)$.


\textbf{Parameter selection.} The parameters used for our distributed algorithm with Gaussian based differential privacy for strongly convex, plain convex and non-convex objective functions are described in Table~\ref{tbl:tbl_async_DP_paramter}. The clipping constant $C$ is set to $0.1$ for strongly convex and plain convex problems and $0.025$ for non-convex problem (this turns out to provide good utility).

\begin{table}[!ht]
\caption{Common parameters of asynchronous DP-SGD framework with differential privacy}
\label{tbl:tbl_async_DP_paramter}
\begin{center}
\begin{small}
\scalebox{0.999}{
\begin{threeparttable}

\begin{tabular}{|l|c|c|c|c|c|c|}
\hline
                & \# of clients $n$ & Diminishing step size $\bar{\eta_t}$ & Regular $\lambda$  & Clipping constant $C$  
               \\ 
               \hline
               \hline
Strongly convex & 5 & $\frac{\eta_0}{1 + \beta {t}} \tnote{\ddag}$ & $\frac{1}{N}$ & 0.1                   
\\ 
\hline
Plain convex & 5 & $\frac{\eta_0}{1 + \beta {t}}$ or $\frac{\eta_0}{1 + \beta \sqrt{t}}$ & $N/A$ & 0.1                                   
\\ 
\hline
Non-convex & 5 & $\frac{\eta_0}{1 + \beta \sqrt{t}}$ & $N/A$ & 0.025
\\ 
\hline
\end{tabular}
    \begin{tablenotes}
       \item {\footnotesize $\ddag$ The $i$-th round step size $\bar{\eta}_i$ is computed by substituting  $t=\sum_{j=0}^{i-1} s_j$ into the diminishing step size formula}.
   \end{tablenotes}
   
\end{threeparttable}
}
\end{small}
\end{center}
\vskip -0.1in
\end{table}

For the plain convex case, we can use  diminishing step size schemes $\frac{\eta_0}{1 + \beta \cdot t}$ or $\frac{\eta_0}{1 + \beta \cdot \sqrt{t}}$. In this paper, we focus our experiments for the plain convex case on $\frac{\eta_0}{1 + \beta \cdot \sqrt{t}}$. Here, $\eta_0$ is the initial step size and we perform a systematic grid search on parameter $\beta=0.001$ for strongly convex case and $\beta=0.01$ for both plain convex and non-convex cases.
Moreover, most of the experiments are conducted with $5$ compute nodes and $1$ central server.
When we talk about accuracy (from Figure~\ref{fig:asyncDP_phishing_sampling_method} and onward), we mean test accuracy defined as  the fraction of samples from a test data set that get accurately labeled by the classifier (as a result of training on a training data set by minimizing a corresponding objective function).


\subsection{Utility graph}
\label{subsec:utility_graph}

The purpose of a utility graph is to help us choose, given the value of
the clipping constant $C$,  the maximum possible noise $\sigma$  for which decent accuracy can be achieved.
A utility graph depicts the test accuracy of model $F(w^{*} + n)$ over $F(w^{*})$, denotes as accuracy fraction, where $w^*$ is a near optimal global model and  $n\sim {\cal N}(0,C^2\sigma^2 {\bf I})$ is Gaussian noise. This shows which maximum $\sigma$ can be chosen with respect to allowed loss in expected test accuracy, clipping constant $C$ and standard deviation $\sigma$. 



\begin{figure}[!htb]
  \centering
  \subfigure[Strong convex.]{\includegraphics[width=0.5\textwidth]{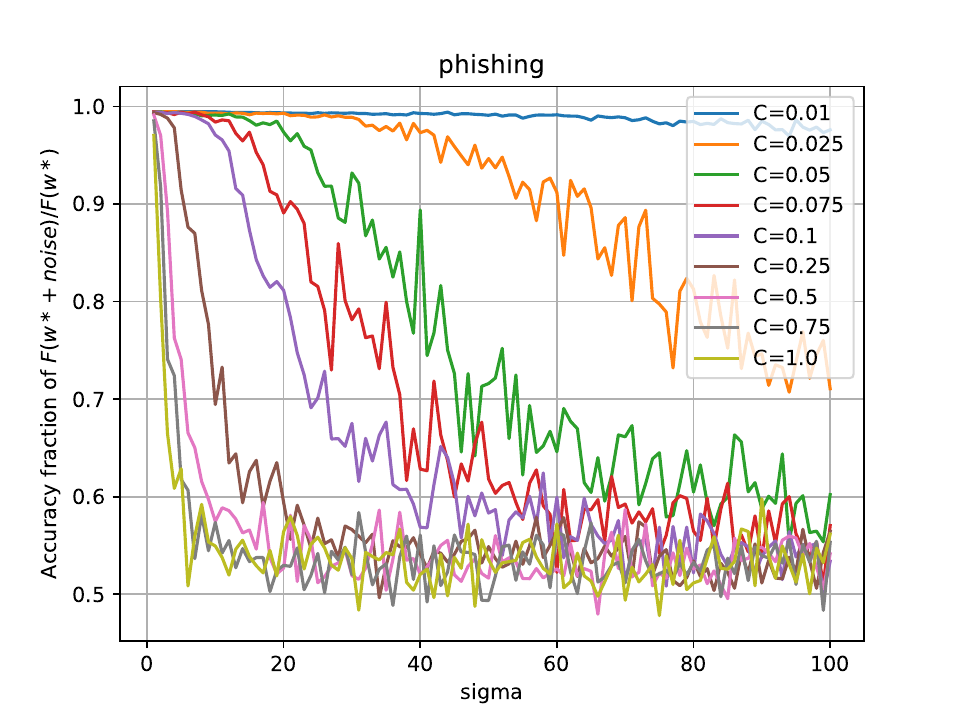}\label{fig:asyncDP_phishing_utility_0}}
  \hfill
  \subfigure[Plain convex.]{\includegraphics[width=0.5\textwidth]{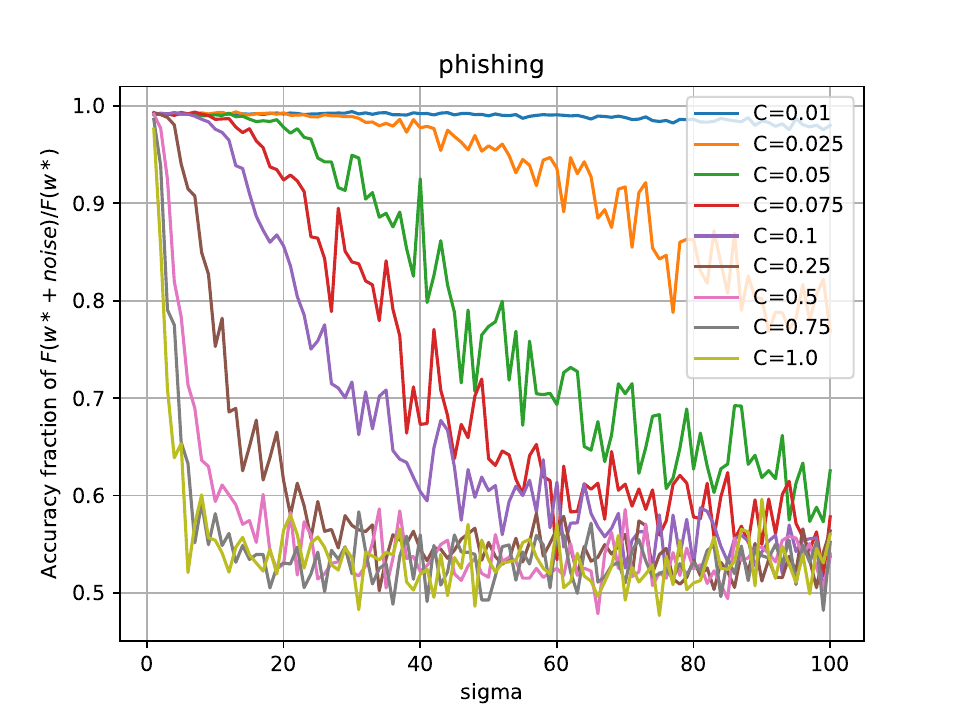}\label{fig:asyncDP_phishing_utility_1}}
  \caption{Utility graph with various gradient norm $C$ and noise level $\sigma$}
  \label{fig:asyncDP_phishing_utility}
\end{figure}

\begin{figure}[!htb]
  \centering
  \subfigure[Strong convex.]{\includegraphics[width=0.5\textwidth]{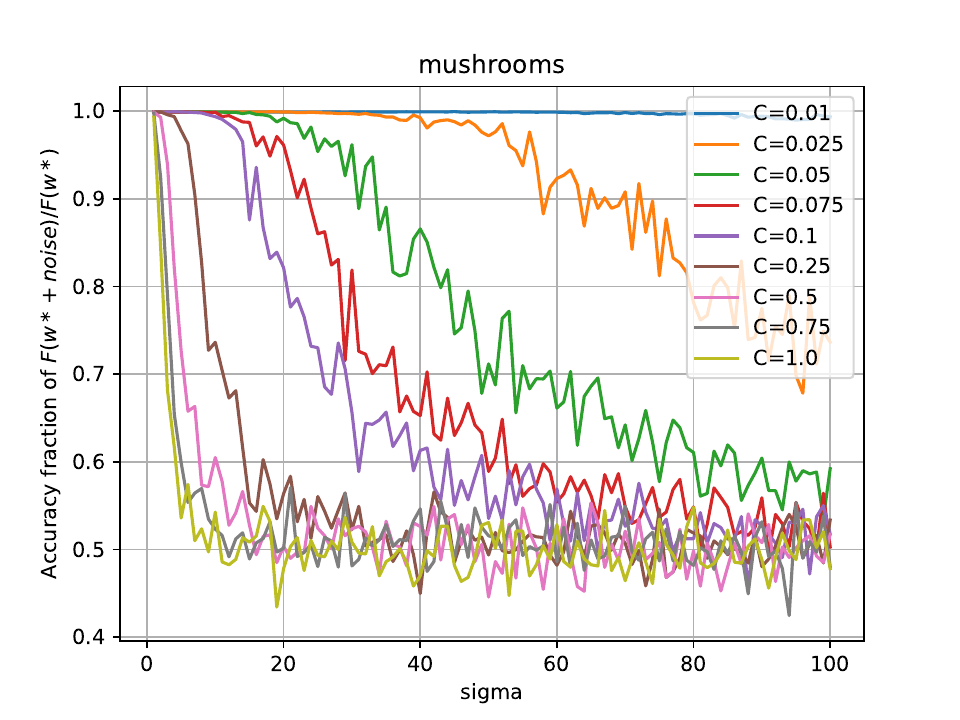}\label{fig:asyncDP_mushrooms_utility_0}}
  \hfill
  \subfigure[Plain convex.]{\includegraphics[width=0.5\textwidth]{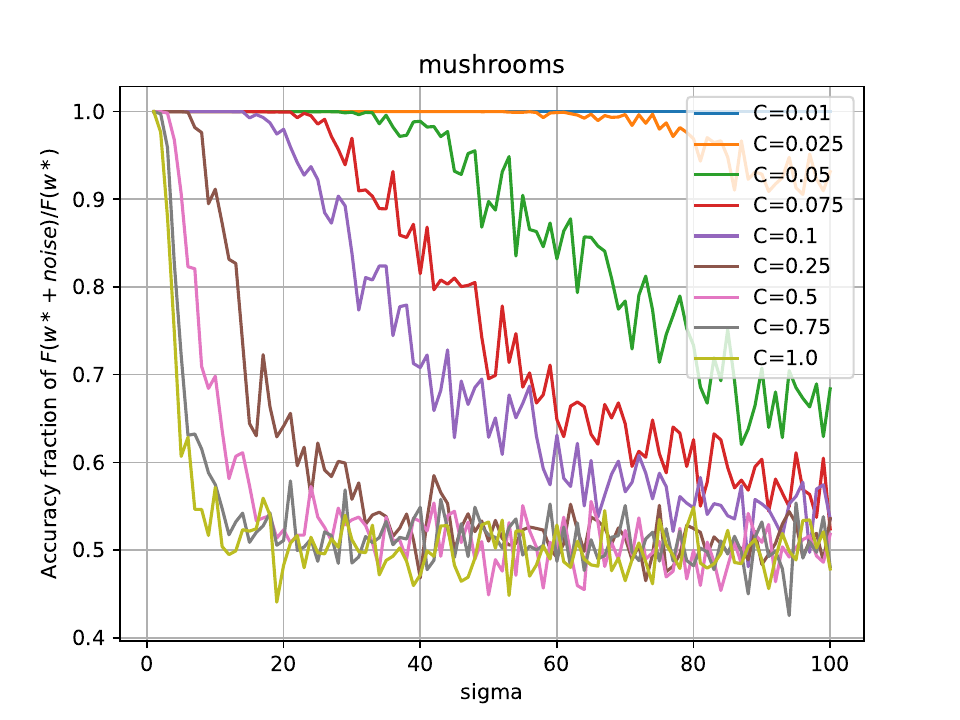}\label{fig:asyncDP_mushrooms_utility_1}}
  \caption{Utility graph with various gradient norm $C$ and noise level $\sigma$}
  \label{fig:asyncDP_mushrooms_utility}
\end{figure}



\begin{figure}[!ht]
\captionsetup[subfigure]{labelformat=empty}
  \centering
  \subfigure[]{\includegraphics[width=0.5\textwidth]{Experiments/Utility_graph/sensitivity_mnist_non_convex.pdf}\label{fig:asyncDP_mnist_utility_0}}
  \hfill
 \subfigure[]{\includegraphics[width=0.5\textwidth]{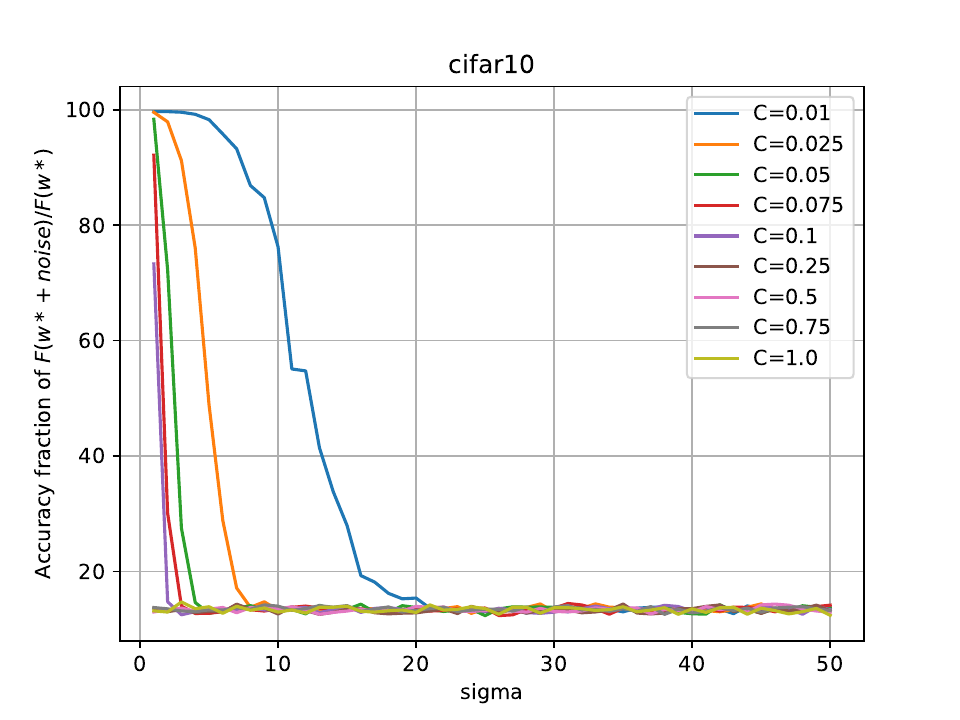}\label{fig:asyncDP_cifar10_utility_0}}
  \caption{Utility graph with various gradient norm $C$ and noise level $\sigma$ for MNIST and CIFAR$10$ data sets.}
  \label{fig:asyncDP_mnist_utility}
\end{figure}

As can be seen from Figure~\ref{fig:asyncDP_phishing_utility} and Figure~\ref{fig:asyncDP_mushrooms_utility}, for clipping constant $C=0.1$, we can choose the maximum $\sigma$ somewhere in the range $\sigma \in \left[18, 22 \right]$ if we want to guarantee there is at most about 10\% accuracy loss
compared to the (near)-optimal solution without noise. Another option is $C=0.075$, where we can tolerate  $\sigma \in \left[18, 30 \right]$ yielding the same accuracy loss guarantee. 
When the gradient bound $C$ gets smaller, our DP-SGD can tolerate bigger noise, i.e, bigger values of $\sigma$. However, we need to increase the number $K$ of iterations during the training process when $C$ is smaller in order to converge and gain a specific test accuracy -- this is the trade-off. For simplicity, we intentionally choose $C=0.1$, $\sigma \leq 20$ and expected test accuracy loss about 10\% for our experiments with strongly convex and plain convex objective functions. 

The utility graph is extended to the non-convex objective function in Figure~\ref{fig:asyncDP_mnist_utility}. To keep the test accuracy loss less or equal to 10\% (of the final test accuracy of the original model $w^{*}$), we choose  $C=0.025$ and  noise level $\sigma \leq 12$ for MNIST data set (as shown in Figure~\ref{fig:asyncDP_mnist_utility_0}) and $C=0.025$ and  noise level $\sigma \leq 6.572$ for CIFAR$10$ data set (as shown in Figure~\ref{fig:asyncDP_cifar10_utility_0}). 
For simplicity, we use this parameter setting for our experiments with the non-convex problem.


\subsection{Asynchronous distributed learning with differential privacy}
\label{subsec:exp_asynDP}

We consider the asynchronous DP-SGD framework with strongly convex, plain convex and non-convex objective functions for different settings, i.e., different levels of privacy budget $\epsilon$ and  different constant sample size sequences.

\subsubsection{Asynchronous DP-SGD with different constant sample size sequences}
\label{subsec:sample_size_sequence}

The purpose of this experiment is to investigate which is the best constant sample size sequence $s_i = s$.
This experiment allows us to choose a decent sample size sequence that will be used in our subsequent experiments. To make the analysis simple, we consider our asynchronous DP-SGD framework with
 $\Upsilon(k,i)$ defined as false if and only if $k< i-1$, i.e., compute nodes are allowed to run fast and/or have small communication latency such that  
 broadcast global models are at most $1$ local round in time behind (so different clients can be asynchronous with respect to one another for 1 local round). We also use iid data sets. The detailed parameters are in Table~\ref{tbl:basic_param_setting}.

\begin{table}[!htb]
\caption{Basic parameter setting for strongly and plain convex problems}
\label{tbl:basic_param_setting}
\begin{center}
\begin{threeparttable}
\begin{tabular}{|c|c|c|}
\hline
Parameter & Value & Note \\ \hline \hline
$\bar{\eta}_0$          & $0.1$     &     initial stepsize
\\ \hline


$N_c$ &    $10,000$  & \# of data points
\\ \hline
$K$ &    $50,000$  & \# of iterations
\\ \hline
$\epsilon$ &    $0.04945$  & 
\\ \hline
$\sigma$ & $19.29962$ &
\\ \hline
$\delta$ & $0.0001$ &
\\ \hline
$C$ & $0.1$ & clipping constant
\\ \hline
$s$ & $\{1, 5, 10, 15, 20, 26 \}$ & constant sample size sequence
\\ \hline
dataset & LIBSVM & iid dataset
\\ \hline
$n$ & $5$ &  \# of nodes
\\ \hline
$\Upsilon$ & $k\geq i-1$ &  $1-$asynchronous round
\\ \hline

\end{tabular}
  \end{threeparttable}
\end{center}
\end{table}

\begin{figure}[!ht]
  \centering
  \subfigure[Strong convex.]{\includegraphics[width=0.5\textwidth]{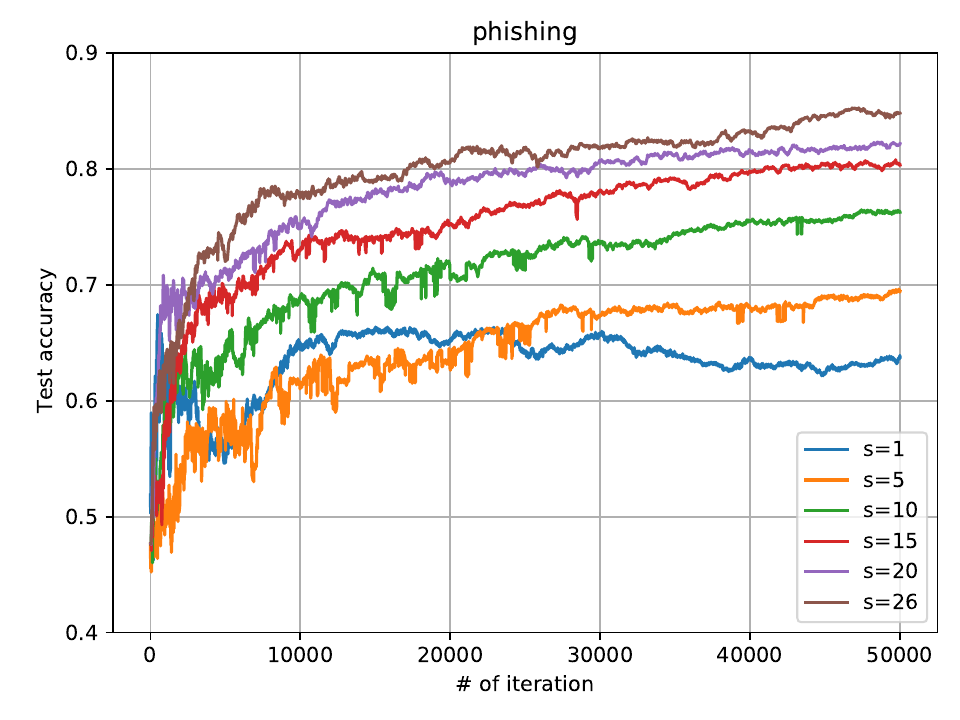}\label{fig:asyncDP_phishing_sample_0}}
  \hfill
  \subfigure[Plain convex.]{\includegraphics[width=0.5\textwidth]{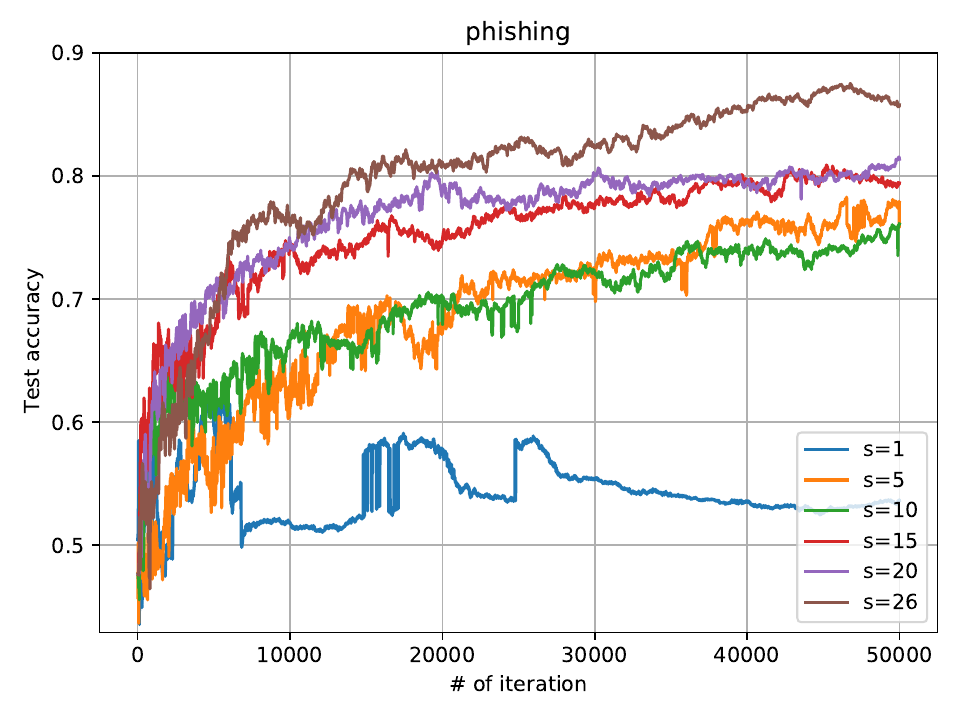}\label{fig:asyncDP_phishing_sample_1}}
  \caption{Effect of different constant sample size sequences}
  \label{fig:asyncDP_phishing_sampling_method}
\end{figure}

\begin{figure}[!ht]
  \centering
  \subfigure[Strong convex.]{\includegraphics[width=0.5\textwidth]{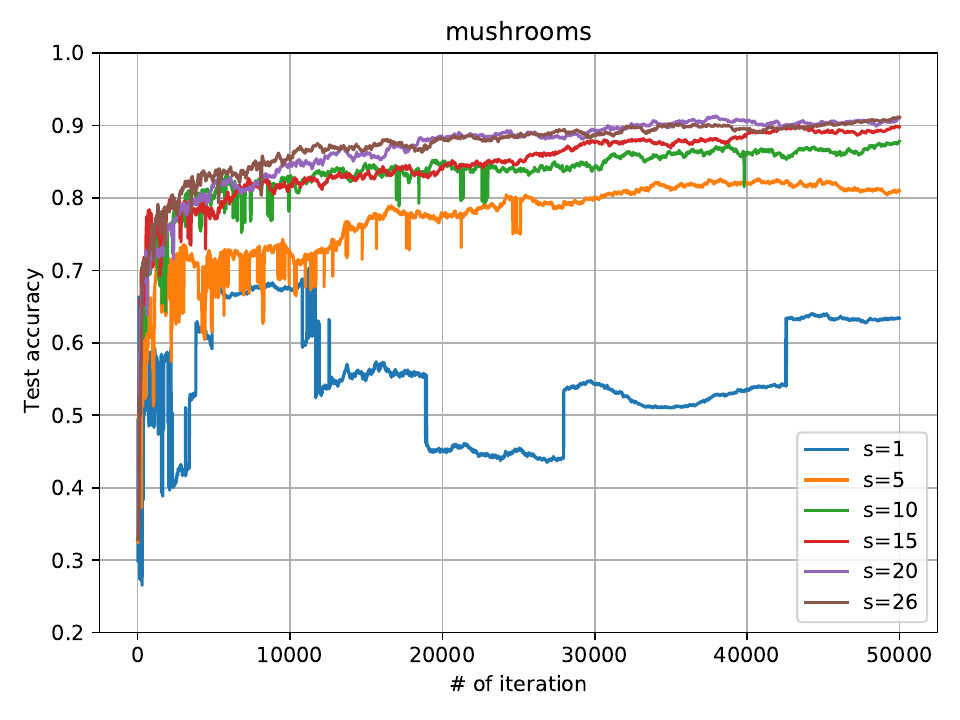}\label{fig:asyncDP_mushrooms_sample_0}}
  \hfill
  \subfigure[Plain convex.]{\includegraphics[width=0.5\textwidth]{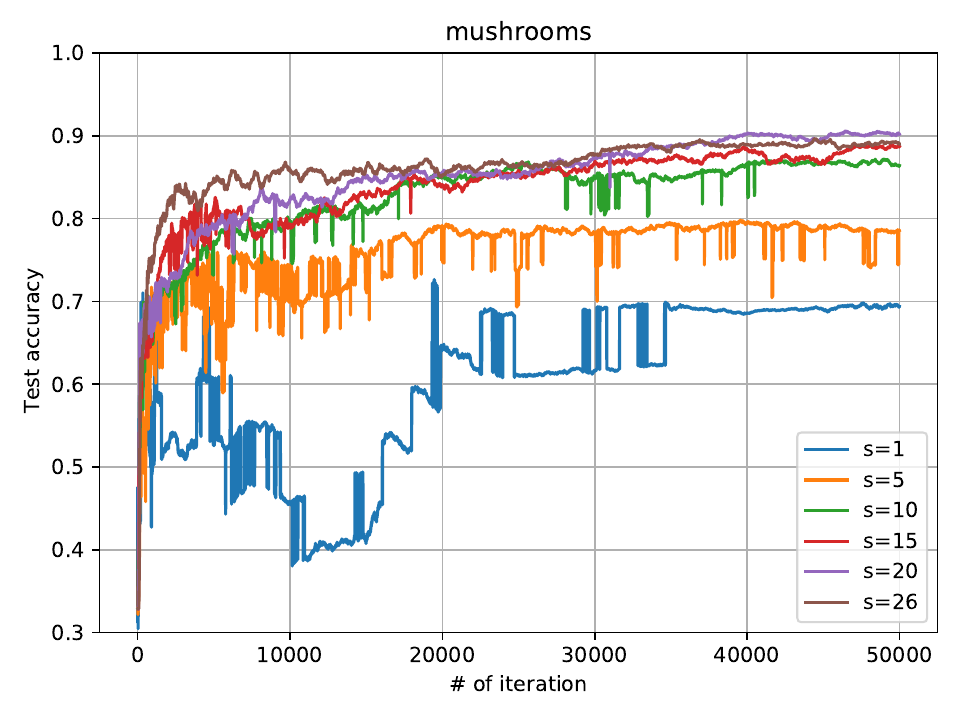}\label{fig:asyncDP_mushrooms_sample_1}}
  \caption{Effect of different constant sample size sequences}
  \label{fig:asyncDP_mushrooms_sampling_method}
\end{figure}

The results from Figure~\ref{fig:asyncDP_phishing_sampling_method} to Figure~\ref{fig:asyncDP_mushrooms_sampling_method} confirm that our asynchronous DP-SGD framework can converge under a very small privacy budget. When the constant sample size $s=1$, it is clear that the DP-SGD algorithm does not achieve  good accuracy compared to other constant sample sizes even though this setting has the maximum number of communication rounds. When we choose constant sample size $s=26$ (this meets the upper bound for constant sample sizes for our small $N=10,000$ and small $\epsilon \approx 0.05$, see Theorem~\ref{thm:simpleF}), our DP-SGD framework converges to a decent test accuracy, i.e, the test accuracy loss is expected less than or equal to 10\% when compared to the original mini-batch SGD without noise. In conclusion, this experiment demonstrates that our asynchronous DP-SGD with diminishing step size scheme and constant sample size sequence works well under DP setting, i.e, our asynchronous DP-SGD framework can gain differential privacy guarantees while maintaining an acceptable accuracy.

\begin{table}[!htb]
\caption{Basic parameter setting for non-convex problem with MNIST data set}
\label{tbl:basic_param_setting_noncvx}
\begin{center}
\begin{threeparttable}
\begin{tabular}{|c|c|c|}
\hline
Parameter & Value & Note \\ \hline \hline
$\bar{\eta}_0$          & $0.1$     &     initial stepsize
\\ \hline
$N_c$ &    $60,000$  & \# of data points
\\ \hline
$K$ &    $360,000$  & \# of iterations
\\ \hline
$\epsilon$ &    $0.15007$  & 
\\ \hline
$\sigma$ & $12.10881$ &
\\ \hline
$\delta$ & $1.667 \cdot 10^{-5}$ &
\\ \hline
$C$ & $0.025$ & clipping constant
\\ \hline
$s$ & $\{ 10, 25, 50, 100, 200, 300, 370 \}$ & constant sample size sequence
\\ \hline
dataset & MNIST & iid dataset
\\ \hline
$n$ & $5$ &  \# of nodes
\\ \hline
$\Upsilon$ & $k\geq i-1$ &  $1-$asynchronous round
\\ \hline

\end{tabular}
  \end{threeparttable}
\end{center}
\end{table}


\begin{table}[!htb]
\caption{Basic parameter setting for non-convex problem for CIFAR$10$ data set}
\label{tbl:basic_param_setting_noncvx_cifar10}
\begin{center}
\begin{threeparttable}
\begin{tabular}{|c|c|c|}
\hline
Parameter & Value & Note \\ \hline \hline
$\bar{\eta}_0$          & $0.1$     &     initial stepsize
\\ \hline
$N_c$ &    $50,000$  & \# of data points
\\ \hline
$K$ &    $350,000$  & \# of iterations
\\ \hline
$\epsilon$ &    $0.50102$  & 
\\ \hline
$\sigma$ & $6.572$ &
\\ \hline
$\delta$ & $2 \cdot 10^{-5}$ &
\\ \hline
$C$ & $0.025$ & clipping constant
\\ \hline
$s$ & $\{ 10, 25, 50, 100, 300, 500, 689 \}$ & constant sample size sequence
\\ \hline
dataset & CIFAR$10$ & iid dataset
\\ \hline
$n$ & $5$ &  \# of nodes
\\ \hline
$\Upsilon$ & $k\geq i-1$ &  $1-$asynchronous round
\\ \hline

\end{tabular}
  \end{threeparttable}
\end{center}
\end{table}



\begin{figure}[!ht]
\captionsetup[subfigure]{labelformat=empty}
  \centering
  \subfigure[Non-convex.]{\includegraphics[width=0.575\textwidth]{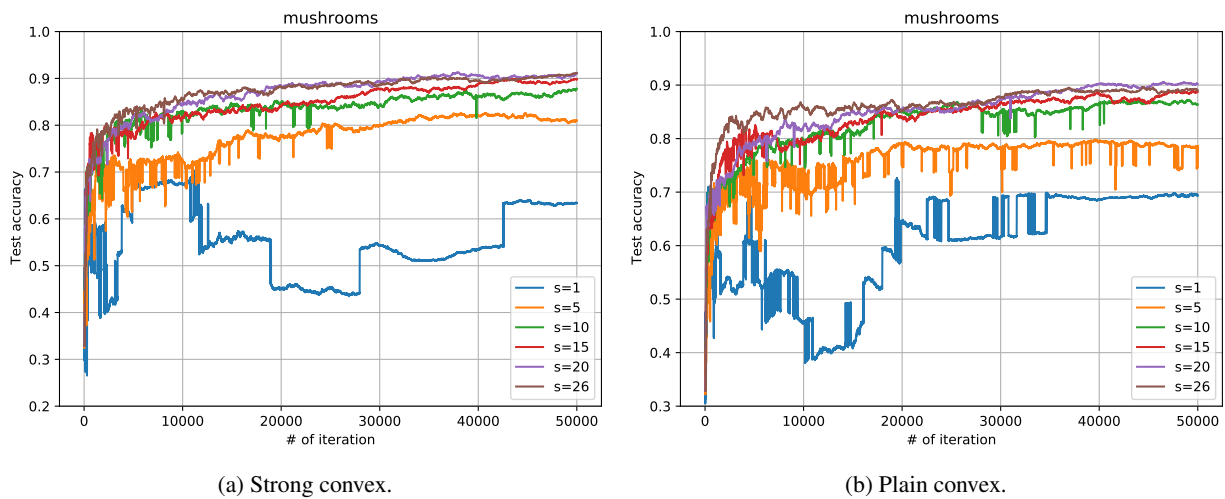}\label{ffig:asyncDP_mnist_sampling_method_0}}
  \caption{Effect of different constant sample size sequences}
  \label{fig:asyncDP_mnist_sampling_method}
\end{figure}

\begin{figure}[!ht]
\captionsetup[subfigure]{labelformat=empty}
  \centering
  \subfigure[Non-convex.]{\includegraphics[width=0.625\textwidth]{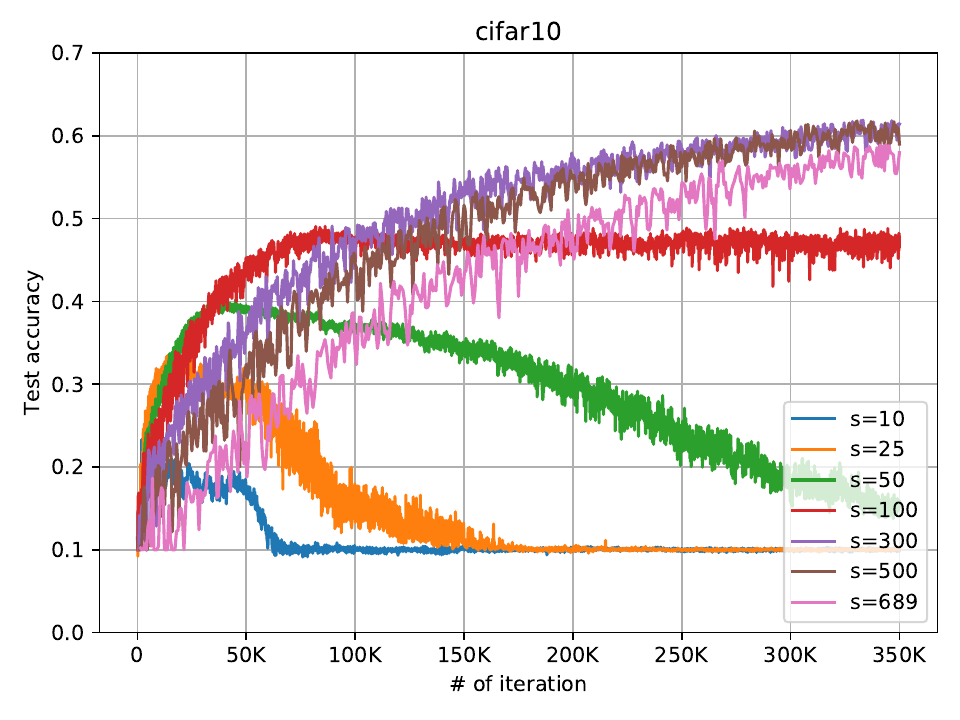}\label{ffig:asyncDP_cifar10_sampling_method_0}}
  \caption{Effect of different constant sample size sequences.}
  \label{fig:asyncDP_cifar10_sampling_method}
\end{figure}

We also conduct the experiment for the  non-convex objective function with MNIST and CIFAR$10$ data sets. 
The detailed parameter settings can be found in Table~\ref{tbl:basic_param_setting_noncvx} and Table~\ref{tbl:basic_param_setting_noncvx_cifar10}. 
Here, we again consider our asynchronous setting where each compute node is allowed to run fast and/or has small communication latency such that  
broadcast global models are at most $1$ local round in time behind. As can be seen from Figure~\ref{fig:asyncDP_mnist_sampling_method} (with MNIST data set), 
our proposed asynchronous DP-SGD still converges under small privacy budget. Moreover, when we use the constant sample size $s=370$ (this meets the upper bound for constant sample sizes for our small $N=60,000$ and small $\epsilon \approx 0.15$, see Theorem~\ref{thm:simpleF}), we can significantly reduce the communication cost compared to other constant sample sizes while keeping the test accuracy loss within 10\%. The constant sample size $s = 10$ (as well as $s \leq 10$) shows a worse performance while this setting requires more communication rounds, compared to other constant sample sizes.
We can observe the same pattern for CIFAR$10$ data set as shown in Figure~\ref{fig:asyncDP_cifar10_sampling_method}, where we can choose the constant sample size $s \leq 689$ with $N=50,000$ data points and $\epsilon \approx 0.5$. While the constant sample size $s$ satisfying $300 \leq s \leq 689$, the test accuracy gets the highest level while the constant sample size $s \leq 50$ deteriorates the performance of accuracy significantly.
This figure again confirms the effectiveness of our asynchronous DP-SGD framework towards a strong privacy guarantee for all types of objective function.

\subsubsection{Asynchronous DP-SGD with different levels of privacy budget}
\label{subsec:different_privacy_budget}

\begin{table}[!htb]
\caption{Different privacy budget settings for strongly and plain convex problems}
\label{tbl:basic_param_setting_epsilon}
\begin{center}
\begin{threeparttable}
\begin{tabular}{|c|c|c|}
\hline
Privacy budget $(\epsilon, \delta)$ & $\sigma$ & Sample size $s$ \\ \hline \hline
$(0.04945, 0.0001)$ & $19.29962$ & $26$
\\ \hline
$(0.1, 0.0001)$ & $13.06742$ & $55$
\\ \hline
$(0.25, 0.0001)$ & $8.59143$ & $103$
\\ \hline
$(0.5, 0.0001)$ & $6.05868$ & $168$
\\ \hline
$(1.0, 0.0001)$ & $4.27273$ & $265$
\\ \hline
$(2.0, 0.0001)$ & $3.03241$ & $400$
\\ \hline
\end{tabular}
  \end{threeparttable}
 \end{center}
\end{table}

We conduct the following experiments to compare the effect of our DP-SGD framework for different levels of privacy budget $\epsilon$
including the non-DP setting (i.e., no privacy at all, hence, no noise). The purpose of this experiment is to show that the test accuracy degradation is at most 10\% even if we use very small $\epsilon$. The detailed constant sample sequence $s$ and noise level $\sigma$ based on Theorem~\ref{thm:simpleF} are illustrated in Table~\ref{tbl:basic_param_setting_epsilon}. Other parameter settings, such as initial stepsize $\eta_0$, 
 are kept the same as in Table~\ref{tbl:basic_param_setting}.

\begin{figure}[!ht]
  \centering
  \subfigure[Strong convex.]{\includegraphics[width=0.5\textwidth]{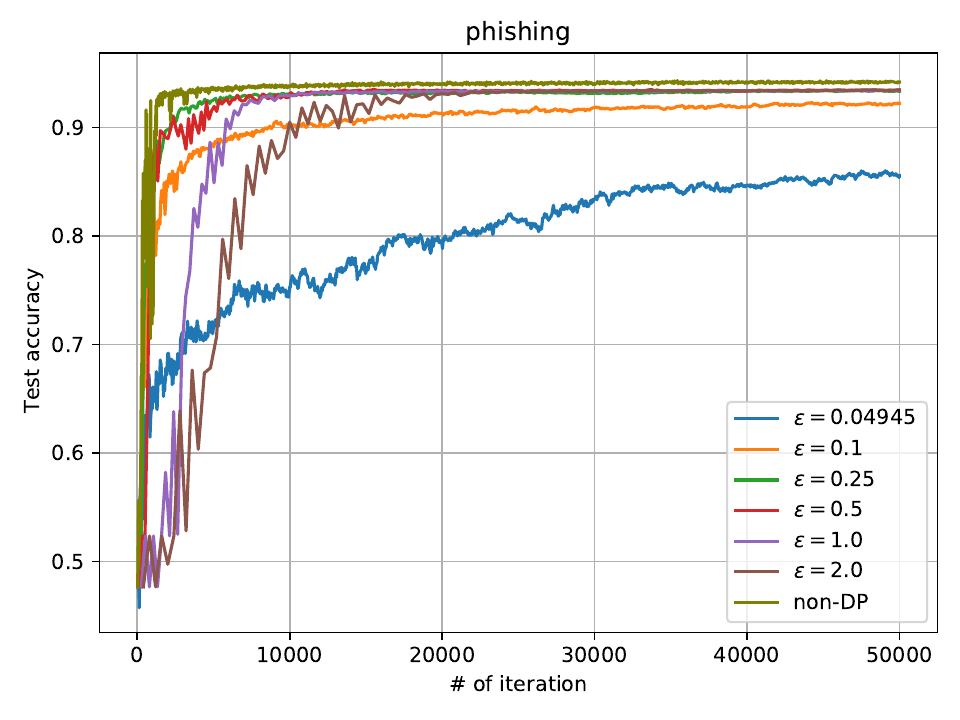}\label{fig:asyncDP_phishing_epsilon_0}}
  \hfill
  \subfigure[Plain convex.]{\includegraphics[width=0.5\textwidth]{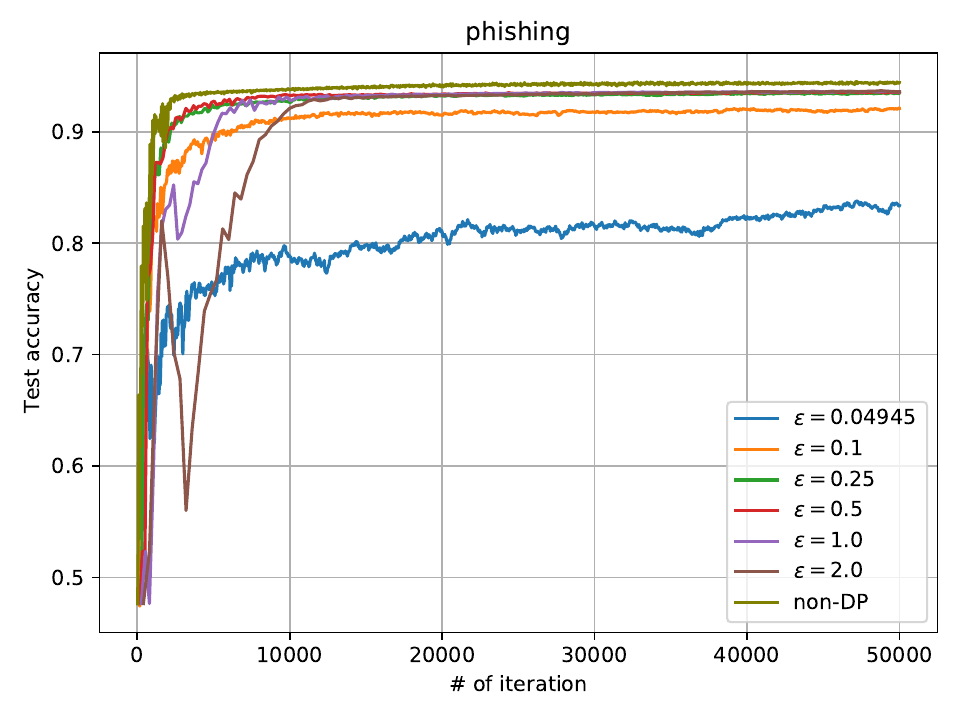}\label{fig:asyncDP_phishing_epsilon_1}}
  \caption{Effect of different levels of privacy budgets $\epsilon$ and non-DP settings}
  \label{fig:asyncDP_phishing_epsilon_method}
\end{figure}

\begin{figure}[!ht]
  \centering
  \subfigure[Strong convex.]{\includegraphics[width=0.5\textwidth]{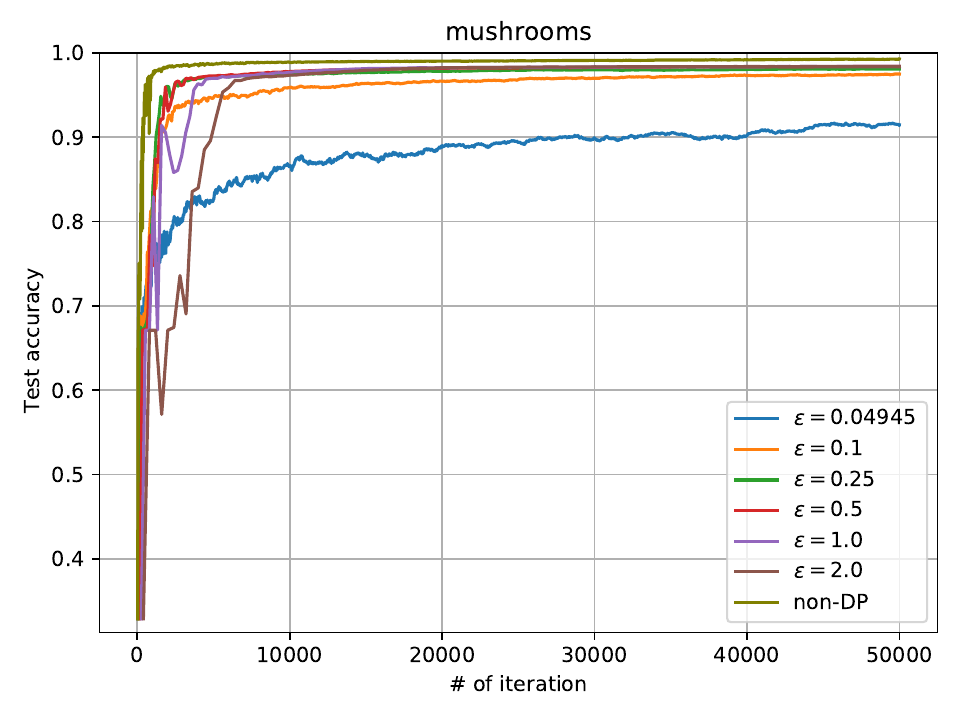}\label{fig:asyncDP_mushrooms_epsilon_0}}
  \hfill
  \subfigure[Plain convex.]{\includegraphics[width=0.5\textwidth]{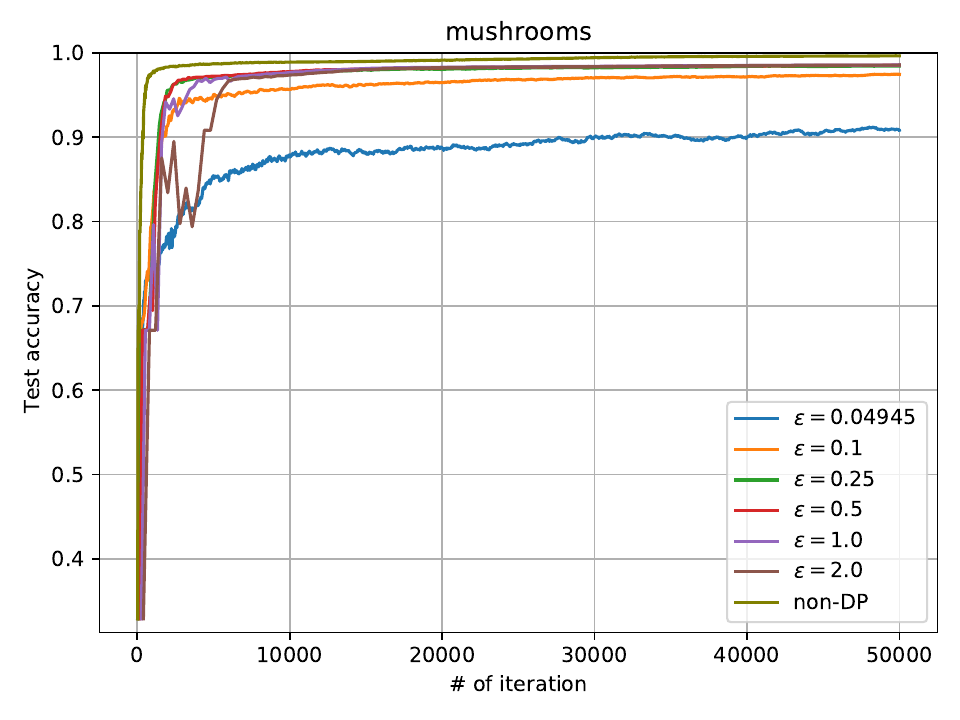}\label{fig:asyncDP_mushrooms_epsilon_1}}
  \caption{Effect of different levels of privacy budgets $\epsilon$ and non-DP settings}
  \label{fig:asyncDP_mushrooms_epsilon_method}
\end{figure}

As can be seen from Figures~\ref{fig:asyncDP_phishing_epsilon_method} and Figure~\ref{fig:asyncDP_mushrooms_epsilon_method}, the test accuracy degradation is about 10\% for $\epsilon=0.04945$ compared to the other graphed privacy settings and  non-DP setting. Privacy budget $\epsilon=0.1$, still significant smaller than what is reported in literature, comes very close to the maximum attainable test accuracy of the non-DP setting.

\begin{figure}[!ht]
\captionsetup[subfigure]{labelformat=empty}
  \centering
  \subfigure[Non-convex.]{\includegraphics[width=0.575\textwidth]{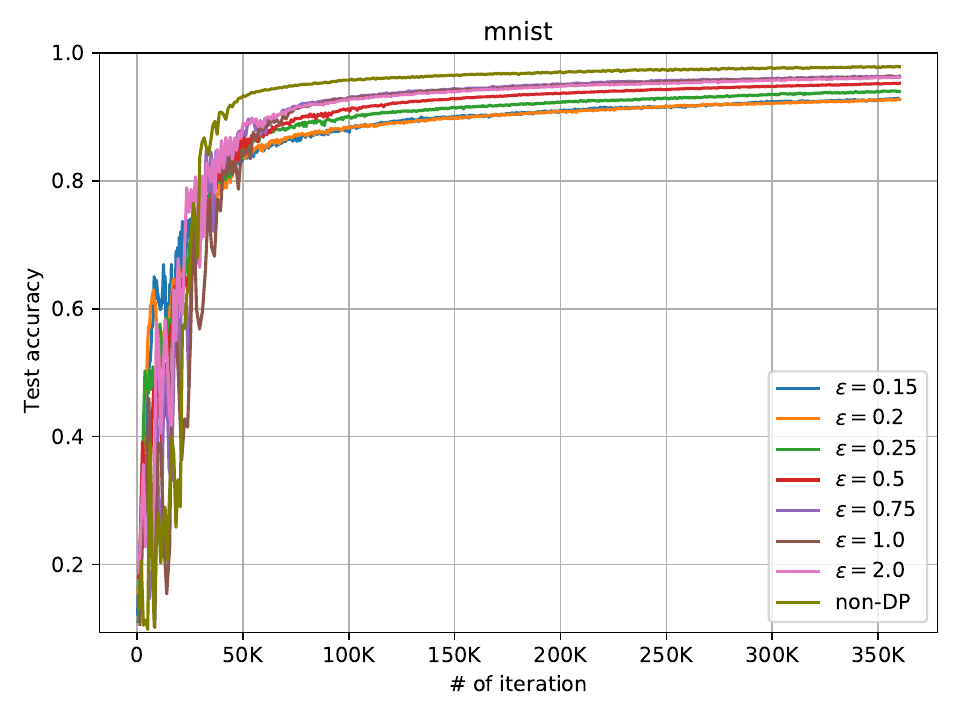}\label{fig:asyncDP_mnist_epsilon_0}}
  \caption{Effect of different levels of privacy budgets $\epsilon$ and non-DP settings}
  \label{fig:asyncDP_mnist_epsilon}
\end{figure}

\begin{figure}[ht!]
\captionsetup[subfigure]{labelformat=empty}
  \centering
  \subfigure[Non-convex.]{\includegraphics[width=0.575\textwidth]{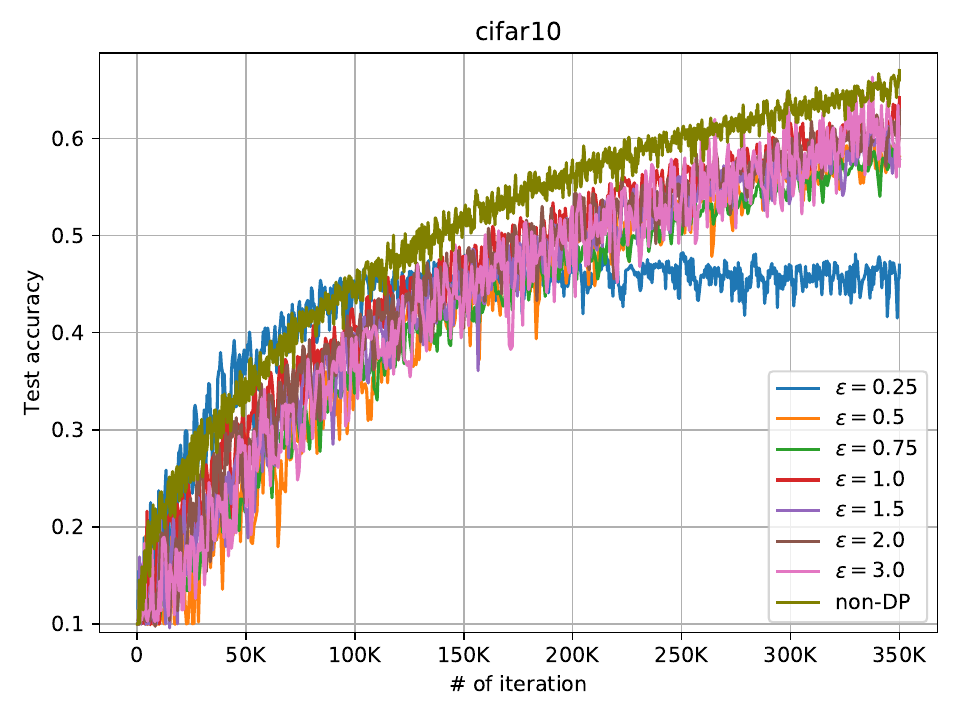}\label{fig:asyncDP_cifar10_epsilon_0}}
  \caption{Effect of different levels of privacy budgets $\epsilon$ and non-DP settings}
  \label{fig:asyncDP_cifar10_epsilon}
\end{figure}

We ran the same  experiment for the non-convex objective function. The detailed setting of different privacy budgets is shown in Table~\ref{tbl:basic_param_setting_epsilon_noncvx}. Note that we also set the asynchronous behavior to be $1$ asynchronous round, and the total of iterations on each compute node is $K=360,000$.  Other parameter settings for the non-convex case, such as initial stepsize $\eta_0$, 
 are kept the same as in Table~\ref{tbl:basic_param_setting_noncvx}.
As can be seen from Figure~\ref{fig:asyncDP_mnist_epsilon} with MNIST data set, the test accuracy loss with $\epsilon \approx 0.15$ is less than 10\% (the expected test accuracy degradation from utility graph at Figure~\ref{fig:asyncDP_mnist_utility}).
Another pattern can be found in Figure~\ref{fig:asyncDP_cifar10_epsilon}. By selecting $\epsilon \approx 0.5$ for CIFAR$10$ data set, the test accuracy reduces less than $10\%$, compared to the non-DP setting. Note that we use AlexNet for CIFAR$10$, which shows $\approx 0.74$ maximum test accuracy in practice\footnote{https://github.com/icpm/pytorch-cifar10}.


\begin{table}[ht!]
\caption{Different privacy budget settings for non-convex problem for MNIST data set}
\label{tbl:basic_param_setting_epsilon_noncvx}
\begin{center}
\begin{threeparttable}
\begin{tabular}{|c|c|c|}
\hline
Privacy budget $(\epsilon, \delta)$ & $\sigma$ & Sample size $s$ \\ \hline \hline
$(0.15007, 1.667 \cdot 10^{-5} ) $ & $12.10881$ & $370$
\\ \hline
$(0.2, 1.667 \cdot 10^{-5} ) $ & $10.48452$ & $460$
\\ \hline
$(0.25, 1.667 \cdot 10^{-5} ) $ & $9.37379$ & $543$
\\ \hline
$(0.5,  1.667 \cdot 10^{-5} )$ & $6.63120$ & $889$
\\ \hline

$(0.75,  1.667 \cdot 10^{-5} )$ & $5.41887$ & $1168$
\\ \hline

$(1.0, 1.667 \cdot 10^{-5} )$ & $4.69244$ & $1409$
\\ \hline
$(2.0, 1.667 \cdot 10^{-5} )$ & $3.31648$ & $2159$
\\ \hline
\end{tabular}
  \end{threeparttable}
 \end{center}
\end{table}

\begin{table}[!htb]
\caption{Different privacy budget settings for non-convex problem for CIFAR$10$ data set}
\label{tbl:basic_param_setting_epsilon_cifar10}
\begin{center}
\begin{threeparttable}
\begin{tabular}{|c|c|c|}
\hline
Privacy budget $(\epsilon, \delta)$ & $\sigma$ & Sample size $s$ \\ \hline \hline
$(0.25, 2.0 \cdot 10^{-5} ) $ & $9.29838$ & $417$
\\ \hline
$(0.5,  2.0 \cdot 10^{-5} )$ & $6.57192$ & $689$
\\ \hline

$(0.75,  2.0 \cdot 10^{-5} )$ & $5.36937$ & $909$
\\ \hline

$(1.0, 2.0 \cdot 10^{-5} )$ & $4.65014$ & $1099$
\\ \hline
$(1.5, 2.0 \cdot 10^{-5} )$ & $4.16111$ & $1267$
\\ \hline
$(2.0, 2.0 \cdot 10^{-5} )$ & $3.28831$ & $1690$
\\ \hline
$(3.0, 2.0 \cdot 10^{-5} )$ & $2.68273$ & $1994$
\\ \hline
\end{tabular}
  \end{threeparttable}
 \end{center}
\end{table}

These figures again confirm the effective performance of our DP-SGD framework, which not only conserves strong privacy, but also keeps a decent convergence rate to good accuracy, even for a very small privacy budget. 

\subsection{\texorpdfstring{Comparison to the $f$-DP Accountant}{}} \label{app:DPaccSim}

We have implemented a simplified differential privacy calculator based on Theorem \ref{thm:Main} for computing the optimal privacy budget $(\epsilon, \delta)$ given the training hyper-parameters $(\sigma, \theta, N, k, C)$. This calculator has the follow steps:

\begin{enumerate}
    \item Set $\delta = 1/N$, $\epsilon = \frac{2\ln{1/\delta}}{\sigma^2 -2}$. 
    
    \item Set $\gamma = 2$ (because $\gamma = 2 + O(\bar{\alpha})$) as the initial value.
    \item According to Theorem \ref{thm:Main}, we compute $T_{min}$ as a result of next steps (3, 4, and 5) as a lower bound on $T$ 
    as follows:
    \begin{itemize}
        \item Compute $\bar{\alpha} = \frac{\epsilon N}{\gamma K}$.
        \item Recompute the new $\gamma_{new} = \frac{2}{1-\bar{\alpha}} + 
\frac{2^4 \cdot \bar{\alpha}  }{1-\bar{\alpha}}
\left( \frac{\sigma}{(1-\sqrt{\bar{\alpha}})^2} +\frac{1}{\sigma(1-\bar{\alpha})-2e\sqrt{\bar{\alpha}}}\frac{e^3}{\sigma}
\right)e^{3/\sigma^2}$.
    \end{itemize}
    \item Repeat steps 3 until $\gamma_{new} - \gamma \leq 0.0001\gamma$ or inequality (\ref{minK}) is violated. In the latter case the calculator cannot find a solution of a set of hyperparameters that satisfies the privacy constraint $(\epsilon,\delta)$ and we lower $\sigma$ and increase $\epsilon$ accordingly (in step 1) and repeat steps 2, 3, and 4.
    \item Based on inequality (\ref{minT}) we compute  minimal value $T_{min} = \frac{\gamma\theta^2}{\epsilon} k^2$. From the asymptotic tightness analysis in section \ref{app:tight} we learn that $T_{min}$ can at most be lowered to 
$T_{min\_asym} = \frac{\theta^2 k^2}{2\epsilon}$.
    \item Corresponding to $T_{min}$ and $T_{min\_asym}$, we set $s_{max} = \frac{kN}{T_{min}}$ and $s_{max\_asym} = \frac{kN}{T_{min\_asym}}$
    \item We obtain the resulting set of parameters $(\epsilon,\delta,\sigma,\gamma,\theta,k,N,s,T)$ for $(s,T)$ equal to  $(s_{max},T_{min})$ and $(s_{max\_asym},T_{min\_asym})$ respectively.
\end{enumerate}
This calculation helps us planning ahead the number of rounds $T$ and the sampling rate $s$ and we can choose $\epsilon_{target}$ by defining an initial $\sigma$. 

Given the hyper-parameters defined in Tables \ref{tbl:basic_param_setting}, \ref{tbl:basic_param_setting_noncvx} and \ref{tbl:basic_param_setting_noncvx_cifar10}, we  use  this calculator to compute $\epsilon_{target}$   and compared this with the value $\epsilon_{f-DP}$ as a result from the exact/tight $f$-DP accountant from \cite{dong2021gaussian}, see \href{https://github.com/tensorflow/privacy/blob/master/tensorflow_privacy/privacy/analysis/gdp_accountant.py}{github.com/tensorflow/privacy}. This leads to Tables \ref{tbl:accountant_comparison_ICML_table_2}, \ref{tbl:accountant_comparison_ICML_table_3}, and \ref{tbl:accountant_comparison_ICML_table_4}.
Since $\epsilon_{f-DP}$ is tight, we conclude that $\epsilon_{target}$ for $T_{min\_asymp}$ cannot be achieved, hence, our provided formula for $T_{min}$ is indeed tight up to a factor $2\gamma$ as mentioned in the main body and studied in Appendix \ref{app:tight}.

\begin{table}[H]
\centering
\begin{tabular}{|l|l|l|}
\hline
 $T_{min}$ and $T_{min\_asym}$  & $\frac{\gamma \theta^2 k^2}{ \epsilon}$ & $\frac{\theta^2k^2}{2\epsilon}$ \\ \hline
 $s_{max} = \frac{kN}{T_{min}}$ and $s_{max\_asym} = \frac{kN}{T_{min\_asym}}$ 
 &  26  & 198\\
 \hline
$\epsilon_{target}$ & 0.0497 & 0.0497 \\ \hline
$\epsilon_{f-DP}$ & 0.0105 & 0.0533 \\ \hline
Multiplication factor $(\epsilon_{target}/\epsilon_{f-DP})$ & 4.7234 & 0.9328 \\ \hline

\end{tabular}
\caption{Comparison of $\epsilon_{target}$ with $\epsilon_{f-DP}$ from the $f$-DP accountant
for Table \ref{tbl:basic_param_setting}
where $\theta = 1,\delta = 1/10000$ and  $\sigma = 19.29962$ for LIBSVM dataset ($N=10000$), $k = 5$.}
\label{tbl:accountant_comparison_ICML_table_2}
\end{table}

\begin{table}[H]
\centering
\begin{tabular}{|l|l|l|}
\hline
 $T_{min}$ and $T_{min\_asym}$  & $\frac{\gamma \theta^2 k^2}{ \epsilon}$ & $\frac{\theta^2k^2}{2\epsilon}$ \\ \hline
 $s_{max} = \frac{kN}{T_{min}}$ and $s_{max\_asym} = \frac{kN}{T_{min\_min}}$ 
 &  288  & 3042 \\ 
 \hline
$\epsilon_{target}$ & 0.1521 & 0.1521 \\ \hline
$\epsilon_{f-DP}$ & 0.0389 & 0.2097 \\ \hline
Multiplication factor $(\epsilon_{target}/\epsilon_{f-DP})$ & 3.9147 & 0.7257 \\ \hline

\end{tabular}
\caption{Comparison of $\epsilon_{target}$ with $\epsilon_{f-DP}$ from the $f$-DP accountant
for Table \ref{tbl:basic_param_setting}
where $\theta = 1,\delta = 1/60000$ and  $\sigma = 12.10881$ for MNIST dataset ($N=60000$), $k = 6$.}
\label{tbl:accountant_comparison_ICML_table_3}
\end{table}

%


\begin{table}[H]
\centering
\begin{tabular}{|l|l|l|}
\hline
 $T_{min}$ and $T_{min\_asym}$  & $\frac{\gamma \theta^2 k^2}{ \epsilon}$ & $\frac{\theta^2k^2}{2\epsilon}$ \\ \hline
 $s_{max} = \frac{kN}{T_{min}}$ and $s_{max\_asym} = \frac{kN}{T_{min\_asym}}$ 
 &  406  & 7504 \\ 
 \hline
$\epsilon_{target}$ & 0.5253 & 0.5253 \\ \hline
$\epsilon_{f-DP}$ & 0.1133 & 0.8239 \\ \hline
Multiplication factor $(\epsilon_{target}/\epsilon_{f-DP})$ & 4.6372 & 0.6376 \\ \hline

\end{tabular}
\caption{Comparison of $\epsilon_{target}$ with $\epsilon_{f-DP}$ from the $f$-DP accountant
for Table \ref{tbl:basic_param_setting}
where $\theta = 1,\delta = 1/50000$ and  $\sigma = 6.572$ for CIFAR10 dataset ($N=50000$), $k = 7$.}
\label{tbl:accountant_comparison_ICML_table_4}
\end{table}

\section{Towards Dynamically Adapting  DP-SGD Parameters} \label{sec:practice}

\begin{figure}[ht!]
\begin{center}
    \includegraphics[width=0.8\textwidth]{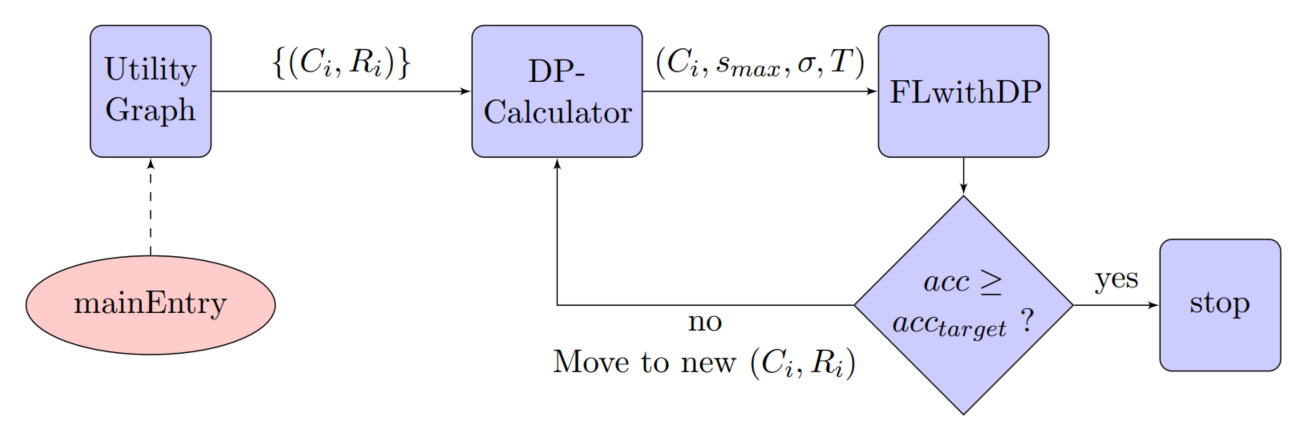}
\end{center}
\caption{The flow chart of our proactive (asynchronous) DP-SGD framework.}
\label{fig:flow_chart}
\end{figure}

In order to apply our theory in practice, we introduce a flow chart of our asynchronous DP-SGD learning from a client's perspective in Figure~\ref{fig:flow_chart}. Each client wants to participate in the collective learning of a global model that achieves a sufficient target accuracy $acc_{target}$ with respect to their test data set. That is, when a client tests the final global model against its own private test set, then the client is satisfied if an accuracy $acc_{target}$ is achieved. Such accuracy can only be achieved for certain combinations of noise $\sigma$ and clipping constant $C$. In particular, the final round introduces Gaussian noise with deviation $\sigma$ and this leads to inherent inaccuracy of the final global model. Experiments in Section \ref{sec:experiment} have shown that reducing the clipping constant $C$ allows a larger $\sigma$ for attaining a good target accuracy $acc_{target}$ (if a sufficient larger number $K$ of gradient computations have been executed). Here, we note that $C$ cannot be reduced indefinitely without hurting accuracy; this is because a reduced clipping constant into some extent plays the role of a reduced step size (or learning rate) and we know that convergence to an accurate global model must start with a large enough step size. Therefore,
before starting any learning, we need to understand how utility relates to $\sigma$ and the clipping constant $C$. This is reflected in the functionality ``Utility Graph'' in the flow chart of Figure \ref{fig:flow_chart}. Based on learning a local model (and based on a-priori information from learning models on similar data sets), ``Utility Graph'' produces a set of pairs $\{(C_i,R_i)\}$ together with a total number $K$ of gradient computations: For various clipping constants $C_i$, a range $R_i$ of possible $\sigma$ is output. That is, if $\sigma\in R_i$ for clipping constant $C=C_i$, then there is a good indication that this will lead to target accuracy $acc_{target}$ of the final model after the client has contributed  $K$ local gradient computations.

A second functionality in the flow chart is the ``DP Calculator''. This calculator takes the pair $(C_i,R_i)$ that allows the maximum possible $\sigma$ in $R_i$. This $\sigma$ defines the best possible $(\epsilon, \delta)$-DP curve given our formula $\sigma=\sqrt{2(\epsilon+\ln(1/\delta))/\epsilon}$. That is, it allows the ``smallest" possible $(\epsilon,\delta)$ pairs. The ``DP Calculator'' checks whether the maximum $\sigma$ allows the target differential privacy budget of the client defined by $(\epsilon_{target},\delta_{target})$. If not, then the client cannot participate in the collective learning. If the target differential privacy budget does fit, then, given $\sigma$, $\epsilon=\epsilon_{target}$, $\delta=\delta_{target}$, $K$, and the client's data set size $N$, the ``DP Calculator'' computes the maximum possible sample size $s_{max}$ according to the conditions in Theorem \ref{thm:simpleF}. This in turn results in a number of rounds $T=K/s_{max}$. We propose to choose the maximum possible sample size as this leads to the best accuracy/utility in our experiments. This is because $s_{max}$ leads to the smallest number of rounds $T$, hence, the smallest number of times noise is added and aggregated into the global model at the server. (Also, as a secondary objective, a smaller number of rounds means less round communication.)

As soon as the ``DP Calculator'' has calculated all parameters, the client executes $T$ rounds of DP-SGD with sample (mini-batch) size $s=s_{max}$. This is represented by the ``FLwithDP'' functionality in the flow chart of Figure \ref{fig:flow_chart}. Once the $T$ rounds are done, the client estimates the accuracy $acc$ of the last received global model based on the client's (public) test data set. Next, the client checks whether it is at least the target accuracy $acc_{target}$. If so, then the client stops participating. That is, with each new global model received by the server, the clients test whether the accuracy is satisfactory; if not, then the client will want to participate again. 
If the test accuracy $acc$ is not sufficient, then the ``DP Calculator'' will work with a new $(C_i,R_i)$. The next subsection details the computations by the ``DP Calculator'' and explains the feedback loop in the flow chart of Figure  \ref{fig:flow_chart}.

We remark that the client can use a complimentary differential privacy accountant to keep track of the exact privacy budget that has been spent.


\subsection{DP Calculator}
\label{appendix:DPcalculator}

We propose each local client to take control over its own privacy budget while making sure the locally measured test accuracy of the final global model is acceptable. The main idea is to start with an initial $\sigma=\sigma_0$ with appropriate clipping constant $C=C_0$ and an estimated number  $K=K_0$ of local gradient computations needed for convergence to ``sufficient'' test accuracy (utility). For local data set size $N$, we want to compute  proper parameter settings  including the batch size $s=s_0$ for each round and the total number of rounds $T=T_0$ (with $K=sT$). Once the $T$ rounds are finished, local test data is used to compute the test accuracy of the final global model. If the accuracy is not satisfactory, then $\sigma$ must be reduced to a lower $\sigma_1$ (and we may re-tune to a larger clipping constant $C_1$). This leads to an additional (estimated) number  of local $K_1$ gradient computations that need to be executed. The lower $\sigma$ corresponds to worse differential privacy since a lower $\sigma$ is directly related to a higher $\epsilon$ for given $\delta=1/N$. The local client is in control of what $\epsilon$ is acceptable -- and if needed, the local client simply stops participating  helping the central server learn a global model.

In order to apply our theory, we pretend as if the initial $T$ rounds used the lower $\sigma=\sigma_1$ -- this means that our analysis provides an advantage to the adversary as we assume less noise is used compared to what was initially  actually used. Hence, the resulting DP guarantee for $\sigma_1$ will hold for all $K=K_0+K_1$ local gradient computations. We use the new $\sigma=\sigma_1$ and $K$ together  with badge size $s_0$ for the first $T_0$ rounds to compute a new parameter setting for the next rounds; this includes the number $T_1$ of additional rounds (making $T=T_0+T_1$) and the new batch size $s_1$. The new batch size implies a new average $\bar{s}=(s_0T_0+s_1T_1)/(T_0+T_1)$ as well as a new variation $\theta_1>1$ of the sequence of batch sizes.

Once all $T=T_0+T_1$ rounds are finished (or equivalently all $K=K_0+K_1$ local gradient computations are finished), the local client again computes the test accuracy of the last global model. If not acceptable $\sigma$ is reduced again and we repeat the above process. If the test accuracy is acceptable, then the local client stops participating, that is, the local client stops gradient computations but continues to receive global models from the central server. As soon as the local client measures a new unacceptable local test accuracy, the client will continue the above process and starts a new series of rounds based on $\sigma$.

Stopping participation and later continuing if needed best fits learning problems over large data: Here, each local client  samples its own local data set according to the `client's behavior'. The local client wants to prevent as much leakage  of its privately selected local data set as possible. Notice that each local data set is too small for a local client to learn a global model on its own -- this is why local clients need to unite in a joint effort to learn a global model (by using distributed SGD). Assuming all samples are iid (all local data sets are themselves sampled from a global distribution), the final global model is not affected by having more or less contribution from local clients (as a result of different stopping and continuation patterns). Notice that if local data sets would be heterogeneous, then the final global model corresponds to a mix of all heterogeneous data sets and here it matters how much each local client participates (as this influences the mix).

The above procedure describes a proactive method for adjusting $\sigma$ to lower values if the locally measured test accuracy is not satisfactory. Of course, the local client sets an a-priori upper bound $\epsilon_{target}$  on the $\epsilon$ (with $\delta=\delta_{target}=1/N$), its privacy budget. This privacy budget cannot be exceeded, even if the local test accuracy becomes unsatisfactory.

We notice that our theory is general in that it can be used to analyse varying sequences of batch sizes, which is needed for our proactive method. We now describe in detail how to calculate parameter settings according to our theorems: 
    
    Suppose the local client has already computed for $T_0+T_1+\ldots + T_{j-1}$ rounds with badge sizes $s_0,s_1,\ldots, s_{j-1}$, hence, $K_0=s_0T_0, K_1=s_1T_1, \ldots, K_{j-1}=s_{j-1}T_{j-1}$. The local client sets/fixes the total $K_{j}$ of gradient computations it wants to compute over the next $T_{j}$ rounds. We want to compute a new $s_{j}$ and $T_{j}$. Notice that $s_{j}=K_{j}/T_{j}$ and $\bar{s}= \sum_{i=0}^{j} s_iT_i/T$, where $T=\sum_{i=0}^{j} T_i$. We want to find a suitable $s_{j}$.

We start our calculation with $s_{j}=1$ and we rerun our calculation for bigger batch sizes until we reach a maximum.
Given a choice $s_{j}=s$, we execute the following steps (we base the calculator on the slightly more complex but more accurate Theorem \ref{thm:simpleF} of Appendix \ref{appDP}):

\begin{enumerate}
    \item Set $\delta = 1/N$, compute $T_{j}=K_{j}/s_{j}$ given the input values $K_{j}$ and $s_{j}$, compute the corresponding $\bar{s}$ (see above) together with corresponding $\theta=\max \{s_i\}/\bar{s}$. Compute $K=\sum_{i=0}^{j} K_i$.
    
    \item Set $\gamma = 2$ (because $\gamma = 2 + O(\bar{\alpha})$) as the initial value.
    \item According to Theorem \ref{thm:simpleF}, we compute $\epsilon$, $\sigma$, and $\bar{\alpha}$ as follows:
    \begin{itemize}
        \item Based on  inequality (\ref{treq3}), set $\epsilon$ as small as possible, that is, $\epsilon = \gamma \theta^2 \bar{s} \frac{K}{N^2}$.
        \item We distinguish two cases:
        \begin{description}
       \item[$j=0$:]  In case we want to determine  $s_0$, we compute $\sigma_0=\sigma$ where $\sigma$ meets (\ref{treq4}) with equality, that is, $\sigma = \sqrt{2 (\epsilon +\ln{1/\delta})/\epsilon}$. 
        \item[$j>0$:] 
        During previous computations we already selected a $\sigma_{j-1}$. As described above, we only perform these calculations if the corresponding test-accuracy is not satisfactory. For this reason we want a lower $\sigma_{j}<\sigma_{j-1}$. The local client chooses a smaller $\sigma_{j}$ with possibly a larger clipping constant $C_{j}$ for which better accuracy within $K_{j}$ local gradient computing steps is expected. 
        We compute $\epsilon$ as a solution of $\sigma_{j} = \sqrt{2 (\epsilon +\ln{1/\delta})/\epsilon}$ and set $\epsilon$ to the maximum of this solution and the minimal possible $\epsilon= \gamma \theta^2 \bar{s} \frac{K}{N^2}$ computed above.
         \end{description}
        \item Compute $\bar{\alpha} = \frac{\epsilon N}{\gamma K}$.
    \end{itemize}
    \item Recompute the new $\gamma_{new} = \frac{2}{1-\bar{\alpha}} + 
\frac{2^4 \cdot \bar{\alpha}  }{1-\bar{\alpha}}
\left( \frac{\sigma}{(1-\sqrt{\bar{\alpha}})^2} +\frac{1}{\sigma(1-\bar{\alpha})-2e\sqrt{\bar{\alpha}}}\frac{e^3}{\sigma}
\right)e^{3/\sigma^2}$.
    \item Repeat steps 3 and 4 with $\gamma$ replaced by $\gamma_{new}$ until $\gamma_{new} - \gamma \leq 0.0001\gamma$, that is, $\gamma$ has converged sufficiently. 
    \item  The resulting set of parameters $(\epsilon,\delta,\sigma,\gamma,\theta,K,N)$ can only be used if  inequalities (\ref{treq1}) and (\ref{treq2}) are satisfied and $\epsilon \leq \epsilon_{target}$. 
    \begin{itemize}
        \item 
   If these conditions are satisfied, then we save the  parameters $(s,\epsilon,\sigma)$ and rerun the above calculation for bigger sample size $s$. Otherwise, we output 
    $(s,\epsilon,\sigma)$ of the previous run (as this corresponds to the maximum $s$ and thus minimal number of communication rounds) and terminate: We set $\sigma_{j}=\sigma$, $s_{j}=s$, and $T_{j}=K_{j}/s_{j}$. Our theory proves that we satisfy $(\epsilon, \delta=1/N)$-differential privacy.
    \item It may be that even the minimal batch size $s_{j}=1$ does not result in valid parameters $(s,\epsilon,\sigma)$. This means that the local client cannot participate any more otherwise its required differential privacy guarantee cannot be met.
     \end{itemize}
\end{enumerate}


\end{document}